%% file: ms.tex
\documentclass[runningheads]{llncs}

\input{packages}
\input{macros}

\newcommand{\etal}{\textit{et al.}}
\newcommand{\ie}{\textit{i.e.}}

\tikzstyle{largewindow} = [red, line width=0.50mm]
\tikzstyle{smallwindow} = [red, line width=0.20mm]

\begin{document}
\pagestyle{headings}
\mainmatter
\def\ECCVSubNumber{4471}  

\title{Neural Image Representations for\\ Multi-Image Fusion and Layer Separation} 

\titlerunning{Neural Image Representations for Multi-Image Fusion and Layer Separation}
%
\author{Seonghyeon Nam \and
Marcus A. Brubaker \and
Michael S. Brown}
\authorrunning{S. Nam et al.}
%
\institute{York University\\
\email{snam0331@gmail.com}, \email{\{mab,mbrown\}@eecs.yorku.ca}}
\maketitle

\begin{abstract}
We propose a framework for aligning and fusing multiple images into a single view using neural image representations (NIRs), also known as implicit or coordinate-based neural representations. Our framework targets burst images that exhibit camera ego motion and potential changes in the scene. We describe different strategies for alignment depending on the nature of the scene motion---namely, perspective planar (\ie, homography), optical flow with minimal scene change, and optical flow with notable occlusion and disocclusion.  With the neural image representation, our framework effectively combines multiple inputs into a single canonical view without the need for selecting one of the images as a reference frame.  We demonstrate how to use this multi-frame fusion framework for various layer separation tasks. The code and results are available at \url{https://shnnam.github.io/research/nir}.
\end{abstract}


\section{Introduction and Related Work}
\input{figures/teaser}
Fusing multiple misaligned images into a single view is a fundamental problem in computer vision.  The underlying assumption for this task is that the multiple images represent varying viewpoints of the same scene, perhaps with small motion in the scene. Many computer vision tasks rely on multi-image fusion, such as image stitching~\cite{Lin:2015:AdaptiveStitching,Levin:2004:SeamlessStitching,Brown:2007:AutomaticPanoramic}, high dynamic range (HDR) imaging~\cite{Kalantari:2017:DeepHDR,Yan:2020:DeepHDRNonLocal,Yan:2019:AttentionHDR}, and image super-resolution~\cite{Bhat:2021:DeepBurstSR,Bhat:2021:NTIREBurstSR,Wronski:2019:HandheldMFSR}.
Most existing image fusion approaches work by first aligning the multiple images based on their assumed motion---for example, homography for planar or nearly planar scenes or optical flow for nonplanar scenes, or when objects in the scene move.  Traditionally, images are aligned to a reference image that is manually chosen among the input images. Since image pixels are represented in a 2D discrete sampled array, such transformations are approximated by interpolation techniques.

Recently, implicit or coordinate-based neural representations were proposed to represent images and videos as a function of pixel coordinates parameterized by multi-layer perceptrons (MLPs)~\cite{Tancik:2020:FFN,Sitzmann:2020:SIREN}.
This new type of image representation, which we call a neural image representation (NIR), is different from conventional discrete grid-based representations in that image signals are continuous with respect to spatial or spatio-temporal pixel coordinates.
Further, the resolution of images no longer depends on the size of the discrete grid, but rather the representational complexity of the MLP.
These representations have been actively studied particularly in view synthesis~\cite{Mildenhall:2020:NeRF,Sitzmann:2019:SRN,Pumarola:2021:D-NeRF,Li:2021:NSFF,Martin:2021:NeRFInTheWild,Park:2021:Nerfies,Park:2021:Hypernerf}, 3D geometry~\cite{Mescheder:2019:Occupancy,Park:2019:DeepSDF}, and image synthesis~\cite{Skorokhodov:2021:Adversarial,Anokhin:2021:ImageGenerators,Shaham:2021:SAPN}.

This work targets multi-frame fusion by leveraging the advantages offered by NIRs. As shown on the left in~\fref{fig:teaser}, we propose to train MLPs to reconstruct a canonical view based on multiple images.  Our approach incorporates image registration techniques into NIRs using coordinate transformations~\cite{Park:2021:Nerfies,Park:2021:Hypernerf}. Unlike existing multi-image fusion, our method does not need an explicit reference image. Instead, a virtual reference image is implicitly learned as the canonical view embedded within the neural representation. Since the space of canonical views is unbounded, all images can be fused regardless of the original image frame as shown in ~\fref{fig:teaser}. In addition, image transformation is achieved in a real-valued coordinate space without the need for interpolation. 


To demonstrate effectiveness of our NIR multi-image fusion, we apply our method to various applications of multi-image layer separation.
As shown in~\fref{fig:teaser}, the goal of multi-image layer separation is to decompose signals from multiple images into a single underlying scene image and interference layers to improve the visibility of the underlying scene.
Many approaches for different tasks have been studied, such as image demoir\'e~\cite{Kim:2020:C3Net,Xu:2020:AFN}, reflection removal~\cite{Li:2013:LiandBrown,Alayrac:2019:VisualCentrifuge,Liu:2020:LearningToSee,Liu:2020:LearningToSeeJournal}, fence removal~\cite{Liu:2020:LearningToSee,Liu:2020:LearningToSeeJournal}, and deraining~\cite{Jiang:2018:FastDeRain,Yang:2019:FCDN,Chen:2018:NTURain,Wei:2017:SE}.
Early works on these problems heavily rely on domain-specific priors for optimization, while recent approaches are driven by deep learning and a large amount of annotated data for supervision.

In our work, we cast the problem as an unsupervised optimization of NIRs.  Specifically, we fuse the underlying scene from the multiple images using NIRs.  Depending on the type of scene motion, we use different deformation strategies when computing the neural image representation for each frame.  To remove the interference layer, we propose two-stream NIRs.  In particular, the underlying ``clean layer'' image without interference is represented by one MLP, while a separate MLP is used to represent the interference layer(s).  We show that standard regularization terms -- for example, total variation -- can be used in the optimization of these NIRs to assist in the layer separation. We demonstrate the effectiveness of our approach on moir\'e removal, obstruction removal, and rain removal.  

Closely related to our approach is DoubleDIP~\cite{Gandelsman:2019:DoubleDIP}, which also studied image layer decomposition on a single or multiple image(s) using coupled deep image priors~\cite{Ulyanov:2018:DIP}.
DoubleDIP exploits self-similarity, an inductive bias in convolutional neural networks (CNNs), to separate different signals.  Our approach uses the parameterization of motion as a general prior to tackle multiple tasks.
Unlike semantic segmentation and matting~\cite{Gandelsman:2019:DoubleDIP,Lu:2021:Omnimatte}, we focus on disentangling low-level signals rather than semantic layers.

\noindent{\textbf{Contribution.}}~We describe a framework to perform multi-frame fusion and layer separation as learning a neural image representation.  We describe variations on the representation and optimization for different scene and camera conditions.  We also demonstrate how to apply this framework to handle several different types of layer separation tasks.  To the best of our knowledge, our work is the first to explicitly address multi-image fusion with neural image representations.

\section{Method Overview}
\input{figures/method_motion}
Neural image representations, also known as implicit or coordinate-based neural representations, have recently been proposed \cite{Sitzmann:2020:SIREN,Tancik:2020:FFN} as a way to represent RGB values of an image as a function of pixel coordinates parameterized by MLPs.  For multiple sequential images, this can be formulated as
\begin{equation}
    \image_{(x, y, t)} = f_{\thetaImage}(x, y, t),
\end{equation}
where $\image_{(x, y, t)}$ is the value at pixel $(x, y)$ in frame $t$ and $f_{\thetaImage}$ is an MLP with parameters $\thetaImage$.
Here each frame is nearly independent due to different values of $t$.

In our work, we assume our multiple images are captured quickly as a burst from a single camera.  Consequently, images are of approximately the same scene, but are expected to have small variations due to the motion of the camera and small amounts of motion in the scene. Furthermore, we do not expect notable variations in scene lighting, appearance, or colors due to the camera's onboard color manipulation. 

Within this context of burst images, our work aims to formulate $f_{\thetaImage}$ differently by learning a joint representation of multiple images using their spatio-temporal correlation.  To this end, we revisit well-established image registration and motion compensation techniques within the new domain of NIRs.  Specifically, our $f_{\thetaImage}$ learns a canonical view of the scene shared across images.
Each image in the burst sequence is modelled by a deformation of the canonical view---for instance, using a perspective planar transform (i.e., a homography) or pixel-wise optical flow.
Since the function is continuous and unbounded, it not only is able store the entire scene regardless of the size of 2D image grid, but also can be easily deformed by transforming input coordinates into a real-valued space.
The model is formally described as
\begin{equation}
    \image_{(x, y, t)} = f_{\thetaImage}(T_{g}(x, y, t)),
\end{equation}
where $\transform$ applies a coordinate transformation with parameters $\phi$.
The parameters of the coordinate transform could be fixed or themselves a function--- that is, $g = g_{\thetaTrans}(x,y,t)$ where $g_{\thetaTrans}$ is an MLP that computes the parameters of the coordinate transform.
The parameters of the MLPs, $\thetaTrans$ and $\thetaImage$, are optimized by minimizing the following pixel reconstruction loss:
\begin{equation}
    \lossRecon = \sum_{x,y,t}~\lVert \image_{(x, y, t)} - \imageGT_{(x, y, t)} \rVert^2_2,
    \label{eq:reconstruction}
\end{equation}
where $\imageGT$ is the original image ground truth.

The explicit parameterization of motion in neural representations enables the simultaneous learning of image and motion representations.
By minimizing~\eref{eq:reconstruction}, our neural representations learn the parameters of scene motion in an unsupervised manner.
More importantly, unlike conventional image registration and motion compensation techniques, our approach does not require a reference image to be selected from the burst input.  Instead, our model learns a virtual reference view of the scene implicitly.

We next show how to extend our multi-frame alignment framework for use in layer separation tasks. In particular, we target tasks where input images are modeled as a combination of two layers: (1) the desired underlying scene image and (2) the undesired corruption in the form as an interference layer. We assume that the contents of the underlying scene remains similar over the multiple images, while the interference layer changes.
To do this, we propose a two-stream architecture for NIRs, with one component that captures the static scene and another that captures the interference.
In the following, we describe our method in detail.


\subsection{NIRs for Multi-Image Fusion}
\input{figures/vis_motion}
\fref{fig:method_motion} shows the overview of the NIRs for multi-image fusion.
We propose three kinds of parameterization according to the assumption of the scene: (a) homography-based NIRs, (b) occlusion-free flow-based NIRs, and (c) occlusion-aware flow-based NIRs.

\paragraph{Homography-based NIRs.}
In case of planar, rigid scenes that are moving globally as shown in \fref{fig:method_motion} (a), we can use a homography as the coordinate transformation.
As shown in the figure, the function $g_{\thetaTrans}$ is learned to estimate parameters of a homography matrix $\homography$ for each frame.
Then the predicted image using the homography-based NIRs is described as:
\begin{equation}
    \image_{(x, y, t)} = f_{\thetaImage}(\homography_t[x, y, 1]^T),
    \label{eq:homography}
\end{equation}
where $\homography_t$ is a 3$\times$3 linear matrix represented as $\homography_t = g_{\thetaTrans}(t)$.
Since $\homography_t$ is applied globally regardless of spatial coordinates, $g_{\thetaTrans}(t)$ only takes $t$ as input.
We omit the normalization of output coordinates in the homography transform for simplicity.

\fref{fig:vis_motion} (a) shows a visualization of the NIR estimated from nine burst images of a distant scene, captured with a horizontally moving camera.
As can be seen in the figure, the homography-based NIR automatically stitches all the images in a single view only using a reconstruction loss.
A single frame $t$ can be recreated by transforming the canonical view using the output of the $g_{\thetaTrans}(t)$ homography matrix.

\paragraph{Occlusion-free flow-based NIRs.}
In many cases, a scene will not be planar or move together rigidly.  However, in burst imagery because frames are temporally close, the motions are likely to be small.
To handle this,  we use a dense optical flow representation to model the per-pixel displacement of scene, which is represented by the displacement of $x$ and $y$ coordinates as shown in~\fref{fig:method_motion} (b).
We assume that the motion is small enough to cause minimal occlusions and disocclusions.
In this case $\phi$ represents an $xy$-displacement that is computed by $g_{\thetaTrans}(x,y,t)$ for each $(x, y, t)$.
Formally, $T(x,y,t) = (x + \Delta x_t, y + \Delta y_t)$ where $(\Delta x_t, \Delta y_t) = g_{\thetaTrans}(x, y, t)$ are the displacement of $x$ and $y$ coordinates.
An output pixel can be computed as:
\begin{equation}
    \image_{(x, y, t)} = f_{\thetaImage}(x + \Delta x_t, y + \Delta y_t).
    \label{eq:occlusion_free_flow}
\end{equation}
In addition to the reconstruction loss in~\eref{eq:reconstruction}, we use a total variation (TV) regularization for the flow smoothness, which is described as
\begin{equation}
    \lossTVFlow = \sum~\lVert J_{g_{\thetaTrans}}(x, y, t) \rVert_1,
    \label{eq:tvflow}
\end{equation}
where $J_{g_{\thetaTrans}}(x, y, t)$ is a Jacobian matrix that consists of gradients of $g_{\thetaTrans}$ with respect to $x$, $y$, and $t$.

\paragraph{Occlusion-aware flow-based NIRs.}
Since the canonical view of the occlusion-free flow-based NIRs is in a 2D plane, it is not enough to store extra information when a scene is occluded or disoccluded.
To address such cases, we add an additional dimension $w$ to the canonical view as shown in \fref{fig:method_motion} (c). Intuitively, different versions of a scene at a certain position caused by occlusion are stored at different values of $w$, while occlusion-irrelevant pixels are stored at the same value of $w$ and shared across images.
This is achieved by regularizing the Jacobian of $g_{\thetaTrans}$ in~\eref{eq:tvflow}.
With $w$, the output image is rendered by the following equation:
\begin{equation}
    \image_{(x, y, t)} = f_{\thetaImage}(x + \Delta x_t, y + \Delta y_t, w_t).
    \label{eq:occlusion_aware_flow}
\end{equation}
\fref{fig:vis_motion} (b) shows a visualization of a learned $xy$-flow map, $w$ map, and canonical views at different values of $w$ after training five consecutive images in~\cite{Perazzi:2016:DAVIS}.
Since the car is moving in the scene, the $xy$-flow map shows spatially varying optical flow on the car.
The $w$ map shows different values in regions of large motion (e.g., wheels), transient lighting effects (e.g., specularities and reflections), and regions that undergo occlusion or disocclusion.
This can be seen more clearly by visualizing the canonical view, with different values of $w$ as shown in the bottom of the figure.

\subsection{Two-Stream NIRs for Layer Separation}
\input{figures/method_twostream}
We now extend NIRs to multi-image layer separation tasks.
\fref{fig:method_twostream} shows the overview of our two-stream NIRs.
We model the images as the combination of two signals,
\begin{equation}
\begin{cases}
    \scene_{(x, y, t)} &= f^1_{\thetaScene}(T_g(x, y, t)),\\
    \interf_{(x, y, t)} &= f^2_{\thetaInterf}(x, y, t)
\end{cases}
\end{equation}
where $f^1_{\thetaScene}$ and $f^2_{\thetaInterf}$ are two different MLPs used to represent the scene and corrupting interference, respectively.

Since we usually have the knowledge of scene motion, we use an explicit parameterization of motion for $f^1_{\thetaScene}$---for example, a homography or a flow field.  To model the interference layers, we use an unconstrained form of MLP for $f^2_{\thetaInterf}$ to store contents that violate the motion in $f^1_{\thetaScene}$, that is beneficial for interference patterns difficult to model.
The generic form of image formation is described as
\begin{equation}
    \image_{(x, y, t)} = \scene_{(x, y, t)} + \interf_{(x, y, t)},
\label{eq:image_formation}
\end{equation}
but the specifics can vary depending on the task.
Due to the flexibility of $f^2_{\thetaInterf}$, it can potentially learn the full contents of the images as a ``video'', effectively ignoring $f^1_{\thetaScene}$.
To prevent this, we regularize $f^2_{\thetaInterf}$ using
\begin{equation}
    \lossftwo = \sum~\lVert \interf_{(x, y, t)} \rVert_1.
\end{equation}

Directly incorporating a spatial alignment into the NIR optimization may appear inefficient at first glance, especially compared to methods that first apply conventional homography and or optical flow estimation and then perform some type of image fusion. However, in the case of corrupted scenes, it is often challenging to estimate the motion of the underlying clean image with the conventional methods.  For instance, existing methods often rely heavily on multiple stages of refinement of motion~\cite{Liu:2020:LearningToSee} to pre-process images to assist with the alignment step. Our method tackles the problem jointly, by incorporating the scene alignment jointly with a layer separation through the benefits of NIRs.

\section{Applications}
We now show the effectiveness of our method on various multi-image layer separation tasks.
Please refer to the supplementary material for more results.

\subsection{Moir\'e Removal}
Moir\'e is a common pattern of interference, often seen when taking a photo of monitor or screen using a digital camera. Moir\'e patterns are caused by the misalignment of the pixel grids in the display and camera sensor.
Burst images usually capture temporally varying moir\'e patterns as camera motion changes the alignment of the sensor and screen and hence the interference pattern.
Typically the movement of the scene in burst images follows homography transform as the screen is planar.
We show that our two-stream NIRs are able to effectively separate the underlying scene and moir\'e pattern.

\paragraph{Formulation.}
We parameterize $f^1_{\thetaScene}$ as a homography-based NIR in~\eref{eq:homography}.
The image formation follows the basic form in~\eref{eq:image_formation}, where we use signed values in the range of $[-1, \, 1)$ for the output of both $\scene_{(x, y, t)} \in \mathbb{R}^3$ and $\interf_{(x, y, t)} \in \mathbb{R}^3$.
The signed output for $\interf_{(x, y, t)}$ is particularly useful to represent color bands of moir\'e patterns.
To further prevent scene content from appearing in both $\scene_{(x, y, t)}$ and $\interf_{(x, y, t)}$, we adopt an exclusion loss used in~\cite{Gandelsman:2019:DoubleDIP,Zhang:2018:ExclusionLoss} to encourage the gradient structure of two signals to be decorrelated.  This is formulated as
\begin{equation}
    \lossExcl = \sum~\lVert \Phi(J_{f^1}(x, y), J_{f^2}(x, y)) \rVert^2_2,
\end{equation}
where $\Phi(J_{f^1}(x, y), J_{f^2}(x, y))=\mathrm{tanh}(N_1 J_{f^1}(x, y)) \otimes \mathrm{tanh}(N_2 J_{f^2}(x, y))$, and $\otimes$ is an element-wise multiplication.
$N_1$ and $N_2$ are normalization terms~\cite{Zhang:2018:ExclusionLoss}.
We optimize $\thetaTrans$, $\thetaScene$, and $\thetaInterf$ using the following training objective:
\begin{equation}
    \lossMoire = \lossRecon + \lambdaftwo\lossftwo + \lambdaExcl\lossExcl,
\end{equation}
where $\lambdaftwo$ and $\lambdaExcl$ are hyperparameters.
We use an MLP with ReLU activation for $g_{\thetaTrans}$ and a SIREN~\cite{Sitzmann:2020:SIREN} for $f^1_{\thetaScene}$ and $f^2_{\thetaInterf}$.

\input{tables/moire}
\input{figures/result_moire}

\paragraph{Experiments.}
Since there are no publicly available datasets for multi-frame screen-captured moir\'e images, we synthesize a dataset from clean images.
To do this, we use the Slideshare-1M~\cite{Araujo:2016:Slideshare1M} dataset, which consists of approximately one million images of lecture slides, to mimic content likely to be captured by students.
Using this dataset, we synthesize 100 test sequences of five burst images for testing following the synthesis procedure in~\cite{Liu:2018:MoireSynthesis}.
For comparison, we compare AFN~\cite{Xu:2020:AFN} and C3Net~\cite{Kim:2020:C3Net}, state-of-the-art deep learning methods, which are trained by our synthetic training set containing 10,000 sequences.
We additionally evaluate Double DIP to compare unsupervised approaches.


\Tref{table:moire} shows a quantitative comparison of methods on the synthetic test set.
In addition to PSNR and SSIM, we compare a normalized cross-correlation (NCC) and structure index (SI).
Even though our method does not outperform AFN, the performance is significantly better than C3Net and Double DIP.
However, notably our method is unsupervised---that is, it does not use a training set of images.
This is in contrast to AFN and C3Net, which require explicit supervision or clean and moir\'e corrupted images.
\fref{fig:moire} shows a qualitative evaluation on real images.
As can be seen, our method outperforms all the baselines on real images.
The performance of AFN and C3Net is degraded because they are not trained on real images.
Double DIP fails to decompose the underlying scene and moir\'e pattern since it relies on an inductive bias in convolutional neural networks, which is not enough to separate complex signals.
Our method removes a moir\'e pattern by restricting the movement of the scene to homography, which acts as a strong prior of moir\'e removal.

\subsection{Obstruction Removal}
The goal of obstruction removal~\cite{Liu:2020:LearningToSeeJournal,Xue:2015:Computational} is to eliminate foreground objects or scenes that hinder the visibility of the background scene.  Obstructions can be in the form of reflection on a window in front of the scene or a physical object, such as a fence.
We apply the two-stream NIRs based on occlusion-free optical flow to a reflection and fence removal.
In this case, the background scenes are not planar, but the movement of the scene is small enough to ignore occlusion.
Similarly to moir\'e removal, we decompose the reflection and fence layer using the fact that they move differently to the background scene.

\paragraph{Formulation.}
We use the occlusion-free flow-based NIRs in~\eref{eq:occlusion_free_flow} for $f^1_{\thetaScene}$.
For reflection removal, we use the image model in~\eref{eq:image_formation}, where $\scene_{(x, y, t)} \in \mathbb{R}^3$ and $\interf_{(x, y, t)} \in \mathbb{R}^3$ are in the range of $[0, 1)$.
We use the following combination of loss functions as a training objective:
\begin{equation}
\begin{split}
    \lossRefl = &\lossRecon + \lambdaTVFlow\lossTVFlow\\
    &+ \lambdaftwo\lossftwo + \lambdaExcl\lossExcl,
\end{split}
\end{equation}
where $\lambdaTVFlow$ is a hyperparameter.
For a fence removal, we use a different image model described as
\begin{equation}
    \image_{(x, y, t)} = (1 - \alpha_{(x, y, t)})\scene_{(x, y, t)} + \alpha_{(x, y, t)}\interf_{(x, y, t)},
\end{equation}
where $(\alpha_{(x, y, t)}, \interf_{(x, y, t)}) = f^2_{\thetaInterf}(x, y, t)$, and $\scene_{(x, y, t)} \in \mathbb{R}^3$ and $\interf_{(x, y, t)} \in \mathbb{R}^3$ are in the range of $[0, 1)$.
$\alpha_{(x, y, t)} \in \mathbb{R}$ is an alpha map of the fence layer in the range of $[0, 1)$.
The training objective is described as:
\begin{equation}
    \lossFence = \lossRecon + \lambdaTVFlow\lossTVFlow + \lambdaftwo\lossftwo.
\end{equation}
We used SIREN for all coordinate functions.

\paragraph{Experiments.}
\fref{fig:reflection} shows qualitative results of our method and existing approaches.
We use real images in~\cite{Li:2013:LiandBrown} for testing.
The methods of Li and Brown~\cite{Li:2013:LiandBrown} and Alayrac et al.~\cite{Alayrac:2019:VisualCentrifuge} are designed for reflection removal, and the method in~\cite{Liu:2020:LearningToSeeJournal} is a general approach for obstruction removal.
As can be seen, our method is able to accurately decompose the background scene and reflection compared with the baseline methods.
\fref{fig:fence} shows a qualitative comparison of fence removal on real images in~\cite{Liu:2020:LearningToSeeJournal}.
Our method achieves comparable quality of results to learning-based methods that heavily rely on large amounts of data and supervision.

\input{figures/result_reflection}
\input{figures/result_fence}

\subsection{Rain Removal}
To show the effectiveness of the occlusion-aware flow-based NIRs, we address the problem of multi-image rain removal as the task deals with various kinds of scenes, from static scenes to dynamic scenes.
Since rain streaks move fast and randomly, the streaks impact the smoothness of the scene motion. We exploit this prior knowledge of the randomness of rain streaks observed in multiple images by imposing a smoothness regularization on the scene flow map.

\paragraph{Formulation.}
We use the occlusion-aware flow-based NIRs in~\eref{eq:occlusion_aware_flow} as a formulation of $f^1_{\thetaScene}$.
Since rain streaks are achromatic, we use the following image formation:
\begin{equation}
    \image_{(x, y, t)} = (1 - \interfScalar_{(x, y, t)})\scene_{(x, y, t)} + \interfScalar_{(x, y, t)}\mathbf{1},
\end{equation}
where $\scene_{(x, y, t)} \in \mathbb{R}^3$ and $\interfScalar_{(x, y, t)} \in \mathbb{R}$ are in the range of $[0, 1)$, and $\mathbf{1} = [1, 1, 1]^T$.
In this form, $\interfScalar_{(x, y, t)}$ acts as an alpha map of rain streaks.
Our final training objective is described as
\begin{equation}
    \lossRain = \lossRecon + \lambdaTVFlow\lossTVFlow + \lambdaftwo\lossftwo.
\end{equation}

\paragraph{Experiments.}
\fref{fig:rain} shows a qualitative evaluation on real images in NTURain~\cite{Chen:2018:NTURain}, with moving cars and pedestrians.
We compare state-of-the-art video deraining methods based on optimization~\cite{Jiang:2018:FastDeRain} and deep learning~\cite{Chen:2018:NTURain}.
We take five consecutive images to run our method which clearly removes rain streaks in the scene and is qualitatively competitive with the baselines on real images.
For quantitative evaluation, we must resort to synthetic data and use RainSynLight25~\cite{Liu:2018:RainSyn},  consisting of 25 synthetic sequences of nine images.
\Tref{table:rain} shows results of ours and baseline methods: SE~\cite{Wei:2017:SE}, FastDeRain~\cite{Jiang:2018:FastDeRain}, SpacCNN~\cite{Chen:2018:NTURain}, and FCDN~\cite{Yang:2019:FCDN}.
Though our method does not outperform deep learning-based methods, it achieves a comparable result to optimization-based approaches without supervision and the domain knowledge of deraining.
We expect incorporating more regularization could further improve the performance.
\input{tables/rain}
\input{figures/result_rain}

\subsection{Discussion}
\input{figures/ablation_combined}
\paragraph{Ablation study on loss functions.}
We conducted an ablation study on various loss functions.
In~\fref{fig:ablation_reflection}, we show the decomposed background and reflection layer of different training objectives by removing each loss function.
As can be seen, the background content is reconstructed in the reflection layer when we do not use $\lossftwo$ since $g_{\thetaInterf}$ is unconstrained.
Without $\lossTVFlow$, on the other hand, both signals are reconstructed in the background layer.
In this case, $g_{\thetaScene}$ has enough freedom to learn the mixture of two layers moving differently.
In addition to $\lossftwo$ and $\lossTVFlow$, the exclusion loss $\lossExcl$ further improves the quality by preventing the structure of two layers from being correlated.

\paragraph{Ablation study on $w$.}
\fref{fig:ablation_w} shows an ablation study on $w$ in the occlusion-aware flow-based NIRs using RainSynLight25~\cite{Liu:2018:RainSyn}.
As shown in the red boxes on the output, the method without $w$ produces artifacts around occlusion and disocclusion, which indicates that it is difficult to represent all contents including occlusion in a 2D canonical view.
Our method stores occluded appearance information in the extra dimension $w$, and enables accurate reconstruction.

\paragraph{Can a complex model take place of a simpler model?}
Note that in principle our NIRs using a complex motion model (e.g. occlusion-aware flow) can be generalized to the scenes with a simpler motion.  For layer separation, however, it is beneficial to use a simpler model that fits well with the motion of a scene as it provides a strong constraint to separate layers effectively.
In~\fref{fig:model_experiment}, we compare the homography-based model and the occlusion-aware flow-based model on a demoir\'eing task.
The PSNR and SSIM of the flow-based model on the synthetic test set are 36.72 and 0.9512, respectively.
The flow-based model removes the moir\'e pattern to some extent, but a part of the pattern still remains.
This is because constraining the representation of motion in the homography-based model is more effective than adding regularization losses in the flow-based model.

\paragraph{The number of input images.}
\fref{fig:number_of_images} shows results of rain removal using the different number of input images.
Better results are obtained with more images, as the additional images provide more information in separating two layers.


\paragraph{Limitations.}
Since our method relies on a pixel distance loss to learn the motion, it may fail when the motion of burst images is too large.
Our method also fails to separate layers when the underlying scene and interference move in a similar manner. 
Although our method is not the top performer in all cases, it achieves competitive results without the need for supervision, which is the case for many of the state-of-the-art methods.

In addition, our method requires a proper assumption of motion to tackle layer separation tasks.
This is because our method relies on the motion of underlying scene as a prior to separate layers.
It may be more desirable to seek a generic model that works for any example without the assumption of motion by incorporating other priors such as an inductive bias learned from a large dataset.

Finally, our method currently takes about 30 minutes at most for optimizing layer separation tasks, which is a limiting factor in a real-world setting.  However, there is already promising research demonstrating how to improve the optimization performance of NIRs~\cite{Tancik:2021:InitMeta}.


\input{figures/further_discussion}
\section{Conclusion}
We presented a framework that uses neural image representations to fuse information from multiple images.   The framework simultaneously registers the images and fuses them into a single continuous image representation.  We outlined multiple variations based on the underlying scene motion: homography-based, occlusion-free optical flow, and occlusion-aware optical flow.  Unlike conventional image alignment and fusion, our approach does not need to select one of the input images as a reference frame.  We showed our framework can be used to address layer separation problems using two NIRs, one for the desired scene layer and the other for the interference layer.

Neural image representations are an exciting new approach to image processing.
This work is a first attempt to extend NIRs to multi-frame inputs with applications to various low-level computer vision tasks.
Despite making only minimal assumptions and without leveraging any supervisory training data, the NIR-based approaches described here are competitive with state-of-the-art, unsupervised methods on individual tasks.
Further, because it is practically impossible to acquire supervisory data in real-world conditions, our approach often qualitatively outperforms supervised methods on real-world imagery.

\noindent{\textbf{Acknowledgement.}}~This work was funded in part by the Canada First Research Excellence Fund (CFREF) for the Vision: Science to Applications (VISTA) program and the NSERC Discovery Grant program.

%
%
\bibliographystyle{splncs04}
\bibliography{reference}

\clearpage


\setcounter{table}{0}
\setcounter{figure}{0}
\renewcommand{\thetable}{A\arabic{table}}
\renewcommand{\thefigure}{A\arabic{figure}}

\appendix


\section{Implementation Details}
We describe our implementation in detail for various multi-image layer separation tasks.   
In particular, we describe the settings and hyperparameters that we found to give the best qualitative results.

\paragraph{Moir\'e removal.}
For $g_{\thetaTrans}$, we use an MLP with two hidden layers, 256 hidden units, and a ReLU activation function.
We also initialize the bias of the output layer to the identity transformation of the input coordinates.
For $f^1_{\thetaScene}$, we use a SIREN~\cite{Sitzmann:2020:SIREN} with four hidden layers and 256 hidden units.
For $f^2_{\thetaInterf}$, we use a SIREN with four hidden layers and 128 hidden units.
We set $\lambdaftwo$ and $\lambdaExcl$ to 0.001 and 0.002, respectively.
At training time, we update the networks for 3,000 iterations using an Adam optimizer~\cite{Kingma:2014:Adam} with the learning rate of 0.0001.
We use the $\frac{1}{4}$ of the size of the input image as the batch size.

\paragraph{Obstruction removal.}
We use a SIREN with 4 hidden layers and 256 hidden units for $g_{\thetaTrans}$, $f^1_{\thetaScene}$, and $f^2_{\thetaInterf}$.
We set $\lambdaftwo$, $\lambdaTVFlow$, and $\lambdaExcl$ to 0.1, 0.02, and 0.001, respectively.
We use the same optimizer and learning rate settings as those in the moir\'e removal.
We use 5,000 and the $\frac{1}{32}$ of the size of the input image as the number of iterations and the batch size.

\paragraph{Rain removal.}
We use a SIREN with 5 hidden layers and 256 hidden units for $g_{\thetaTrans}$, $f^1_{\thetaScene}$, and $f^2_{\thetaInterf}$.
We set $\lambdaftwo$ and $\lambdaTVFlow$ to 0.01 and 0.02, respectively.
Similarly, the number of training iterations is set to 5,000, and the batch size is set to $\frac{1}{32}$ of the size of the input image.

\input{figures/analysis_supp}
\section{Additional Experiments}
\subsection{Learning motion in NIRs vs. conventional motion estimation}
In our approach, the motion of a scene is optimized jointly with a layer separation task from scratch.
To show the effectiveness of it, we compare our method with a conventional motion estimation.
Specifically, we use a homography-based NIR in the task of moir\'e removal.
For a baseline, we replace the MLP for estimating homography matrices with a conventional homography estimation method~\cite{Brown:2007:Automatic}, and use the center frame as a reference.
We train both the baseline and the original NIR on the same synthesized dataset.
As a result, the baseline achieves an average PSNR of 23.94 and SSIM of 0.7690, while those of the original NIR are 38.68 and 0.9751.
\fref{fig:analysis_motion} shows a qualitative comparison on real burst images.
As can be seen, our result is visually plausible compared to the baseline result.
Since the input scene is highly corrupted by interference patterns, it is difficult to estimate the motion of the underlying scene accurately using the conventional motion estimation method.
In this case, it is often required to do an additional refinement~\cite{Liu:2020:LearningToSee}.
On the other hand, our method achieves a better performance by jointly learning the motion and layer separation in a single framework of NIR.

\subsection{Further Analysis on $w$.}
In~\fref{fig:analysis_reg_w}, we additionally conduct an experiment to further understand the space represented by $w$.
Specifically, we explicitly enforce our model to represent most of pixel values at $w = 0$ by adding the regularization $\lossW = \sum~\lVert w \rVert_1$.
We also evaluate the performance by training the same model five times and computing the mean and standard deviation of PSNRs.
As can be seen, the model with $\lossW$ uses 0 as the center of $w$ space, while the original model uses an arbitrary value varying according to initialization.
However, the output quality is similar, as shown in both qualitative and quantitative results.
This finding implies that the smoothness prior, driven by $\lossTVFlow$, is more important to learn multi-image representation than the center of $w$.

\subsection{Other Applications}
\input{figures/result_denoising}
\input{figures/result_sr}
\paragraph{Burst image denoising.}
We additionally apply our method to burst image denoising.
To do this, we cast the problem as a layer separation that is to decompose the signal of images into the underlying scene and noise.
In this task, we use an occlusion-free flow-based neural representation as a function of the scene, which is reasonable since the motion of burst images is typically small but may not be planar.
We use the same objective as used in moir\'e removal.
As shown in~\fref{fig:denoising}, we evaluate the performance using a burst image denoising dataset in~\cite{Liu:2014:FBID}.
Our method demonstrates a competitive result compared with a burst image denoising method~\cite{Liu:2014:FBID}, which indicates that our multi-image fusion based on NIRs is also useful to remove random noise signals.

\paragraph{Joint demosaicing and burst super-resolution.}
To further understand the effectiveness of our multi-image fusion, we apply our method to a more challenging task: burst image super-resolution.
Since our method deals with a real-valued coordinate space, it is technically applicable to a sub-pixel registration in burst super-resolution.
For the experiment, we use a real burst dataset in~\cite{Bhat:2021:DeepBurstSR,Bhat:2021:NTIREBurstSR} that contains sequences of 14 raw burst images for testing.
To reconstruct RGB values from Bayer color filter array (CFA) images, we multiply a channel mask to the 3-channel output of a NIR before comparing it to ground truth.
At inference time, we take all channels of RGB output, where missing channels are interpolated.
For implementation, we use an occlusion-free flow-based NIR.
\fref{fig:sr} shows results of a joint demosaicing and burst super-resolution (x4).
The original image is upsampled using bicubic interpolation, and all images are post-processed using the code in~\cite{Bhat:2021:NTIREBurstSR}.
As can be seen, our results show clearer details of images than bicubic results but are still blurry.
We conjecture that the sub-pixel alignment may not be accurate since the optimization only relies on the error of pixel intensities.
Thus, an interesting direction of follow-up research would be to add more loss functions and regularization to assist with sub-pixel registration.
Image noise is also reduced in our results without an explicit noise layer, but using our two-stream architecture would improve the performance.

\subsection{Additional Comparisons}
\Tref{table:obstruction} compares quantitative results of obstruction removal on three controlled sequences in~\cite{Xue:2015:Computational}.
We use the baseline results reported in~\cite{Liu:2020:LearningToSeeJournal}.
Like other tasks, our method (without supervision) achieves comparable results to the unsupervised method, but does not outperform supervised approaches.

\fref{fig:supp_burst} shows burst images used for the layer separation applications addressed in the main paper, and \fref{fig:supp_moire} to \fref{fig:supp_rain} show additional qualitative results.
Particularly, the last example in~\fref{fig:supp_reflection} shows a failure case.
In the example, our method fails to remove the floor of the reflected scene.
Since our layer separation relies on the difference of scene motion, our method does not work well when the movement of the two layers is similar.
Note that we use an occlusion-aware flow-based NIR for the results in~\fref{fig:supp_fence} due to occlusion and disocclusion.
Even though we choose one of three models for each application, it is generally desirable to use the best model to fit the problem's setting.
For the last example in~\fref{fig:supp_rain}, we use synthetic images to show our result on a dynamic scene.

\input{tables/obstruction}
\input{figures/supp_burst}
\input{figures/supp_moire}
\input{figures/supp_reflection}
\input{figures/supp_fence}
\input{figures/supp_rain}


\end{document}

%% file: packages.tex
\usepackage{graphicx}
\usepackage{comment}
\usepackage{amsmath,amssymb} 
\usepackage{color}

\usepackage[accsupp]{axessibility}  

\usepackage[pagebackref=true,breaklinks=true,colorlinks,bookmarks=false]{hyperref}

\usepackage{booktabs}
\usepackage{caption}
\usepackage{subcaption}
\usepackage{cuted}
\usepackage{multirow}
\usepackage[title]{appendix}
\usepackage{wrapfig}

\usepackage{tikz}
\usetikzlibrary{calc,spy}

%% file: macros.tex
\newcommand{\Tref}[1]{Table~\ref{#1}}
\newcommand{\eref}[1]{Eq.~(\ref{#1})}

\newcommand{\fref}[1]{Fig.~\ref{#1}}

\newcommand{\image}{\hat{\mathbf{I}}}
\newcommand{\imageGT}{\mathbf{I}}
\newcommand{\scene}{\hat{\mathbf{O}}}
\newcommand{\interf}{\hat{\mathbf{U}}}
\newcommand{\interfScalar}{\hat{\mathrm{U}}}
\newcommand{\thetaImage}{\theta_{I}}
\newcommand{\thetaTrans}{\theta_{T}}
\newcommand{\thetaScene}{\theta_{O}}
\newcommand{\thetaInterf}{\theta_{U}}
\newcommand{\transform}{T}
\newcommand{\homography}{M}
\newcommand{\loss}[1]{\mathcal{L}_{\mathrm{#1}}}
\newcommand{\lambdaLoss}[1]{\lambda_{\mathrm{#1}}}
\newcommand{\lossTVFlow}{\loss{TVFlow}}
\newcommand{\lossRecon}{\loss{Recon}}
\newcommand{\lossMoire}{\loss{Moire}}
\newcommand{\lossftwo}{\loss{Interf}}
\newcommand{\lossExcl}{\loss{Excl}}
\newcommand{\lossRefl}{\loss{Refl}}
\newcommand{\lossFence}{\loss{Fence}}
\newcommand{\lossRain}{\loss{Rain}}
\newcommand{\lossW}{\loss{w}}
\newcommand{\lambdaftwo}{\lambdaLoss{Interf}}
\newcommand{\lambdaExcl}{\lambdaLoss{Excl}}
\newcommand{\lambdaTVFlow}{\lambdaLoss{TVFlow}}

%% file: figures/teaser.tex
\begin{figure}
\centering
\includegraphics[width=0.97\textwidth]{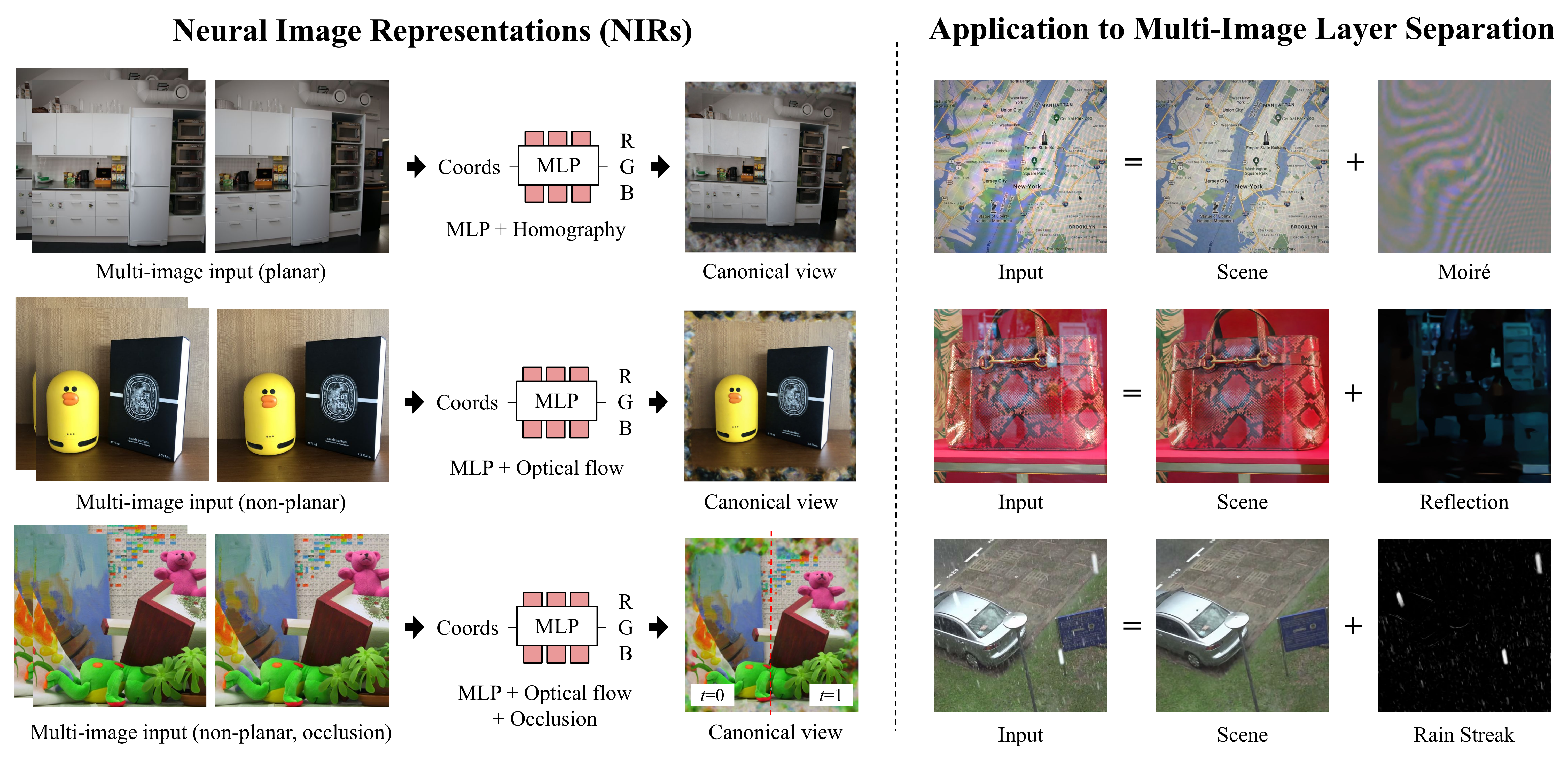}
\caption{This figure provides an overview of our work which fuses multiple images to a single canonical view in a continuous image representation. Our method incorporates motion models such as homography and optical flow into the formulation of implicit or coordinate-based neural representations. We demonstrate the effectiveness of our method on various applications of multi-image layer separation. Images from~\cite{Baker:2011:Middlebury,Li:2013:LiandBrown,Chen:2018:NTURain,Meneghetti:2015:StitchingDataset} are used here for visualization.}
\label{fig:teaser}
\end{figure}

%% file: figures/method_motion.tex
\begin{figure}
\centering
\captionsetup{justification=centering}
\begin{subfigure}[b]{0.32\textwidth}
    \centering
    \includegraphics[width=\textwidth]{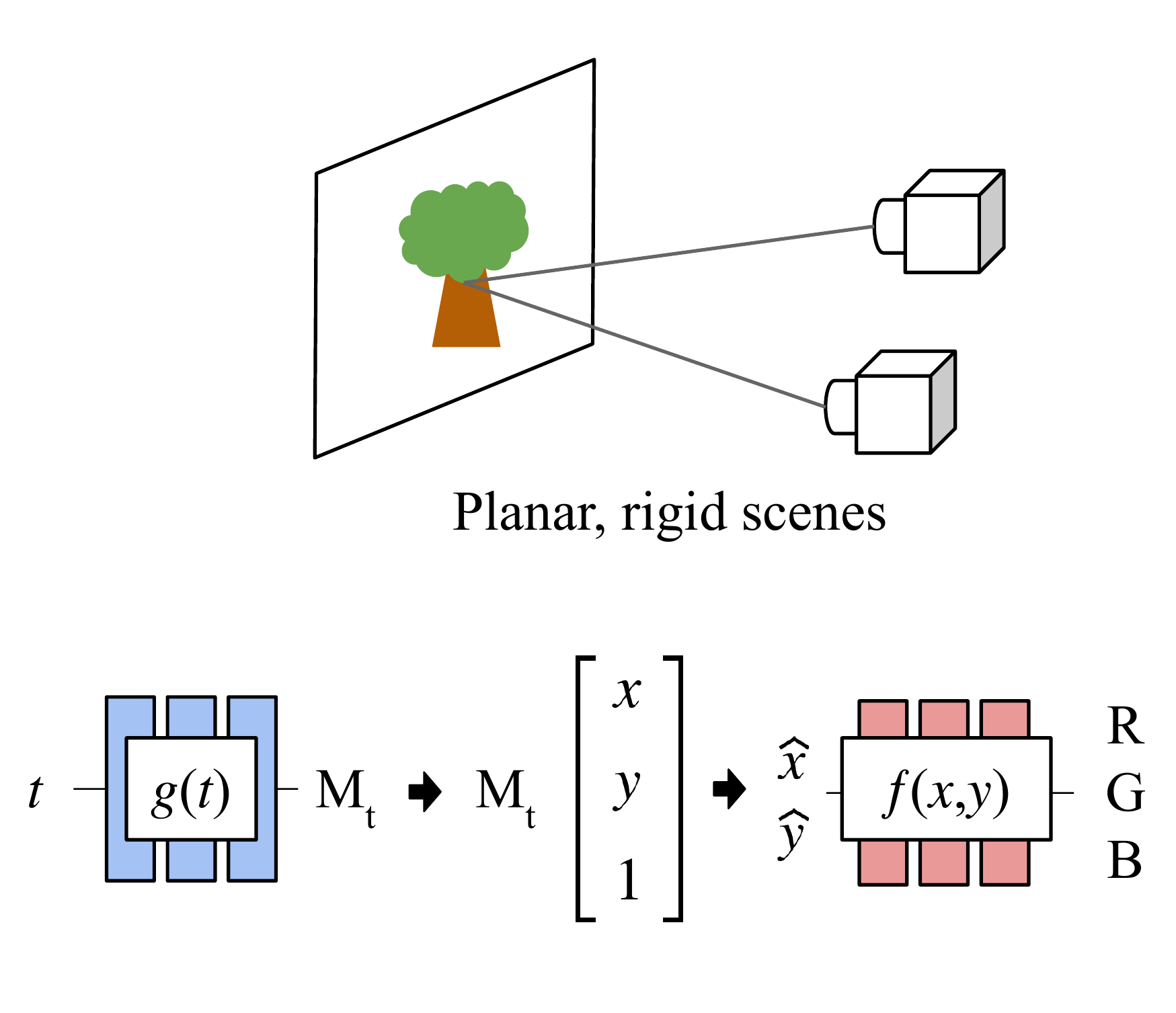}
    \caption{Homography-based NIR}
\end{subfigure}
\begin{subfigure}[b]{0.32\textwidth}
    \centering
    \includegraphics[width=\textwidth]{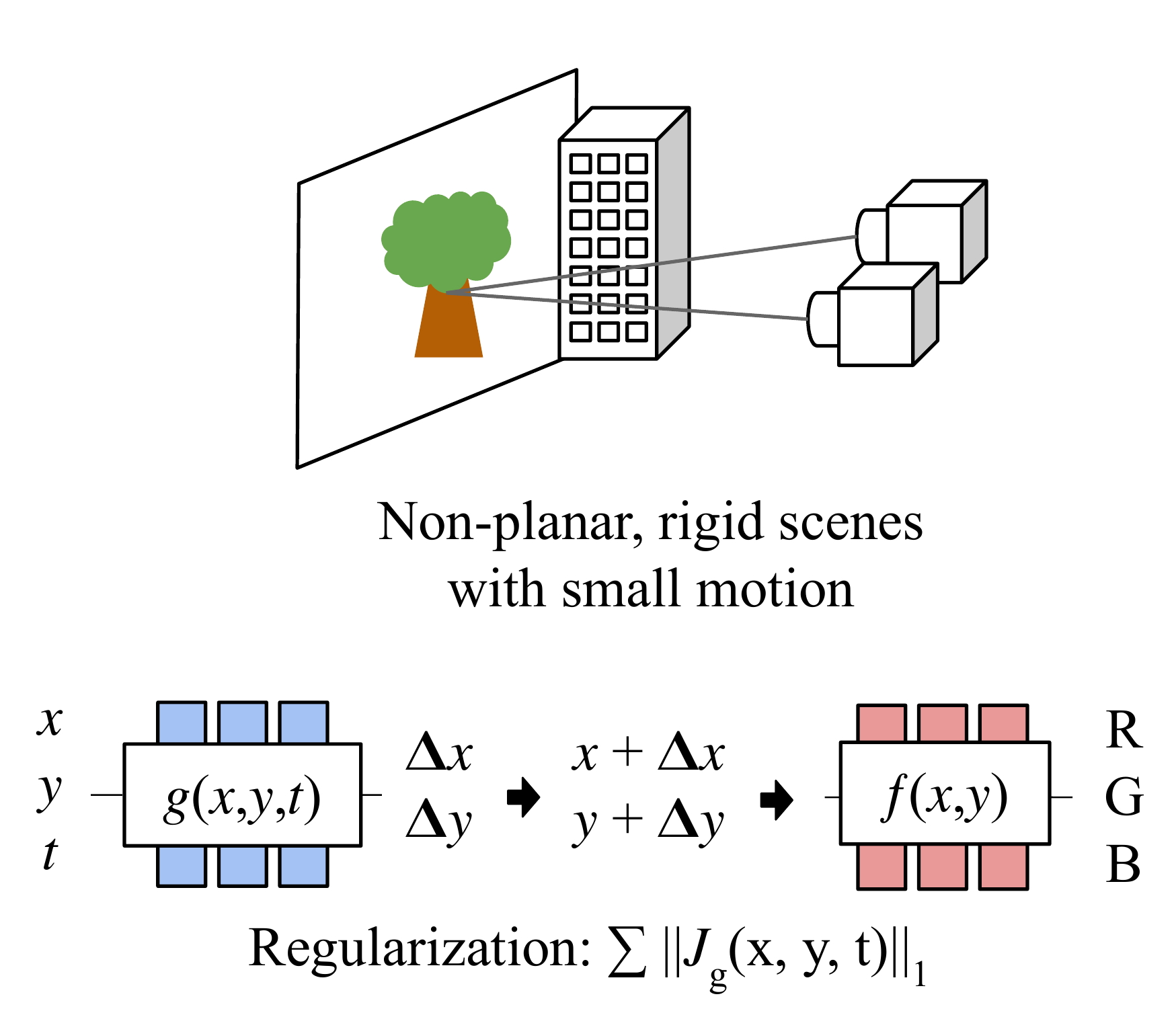}
    \caption{Occlusion-free flow-based NIR}
\end{subfigure}
\begin{subfigure}[b]{0.32\textwidth}
    \centering
    \includegraphics[width=\textwidth]{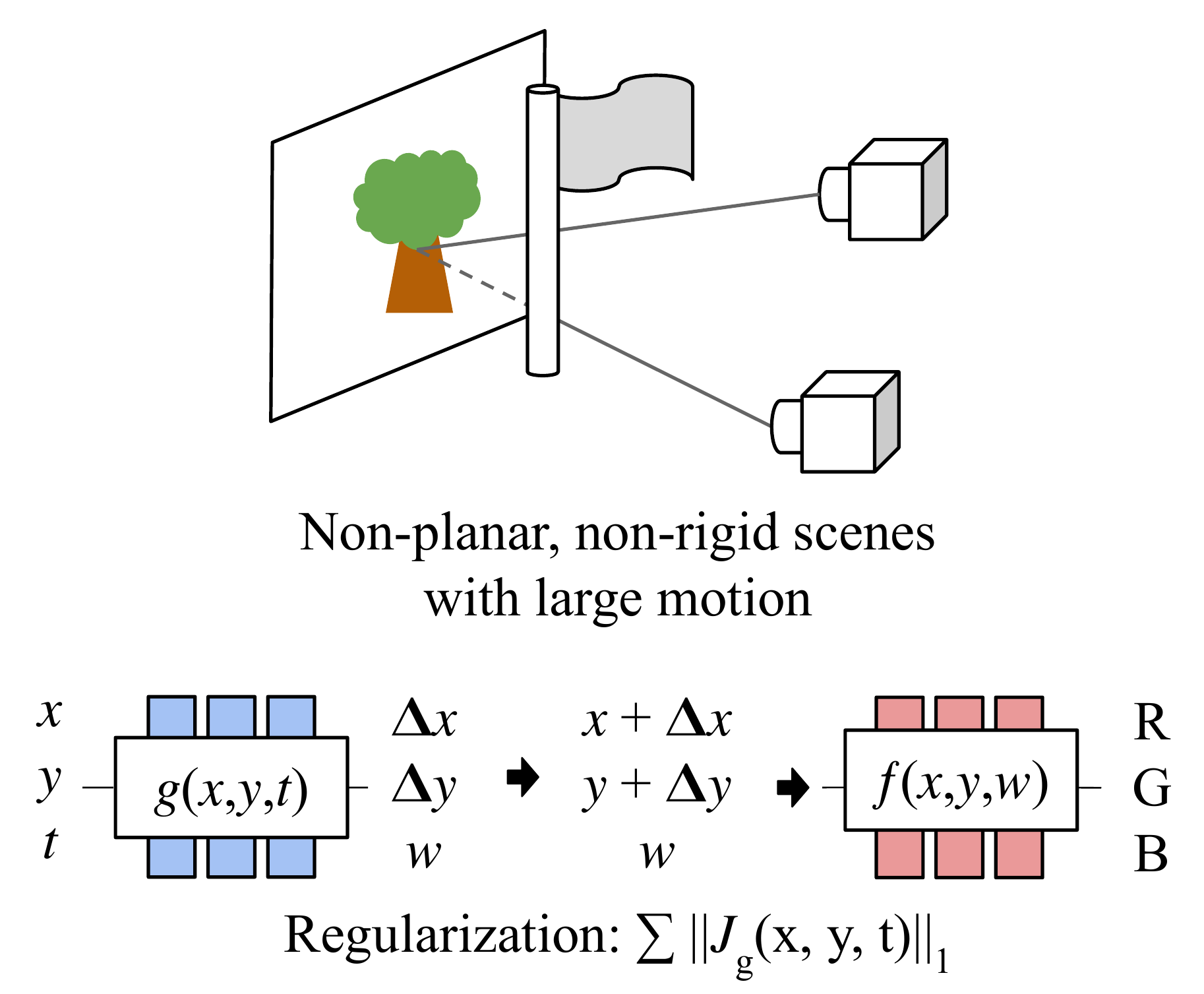}
    \caption{Occlusion-aware flow-based NIR}
\end{subfigure}
\captionsetup{justification=justified}
\caption{Illustration of our neural image representations (NIRs). Assuming that the MLP $f$ learns a canonical view where all burst images are fused, we render each image by projecting the canonical view to the frame-specific view, which is achieved by transforming the input coordinates fed into the $f$. We estimate the transform using another MLP $g$. According to different assumptions of the world, we formulate our framework differently; we formulate the transform of coordinates using (a) homography, (b) optical flow without occlusion/disocclusion, and (c) optical flow with occlusion/disocclusion.}
\label{fig:method_motion}
\end{figure}

%% file: figures/vis_motion.tex
\begin{figure}[t]
\centering
\setlength{\tabcolsep}{1pt}
\begin{subfigure}[b]{0.47\textwidth}
    \centering
    \begin{tabular}{ccc}
    \includegraphics[width=0.3\textwidth]{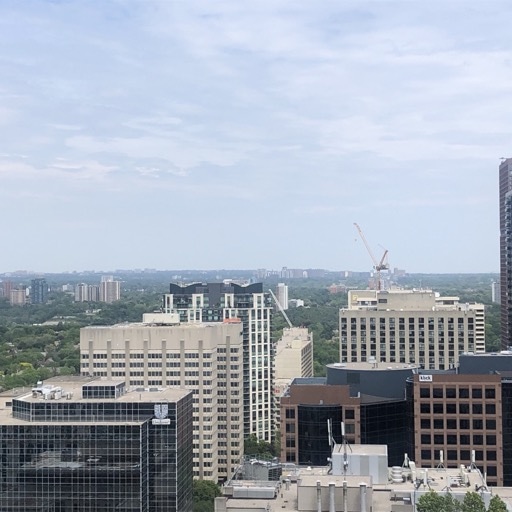} & \includegraphics[width=0.3\textwidth]{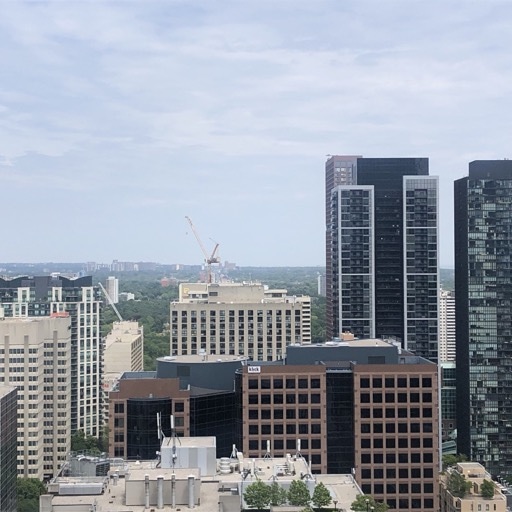} & \includegraphics[width=0.3\textwidth]{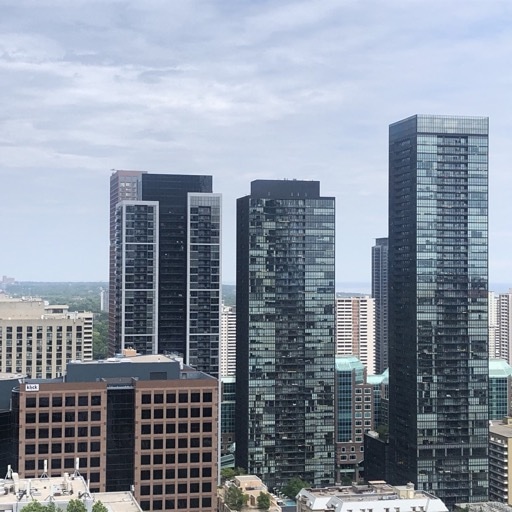} \\
    \multicolumn{3}{c}{\small Input} \\
    \multicolumn{3}{c}{\includegraphics[width=0.915\textwidth]{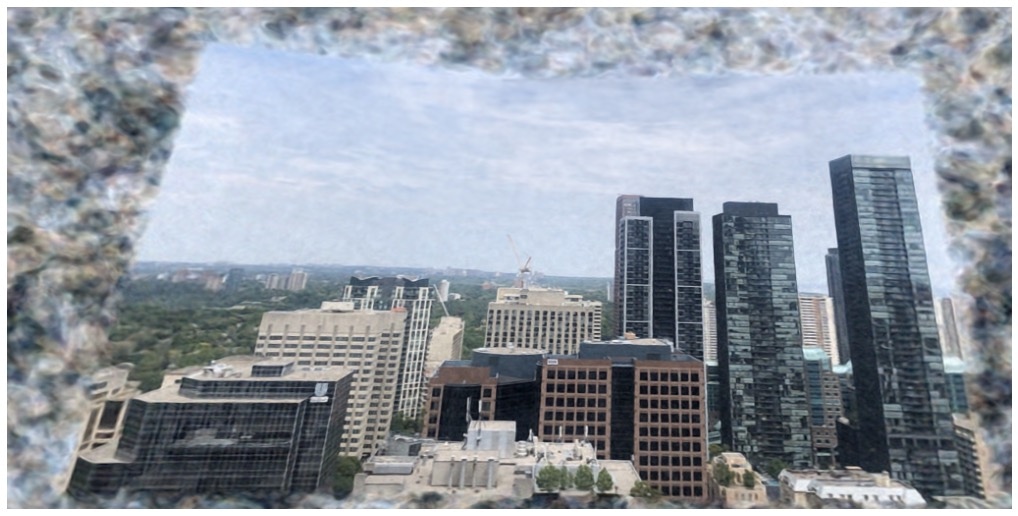}} \\
    \multicolumn{3}{c}{\small Canonical view} \\
    \end{tabular}
    \caption{Homography-based NIR}
\end{subfigure}
\hfill
\begin{subfigure}[b]{0.47\textwidth}
    \centering
    \begin{tabular}{cccccc}
    \multicolumn{3}{c}{\includegraphics[trim=30 20 30 40,clip,width=0.45\textwidth]{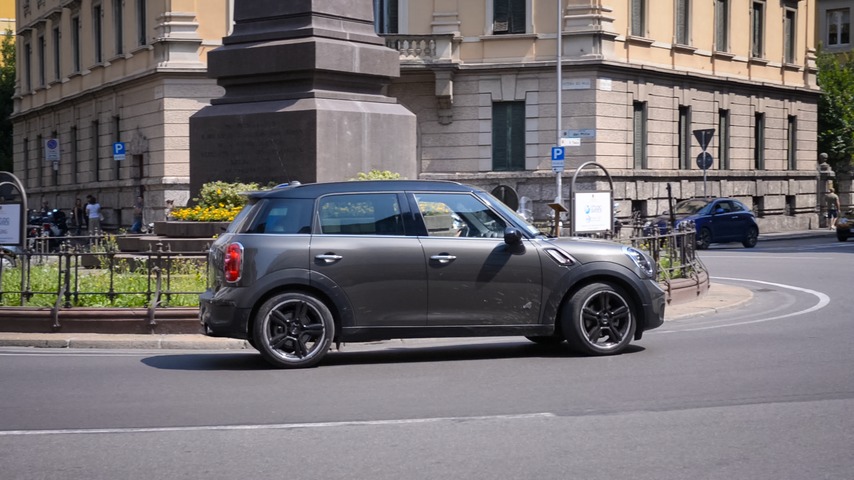}} & \multicolumn{3}{c}{\includegraphics[trim=100 70 100 130,clip,width=0.45\textwidth]{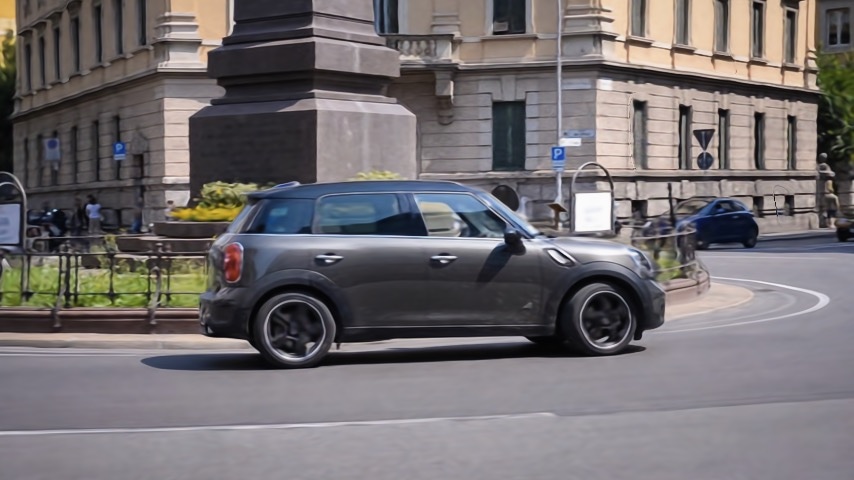}} \\
    \multicolumn{3}{c}{\small Input} & \multicolumn{3}{c}{\small Reconstruction} \\
    \multicolumn{3}{c}{\includegraphics[trim=100 70 100 130,clip,width=0.45\textwidth]{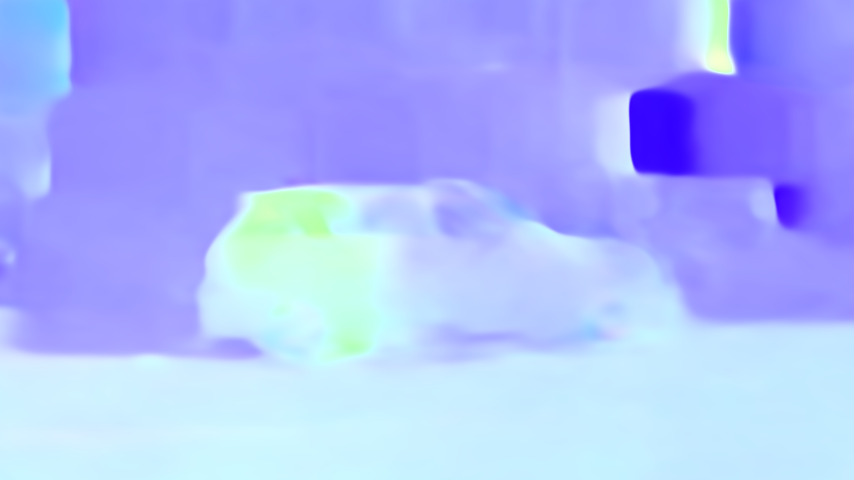}} & \multicolumn{2}{c}{\includegraphics[trim=100 70 245 130,clip,width=0.35\textwidth]{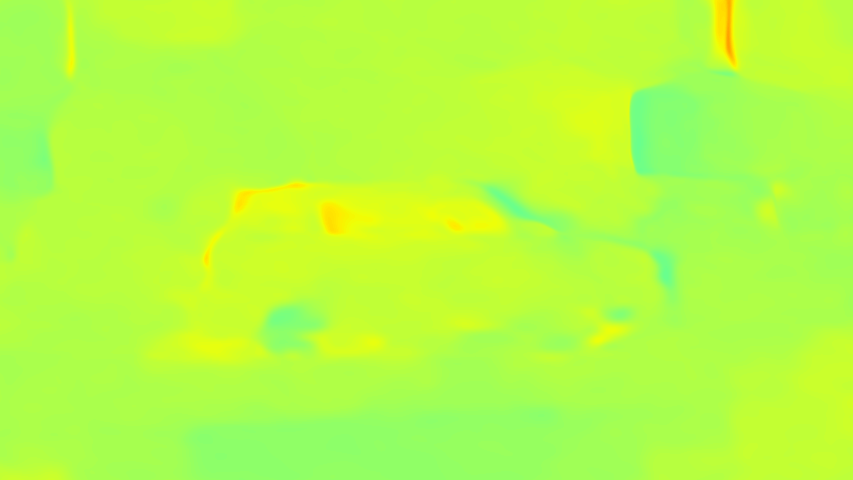}} & \includegraphics[width=0.1\textwidth]{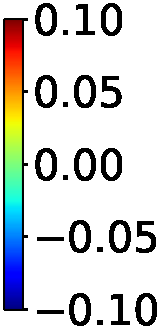} \\
    \multicolumn{3}{c}{\small $xy$-flow map} & \multicolumn{3}{c}{\small $w$ map} \\
    \multicolumn{2}{c}{\includegraphics[trim=150 80 400 110,clip,width=0.3\textwidth]{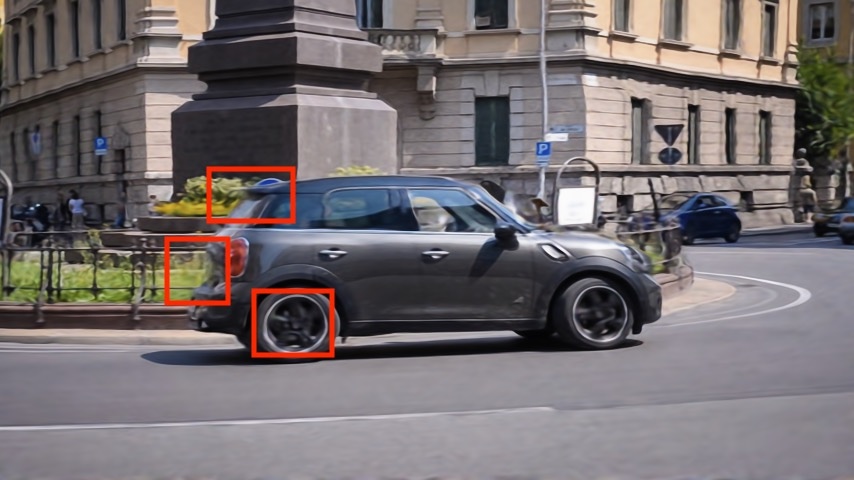}} & \multicolumn{2}{c}{\includegraphics[trim=150 80 400 110,clip,width=0.3\textwidth]{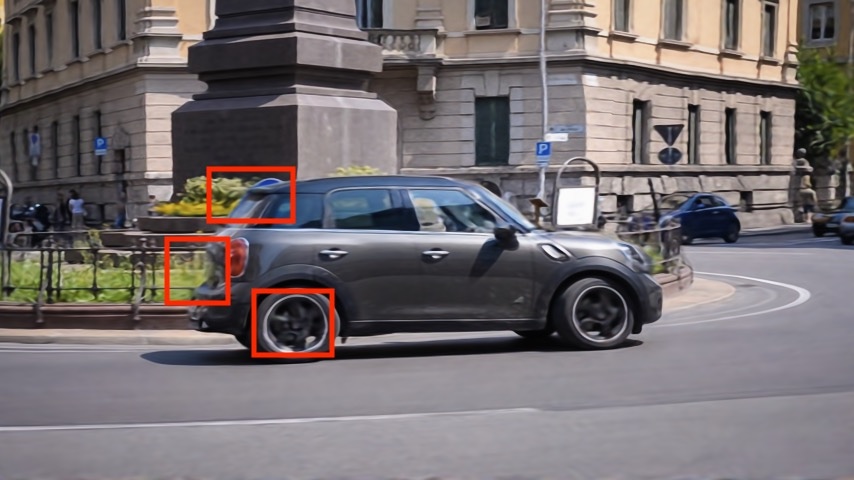}} & \multicolumn{2}{c}{\includegraphics[trim=150 80 400 110,clip,width=0.3\textwidth]{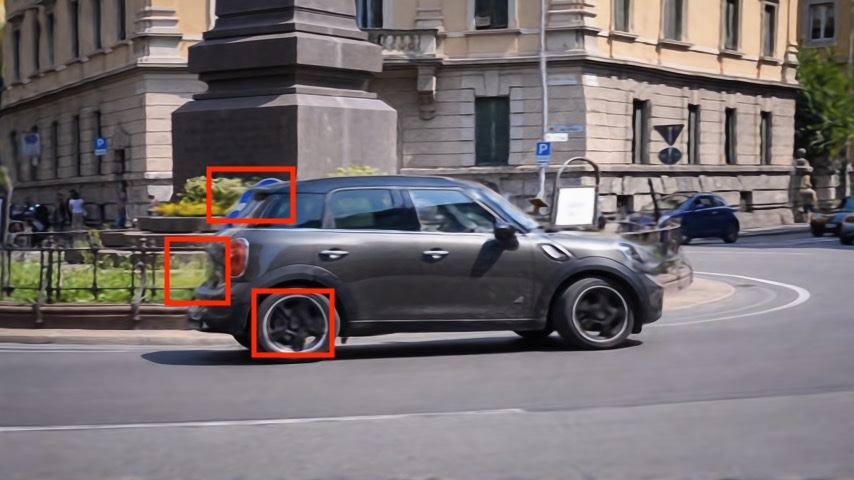}} \\
    \multicolumn{6}{c}{\small Canonical views at $t = 0, 2, 3$}
    \end{tabular}
    \caption{Occlusion-aware flow-based NIR}
\end{subfigure}
\caption{Visualization of learned representations. In (a), the top row shows three of nine representative images used to learn a homography-based NIR, and the bottom shows a learned canonical view. In (b), the first row shows one of the input and reconstruction images, the second row shows a $xy$-flow map and $w$ map learned by a occlusion-aware flow-based NIR, and the third row shows canonical views at $t = 0, 2, 3$.}
\label{fig:vis_motion}
\end{figure}

%% file: figures/method_twostream.tex
\begin{wrapfigure}{r}{.5\linewidth}
  \vspace{-22pt}
  \centering
\includegraphics[width=0.48\textwidth]{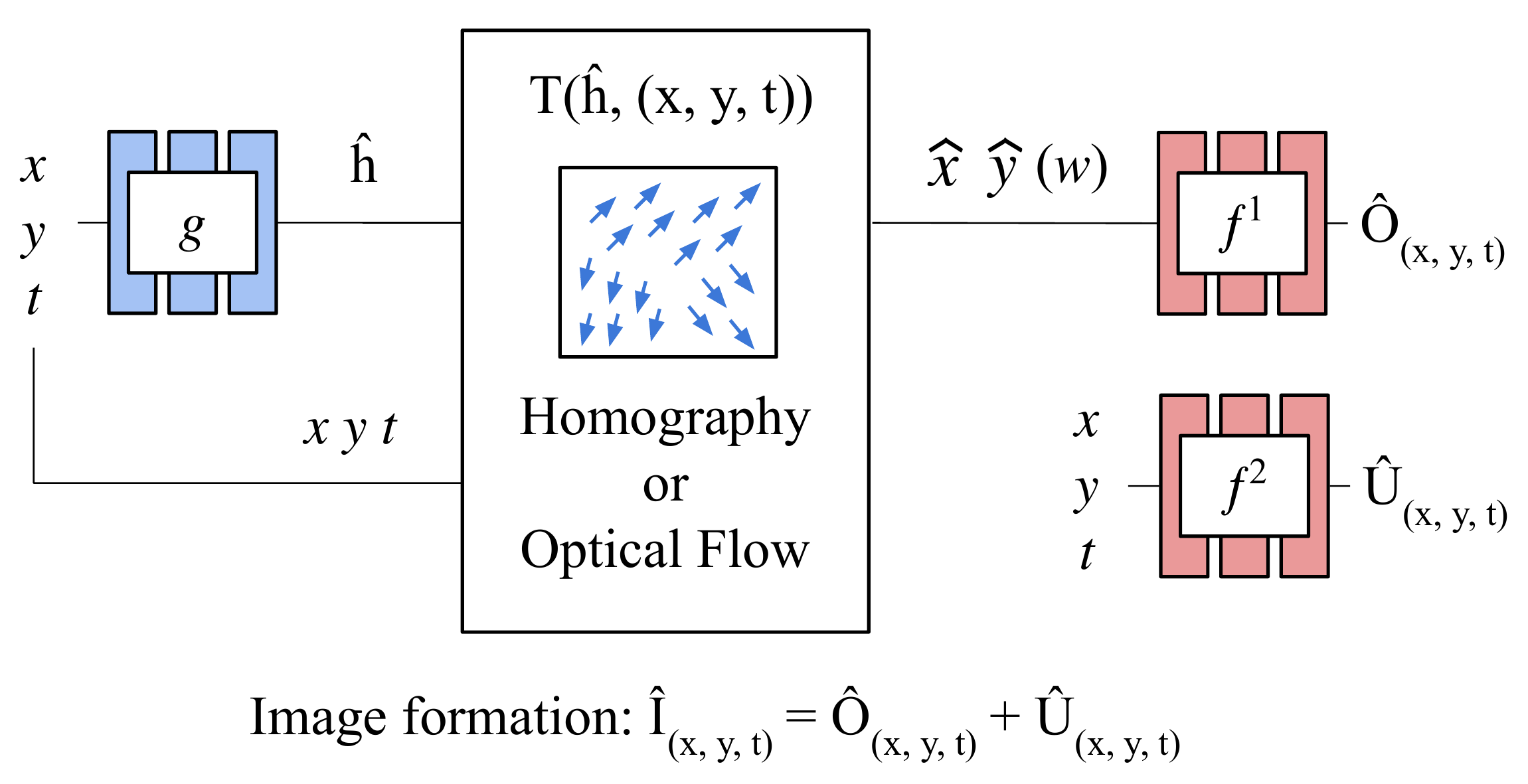}
\caption{Multi-image layer separation.}
\label{fig:method_twostream}
  \vspace{-30pt}
\end{wrapfigure}

%% file: tables/moire.tex
\begin{table}[t]
\begin{center}
\setlength{\tabcolsep}{10pt}
\begin{tabular}{ccccc}
\toprule
& \multicolumn{2}{c}{Supervised} & \multicolumn{2}{c}{Unsupervised} \\
& AFN~\cite{Xu:2020:AFN} & C3Net~\cite{Kim:2020:C3Net} & Double DIP~\cite{Gandelsman:2019:DoubleDIP} & Ours \\
\hline
Input & Single & Burst & Burst & Burst \\
\hline
PSNR & 43.63 & 27.99 & 18.53 & 38.68 \\
SSIM & 0.9952 & 0.8071 & 0.8762 & 0.9751 \\
NCC & 0.9963 & 0.7724 & 0.5120 & 0.9865 \\
SI & 0.9962 & 0.7721 & 0.4895 & 0.9856 \\
\bottomrule
\end{tabular}
\end{center}
\caption{Quantitative evaluation of moir\'e removal with a synthetic dataset. AFN~\cite{Xu:2020:AFN} uses a single image; other methods use five images as input.}
\label{table:moire}
\vspace{-7.0pt}
\end{table}


%% file: figures/result_moire.tex
\tikzstyle{closeup} = [
  opacity=1.0,          
  height=0.7cm,         
  width=0.178\textwidth, 
  connect spies, red  
]

\begin{figure*}[t]
\centering
\begin{tikzpicture}[x=0.19\textwidth, y=0.17\textheight, spy using outlines={every spy on node/.append style={smallwindow}}]
\node[anchor=south] (Fig1A) at (0,0) {\includegraphics[width=0.18\textwidth,angle=-90]{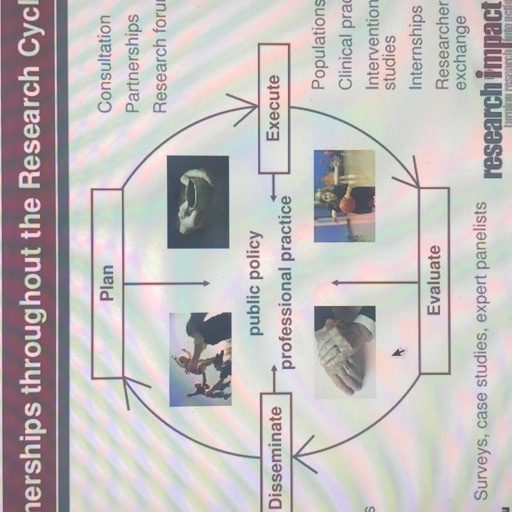}};
\spy [closeup,magnification=2] on ($(Fig1A)+(-0.15,+0.15)$) 
    in node[largewindow,anchor=north west] at ($(Fig1A.south west) + (0.06,0)$);

\node[anchor=south] (Fig1B) at (1,0) {\includegraphics[width=0.18\textwidth,angle=-90]{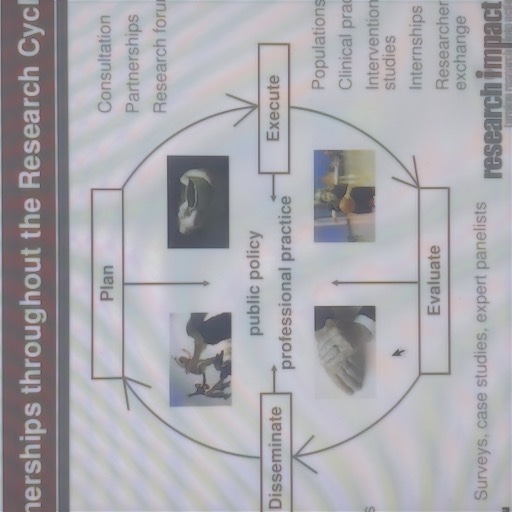}};
\spy [closeup,magnification=2] on ($(Fig1B)+(-0.15,+0.15)$) 
    in node[largewindow,anchor=north west] at ($(Fig1B.south west) + (0.06,0)$);

\node[anchor=south] (Fig1C) at (2,0) {\includegraphics[width=0.18\textwidth,angle=-90]{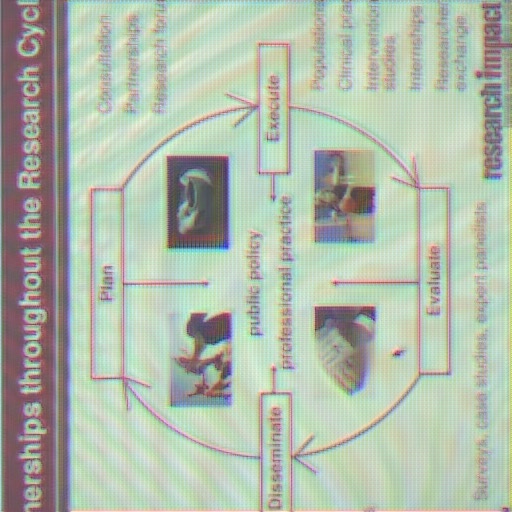}};
\spy [closeup,magnification=2] on ($(Fig1C)+(-0.15,+0.15)$) 
    in node[largewindow,anchor=north west] at ($(Fig1C.south west) + (0.06,0)$);
    
\node[anchor=south] (Fig1D) at (3,0) {\includegraphics[width=0.18\textwidth,angle=-90]{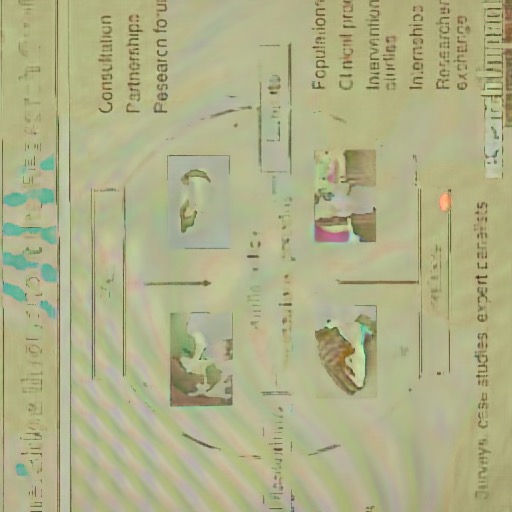}};
\spy [closeup,magnification=2] on ($(Fig1D)+(-0.15,+0.15)$) 
    in node[largewindow,anchor=north west] at ($(Fig1D.south west) + (0.06,0)$);
    
\node[anchor=south] (Fig1E) at (4,0) {\includegraphics[width=0.18\textwidth,angle=-90]{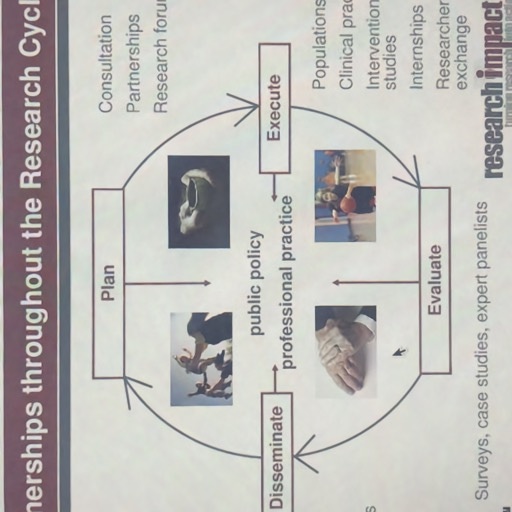}};
\spy [closeup,magnification=2] on ($(Fig1E)+(-0.15,+0.15)$) 
    in node[largewindow,anchor=north west] at ($(Fig1E.south west) + (0.06,0)$);
    
\node[anchor=south] (Fig2A) at (0,-1) {\includegraphics[width=0.18\textwidth,angle=-90]{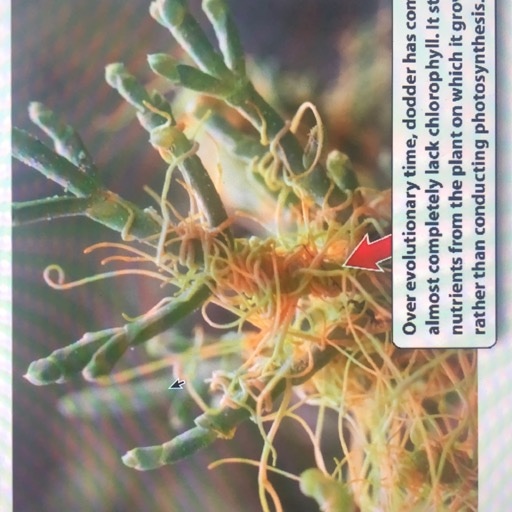}};
\spy [closeup,magnification=2] on ($(Fig2A)+(+0.23,+0.23)$) 
    in node[largewindow,anchor=north west] at ($(Fig2A.south west) + (0.06,0)$);
\node [anchor=north] at ($(Fig2A.south)+(0,-0.23)$) {\small Input};
    
\node[anchor=south] (Fig2B) at (1,-1) {\includegraphics[width=0.18\textwidth,angle=-90]{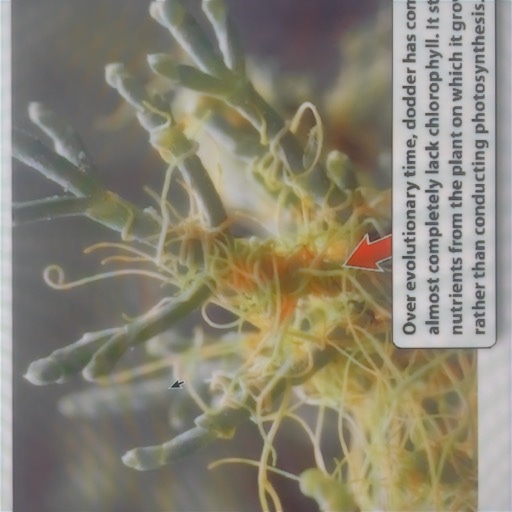}};
\spy [closeup,magnification=2] on ($(Fig2B)+(+0.23,+0.23)$) 
    in node[largewindow,anchor=north west] at ($(Fig2B.south west) + (0.06,0)$);
\node [anchor=north] at ($(Fig2B.south)+(0,-0.23)$) {\small AFN~\cite{Xu:2020:AFN}};
    
\node[anchor=south] (Fig2C) at (2,-1) {\includegraphics[width=0.18\textwidth,angle=-90]{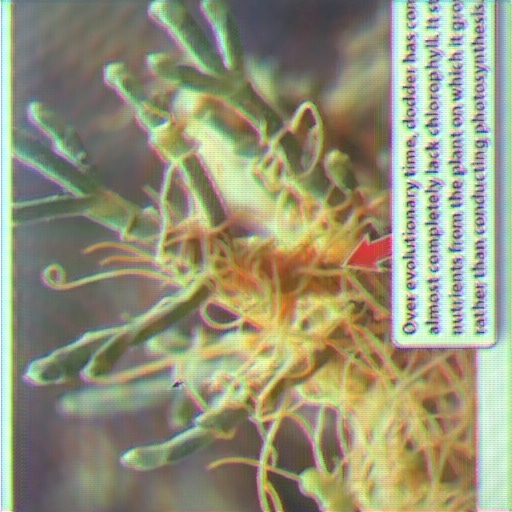}};
\spy [closeup,magnification=2] on ($(Fig2C)+(+0.23,+0.23)$) 
    in node[largewindow,anchor=north west] at ($(Fig2C.south west) + (0.06,0)$);
\node [anchor=north] at ($(Fig2C.south)+(0,-0.23)$) {\small C3Net~\cite{Kim:2020:C3Net}};
    
\node[anchor=south] (Fig2D) at (3,-1) {\includegraphics[width=0.18\textwidth,angle=-90]{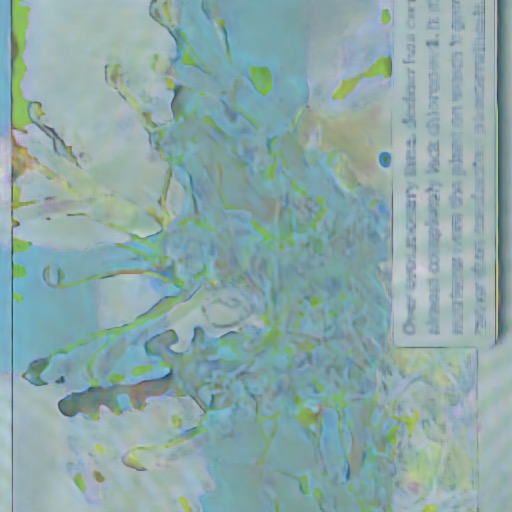}};
\spy [closeup,magnification=2] on ($(Fig2D)+(+0.23,+0.23)$) 
    in node[largewindow,anchor=north west] at ($(Fig2D.south west) + (0.06,0)$);
\node [anchor=north] at ($(Fig2D.south)+(0,-0.23)$) {\small Double DIP~\cite{Gandelsman:2019:DoubleDIP}};
    
\node[anchor=south] (Fig2E) at (4,-1) {\includegraphics[width=0.18\textwidth,angle=-90]{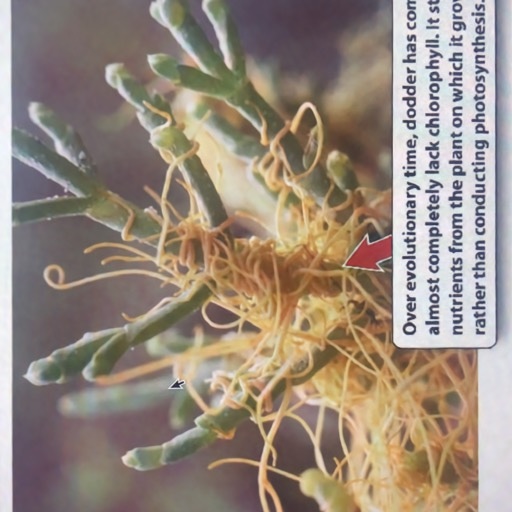}};
\spy [closeup,magnification=2] on ($(Fig2E)+(+0.23,+0.23)$) 
    in node[largewindow,anchor=north west] at ($(Fig2E.south west) + (0.06,0)$);
\node [anchor=north] at ($(Fig2E.south)+(0,-0.23)$) {\small Ours};
\end{tikzpicture}

\caption{Qualitative comparison of moir\'e removal on real images. Our method outperforms all methods including AFN~\cite{Xu:2020:AFN}. AFN was better than ours on the synthetic data in~\Tref{table:moire} which is unrepresentative of real-world images.}
\label{fig:moire}
\end{figure*}
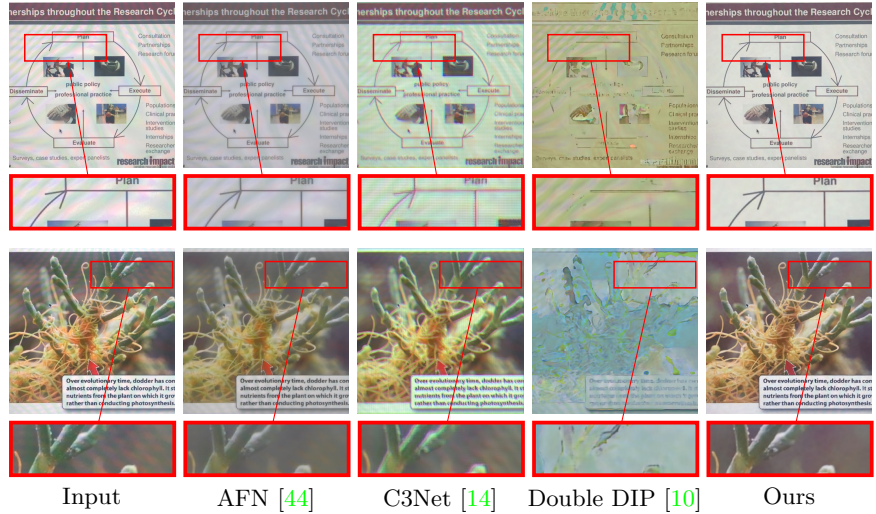

%% file: figures/result_reflection.tex
\begin{figure*}[t]
    \centering
    \setlength{\tabcolsep}{1.1pt}
    \begin{tabular}{cccccc}
    \includegraphics[width=0.16\textwidth]{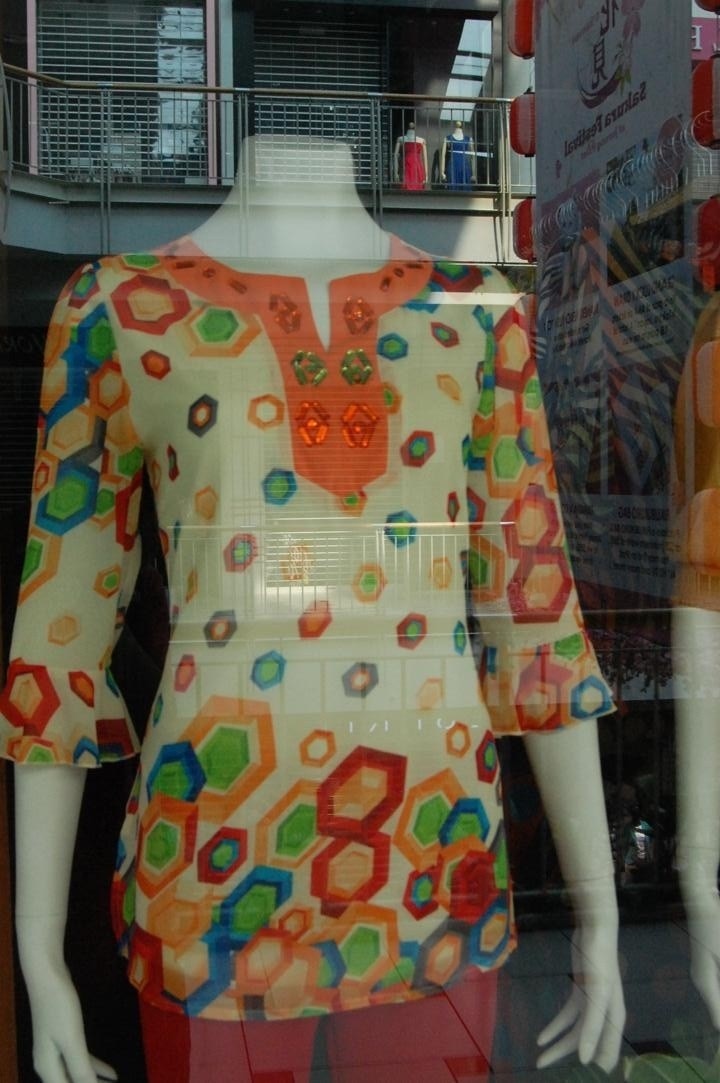} & \includegraphics[width=0.16\textwidth]{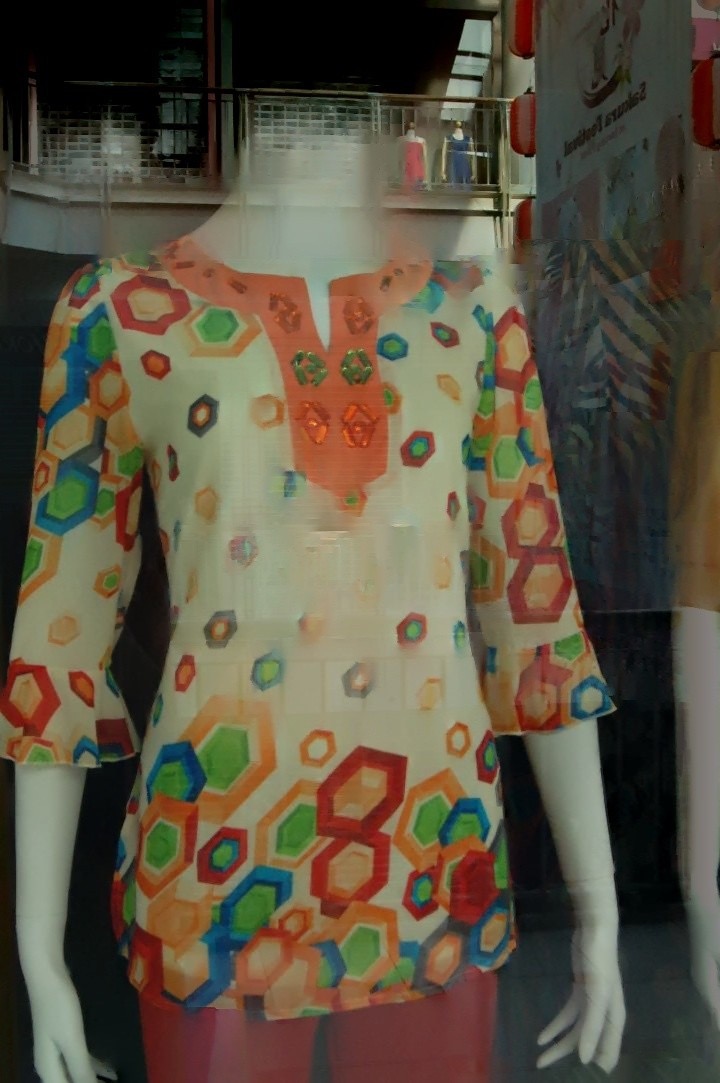} & \includegraphics[width=0.16\textwidth]{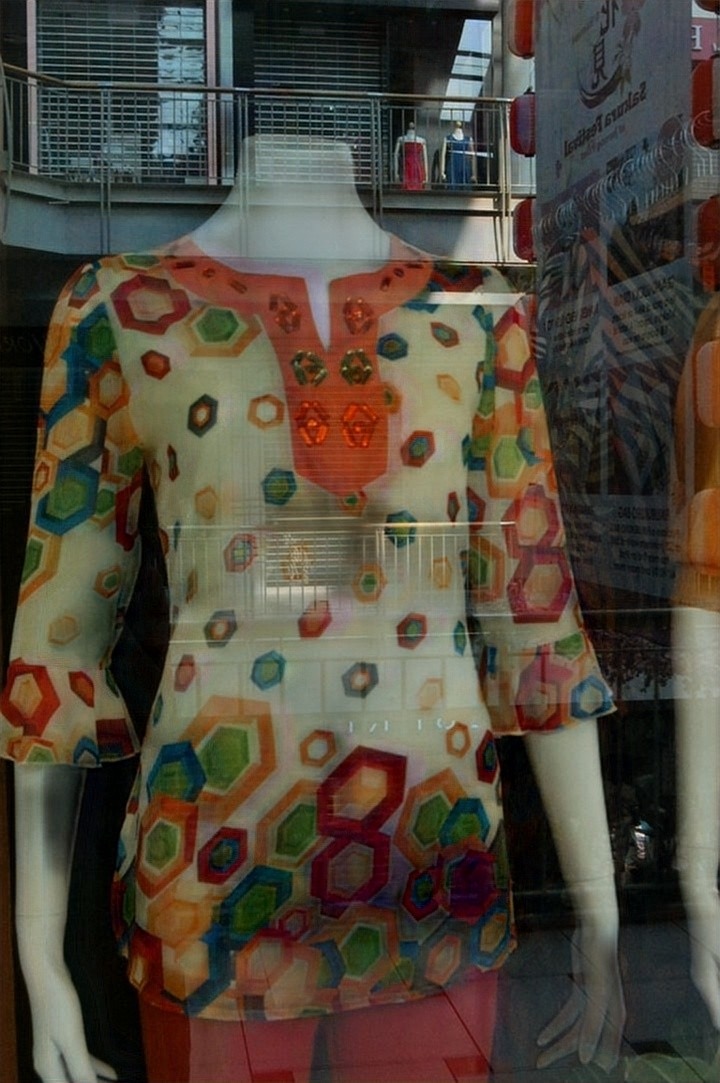} & \includegraphics[width=0.16\textwidth]{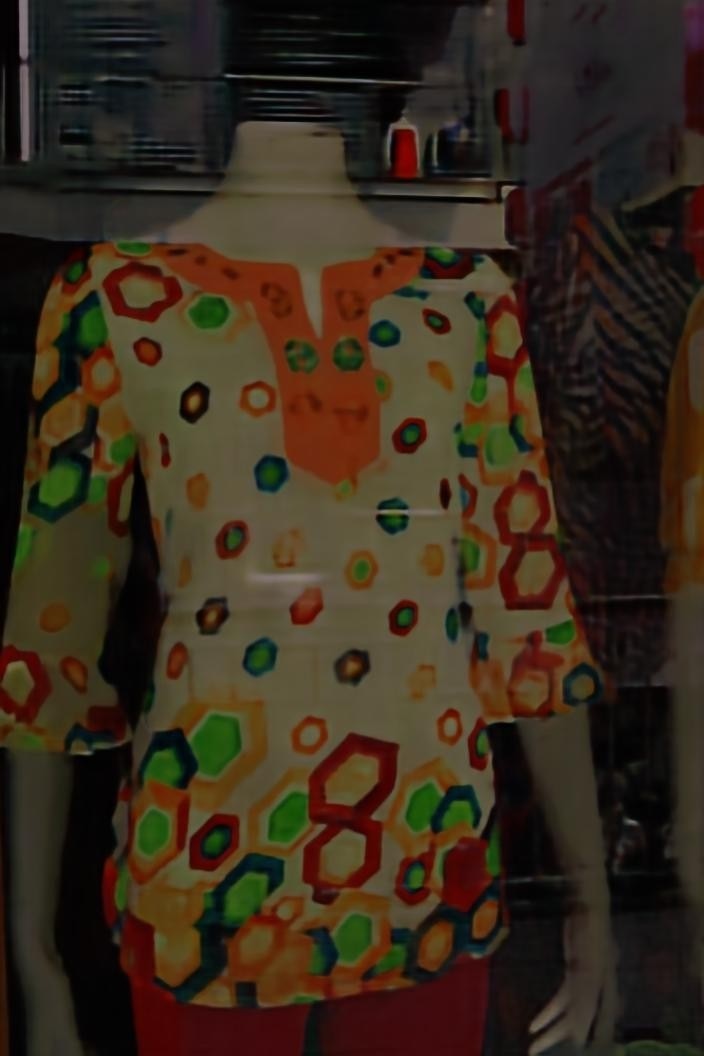} & \includegraphics[width=0.16\textwidth]{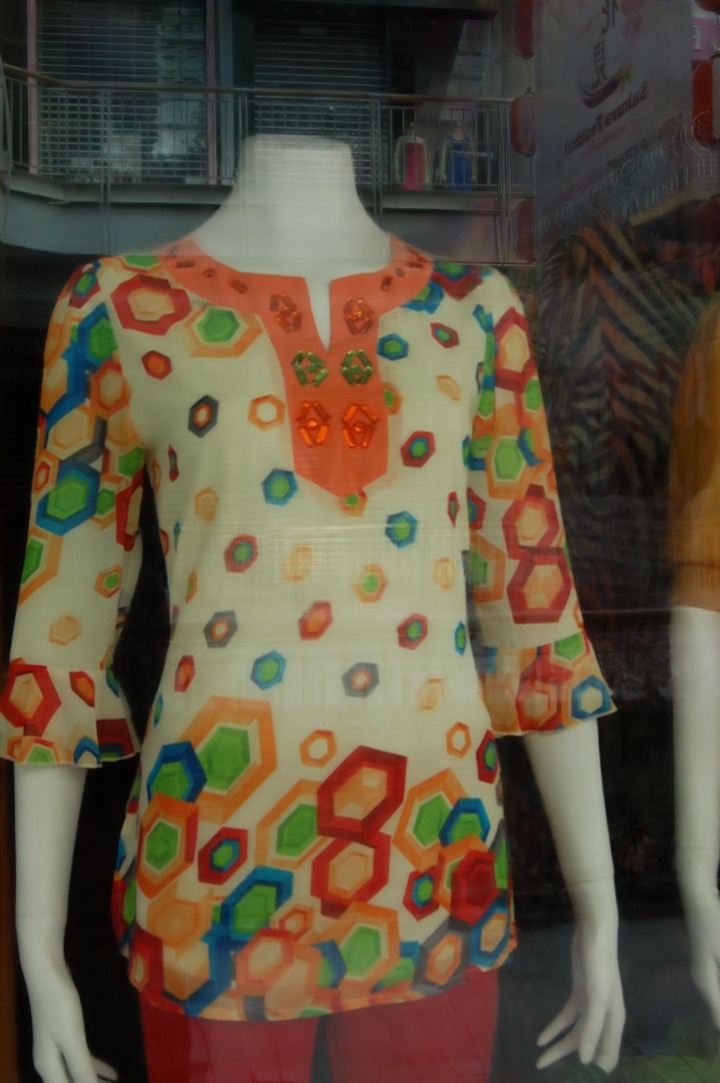} & \includegraphics[width=0.16\textwidth]{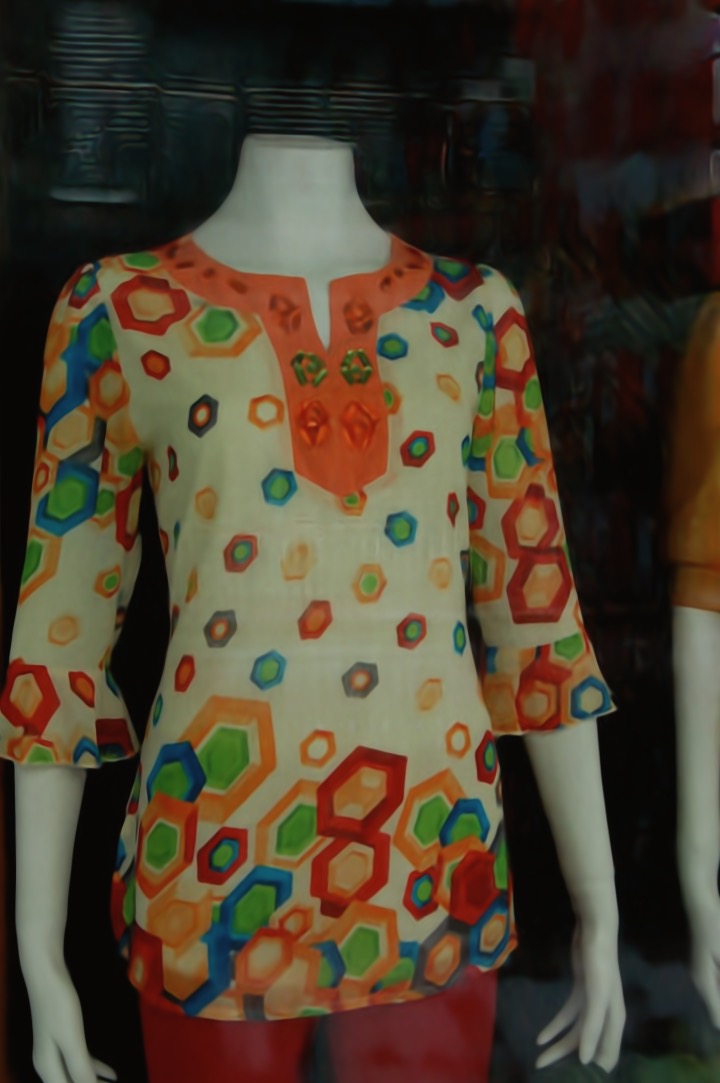} \\
    \includegraphics[width=0.16\textwidth]{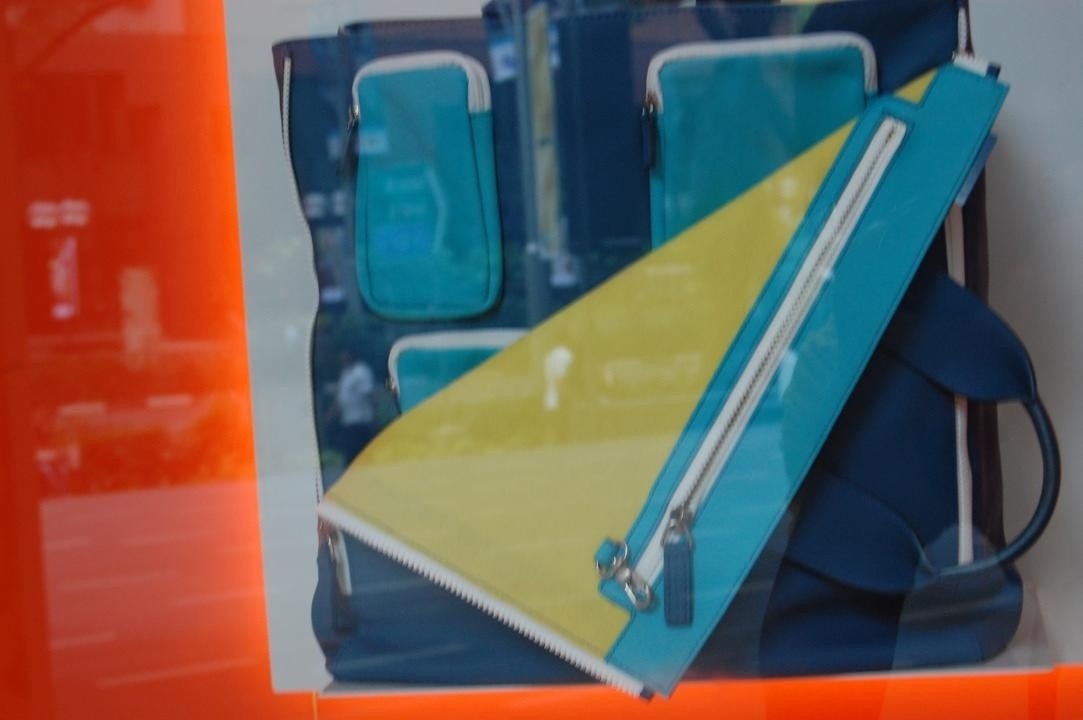} & \includegraphics[width=0.16\textwidth]{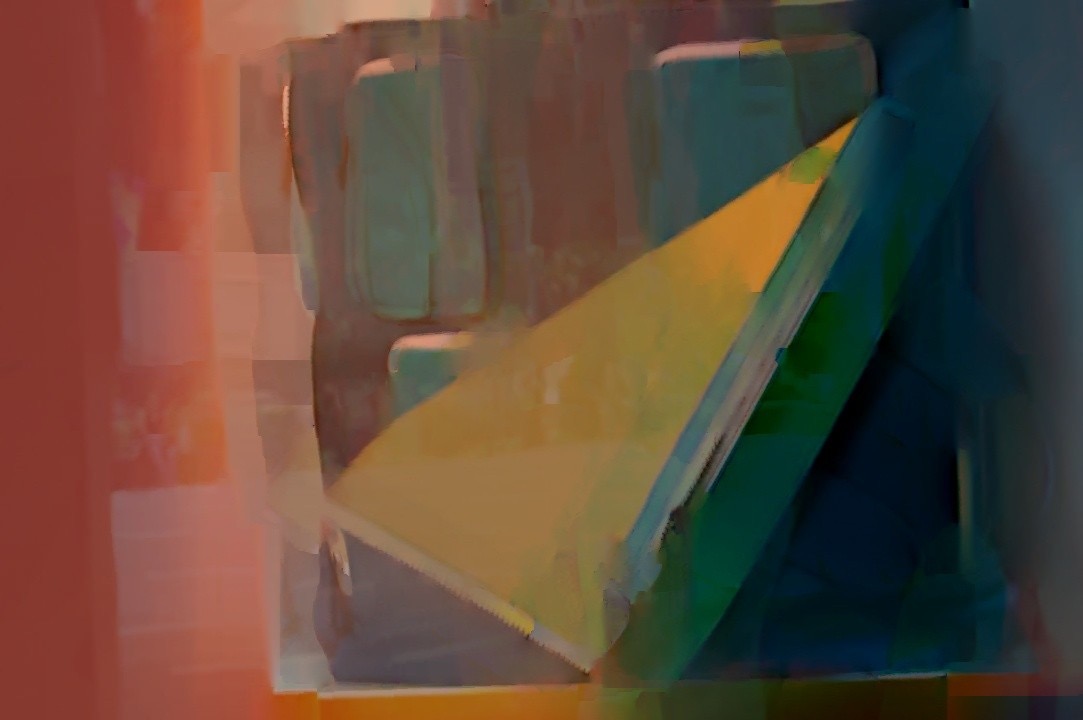} & \includegraphics[width=0.16\textwidth]{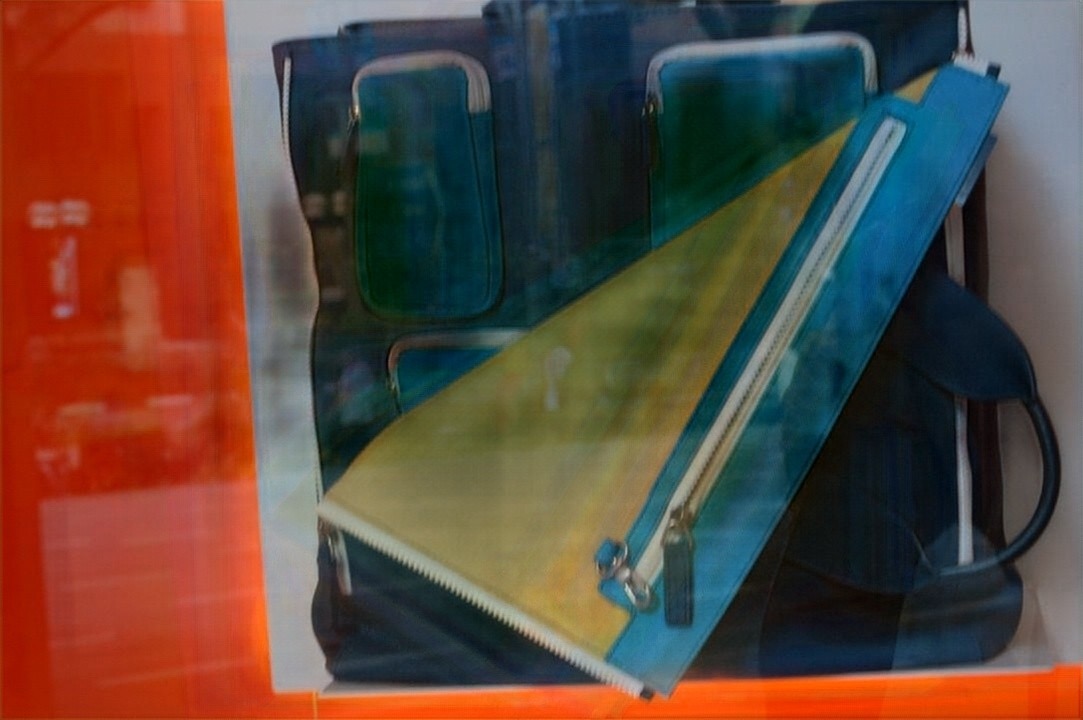} & \includegraphics[width=0.16\textwidth]{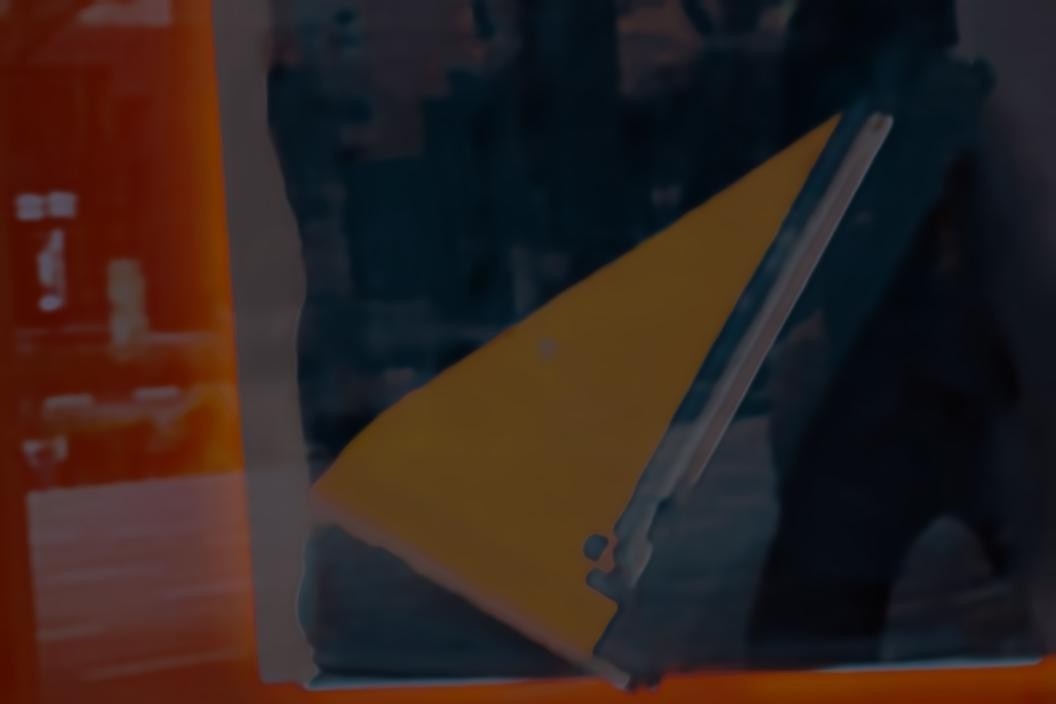} & \includegraphics[width=0.16\textwidth]{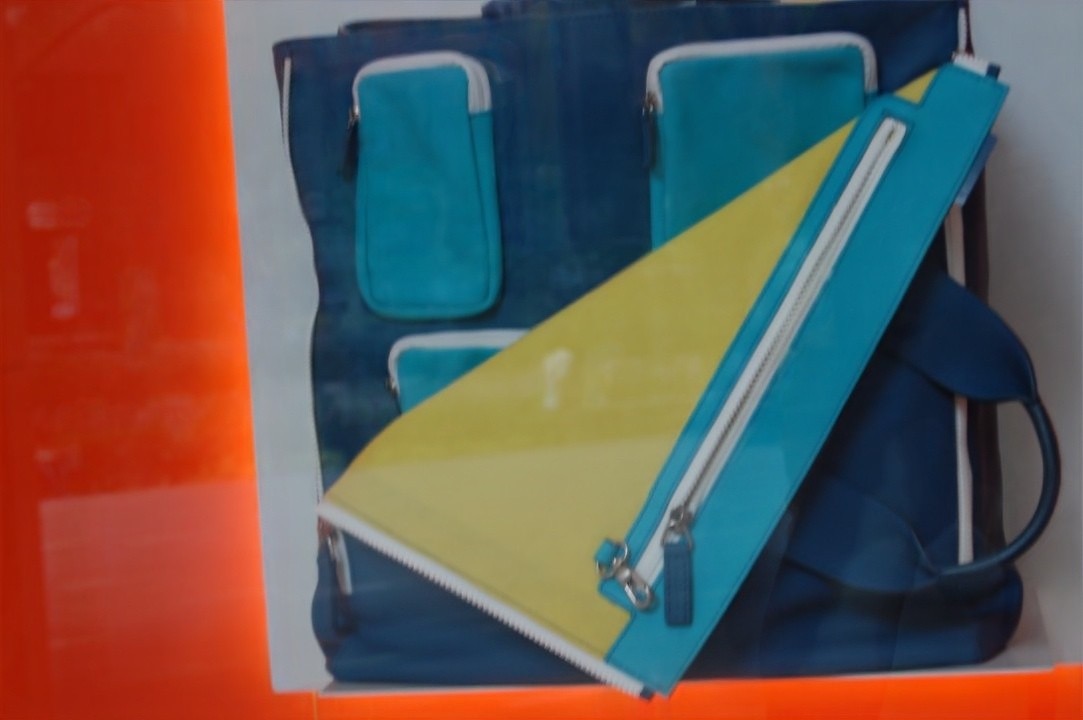} & \includegraphics[width=0.16\textwidth]{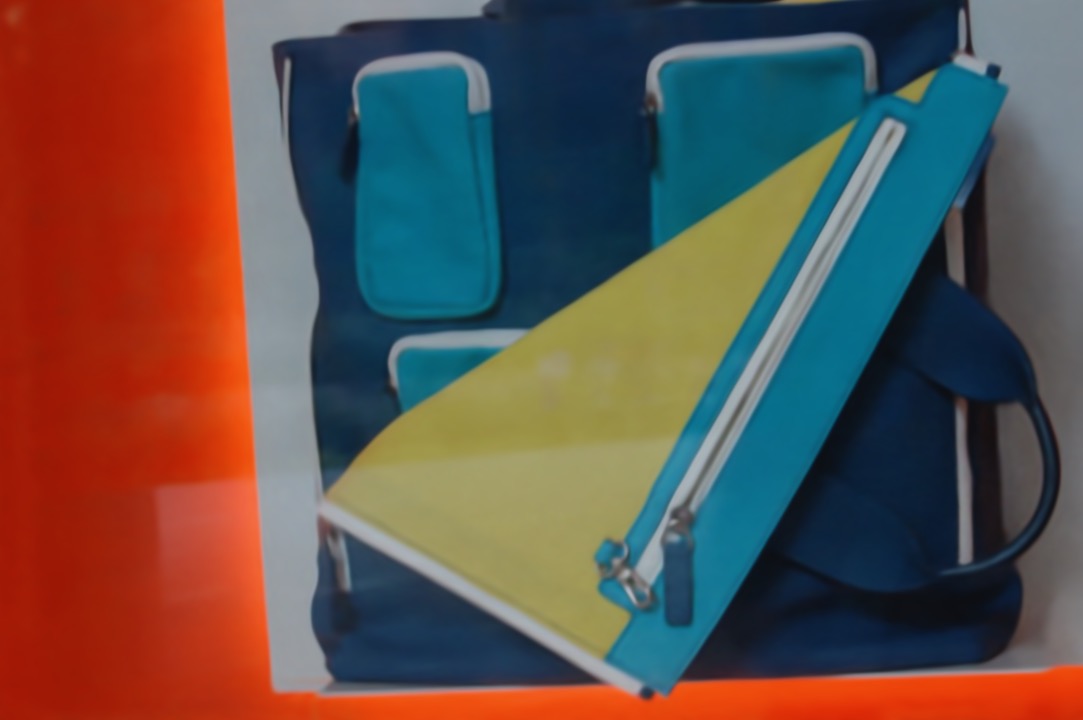} \\
    \multirow{2}{*}{\small Input} & {\small Li and Brown} & {\small Alayrac~\etal} & {\small Double DIP} & {\small Liu~\etal} & \multirow{2}{*}{\small Ours} \\
    & {\small \cite{Li:2013:LiandBrown}} & {\small \cite{Alayrac:2019:VisualCentrifuge}} & {\small \cite{Gandelsman:2019:DoubleDIP}} & {\small \cite{Liu:2020:LearningToSeeJournal}} &
    \end{tabular}
    \caption{Qualitative results of reflection removal on real images in~\cite{Li:2013:LiandBrown}.}
    \label{fig:reflection}
\end{figure*}

%% file: figures/result_fence.tex
\tikzstyle{closeup_fence} = [
  opacity=1.0,          
  height=0.7cm,         
  width=0.23\textwidth, 
  connect spies, red  
]

\begin{figure*}[t]
\centering
\begin{tikzpicture}[x=0.25\textwidth, y=0.2\textheight, spy using outlines={every spy on node/.append style={smallwindow}}]
\node[anchor=south] (Fig1A) at (0,0) {\includegraphics[width=0.235\textwidth]{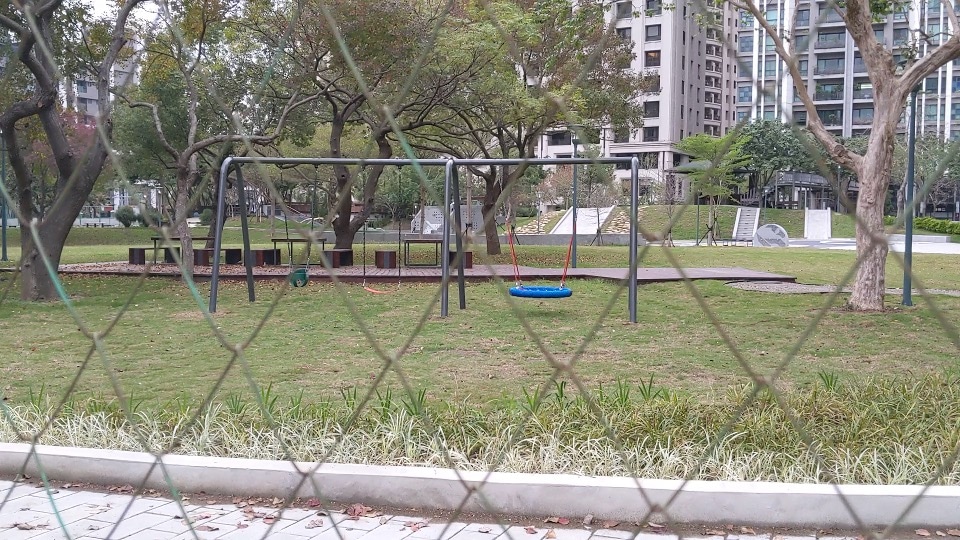}};
\spy [closeup_fence,magnification=2] on ($(Fig1A)+(-0.15,-0.15)$) 
    in node[largewindow,anchor=north west] at ($(Fig1A.south west) + (0.05,0)$);
\node [anchor=north] at ($(Fig1A.south)+(0,-0.2)$) {\small Input};

\node[anchor=south] (Fig1B) at (1,0) {\includegraphics[width=0.235\textwidth]{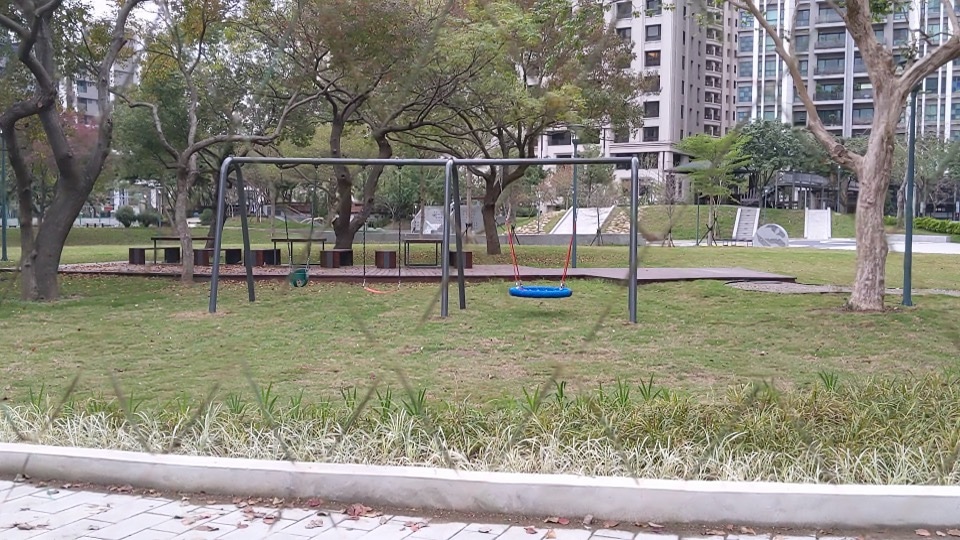}};
\spy [closeup_fence,magnification=2] on ($(Fig1B)+(-0.15,-0.15)$) 
    in node[largewindow,anchor=north west] at ($(Fig1B.south west) + (0.05,0)$);
\node [anchor=north] at ($(Fig1B.south)+(0,-0.2)$) {\small Liu~\etal~\cite{Liu:2020:LearningToSee}};

\node[anchor=south] (Fig1C) at (2,0) {\includegraphics[width=0.235\textwidth]{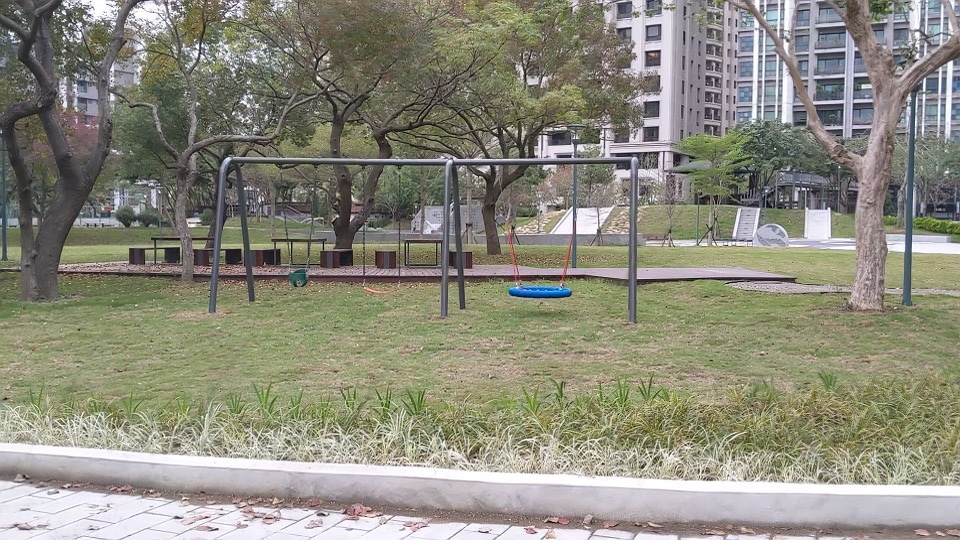}};
\spy [closeup_fence,magnification=2] on ($(Fig1C)+(-0.15,-0.15)$) 
    in node[largewindow,anchor=north west] at ($(Fig1C.south west) + (0.05,0)$);
\node [anchor=north] at ($(Fig1C.south)+(0,-0.2)$) {\small Liu~\etal~\cite{Liu:2020:LearningToSeeJournal}};
    
\node[anchor=south] (Fig1D) at (3,0) {\includegraphics[width=0.235\textwidth]{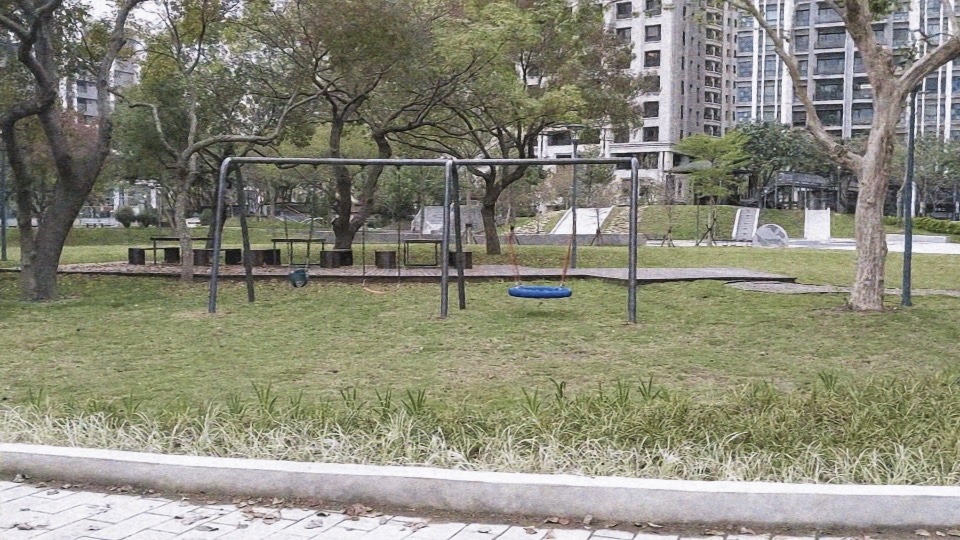}};
\spy [closeup_fence,magnification=2] on ($(Fig1D)+(-0.15,-0.15)$) 
    in node[largewindow,anchor=north west] at ($(Fig1D.south west) + (0.05,0)$);
\node [anchor=north] at ($(Fig1D.south)+(0,-0.2)$) {\small Ours};
\end{tikzpicture}
\caption{Qualitative comparison of fence removal on real images in~\cite{Liu:2020:LearningToSeeJournal}.}
\label{fig:fence}
\end{figure*}

%% file: tables/rain.tex
\begin{table}[t]
\setlength{\tabcolsep}{10pt}
\begin{center}
\begin{tabular}{c|cc|ccc}
\toprule
& \multicolumn{2}{c|}{Supervised} & \multicolumn{3}{c}{Unsupervised} \\
\hline
& SpacCNN & FCDN & SE & FastDeRain & Ours \\
& \cite{Chen:2018:NTURain} & \cite{Yang:2019:FCDN} & \cite{Wei:2017:SE} & \cite{Jiang:2018:FastDeRain} & \\
\hline
PSNR & 32.78 & 35.80 & 26.56 & 29.42 & 28.61 \\
SSIM & 0.9239 & 0.9622 & 0.8006 & 0.8683 & 0.8604 \\
\bottomrule
\end{tabular}
\end{center}
\caption{Result of rain removal on RainSynLight25~\cite{Liu:2018:RainSyn}.}
\label{table:rain}
\vspace{-10.0pt}
\end{table}

%% file: figures/result_rain.tex
\tikzstyle{closeup_rain} = [
  opacity=1.0,          
  height=0.7cm,         
  width=0.226\textwidth, 
  connect spies, red  
]

\begin{figure*}[t]
\centering
\begin{tikzpicture}[x=0.24\textwidth, y=0.16\textheight, spy using outlines={every spy on node/.append style={smallwindow}}]
\node[anchor=south] (Fig1A) at (0,0) {\includegraphics[width=0.23\textwidth]{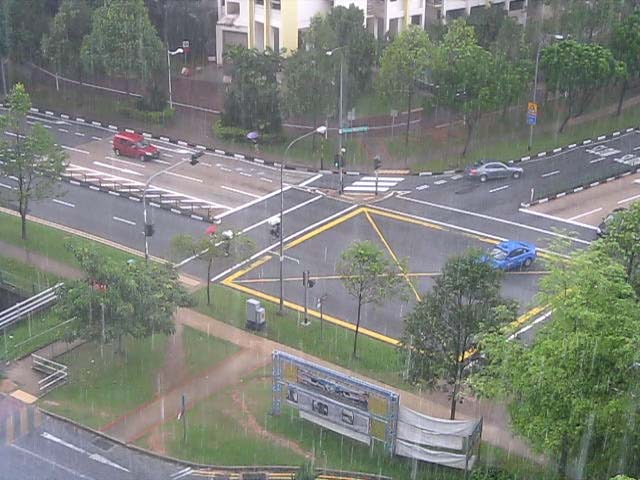}};
\spy [closeup_rain,magnification=3] on ($(Fig1A)+(-0.25,-0.1)$) 
    in node[largewindow,anchor=north west] at ($(Fig1A.south west) + (0.04,0)$);

\node[anchor=south] (Fig1B) at (1,0) {\includegraphics[width=0.23\textwidth]{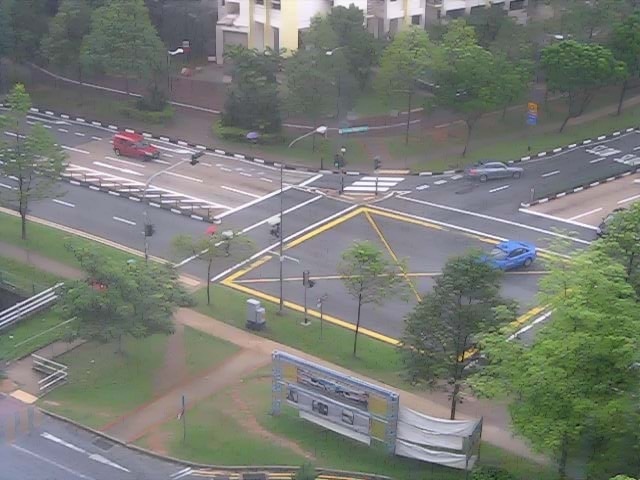}};
\spy [closeup_rain,magnification=3] on ($(Fig1B)+(-0.25,-0.1)$) 
    in node[largewindow,anchor=north west] at ($(Fig1B.south west) + (0.04,0)$);

\node[anchor=south] (Fig1C) at (2,0) {\includegraphics[width=0.23\textwidth]{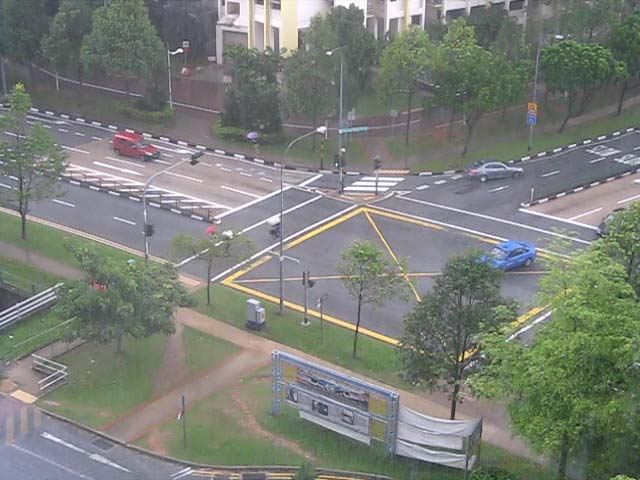}};
\spy [closeup_rain,magnification=3] on ($(Fig1C)+(-0.25,-0.1)$) 
    in node[largewindow,anchor=north west] at ($(Fig1C.south west) + (0.04,0)$);
    
\node[anchor=south] (Fig1D) at (3,0) {\includegraphics[width=0.23\textwidth]{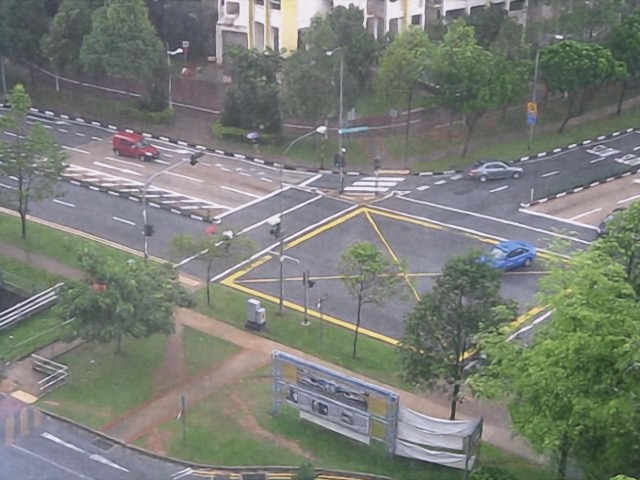}};
\spy [closeup_rain,magnification=3] on ($(Fig1D)+(-0.25,-0.1)$) 
    in node[largewindow,anchor=north west] at ($(Fig1D.south west) + (0.04,0)$);

\node[anchor=south] (Fig2A) at (0,-1) {\includegraphics[width=0.23\textwidth]{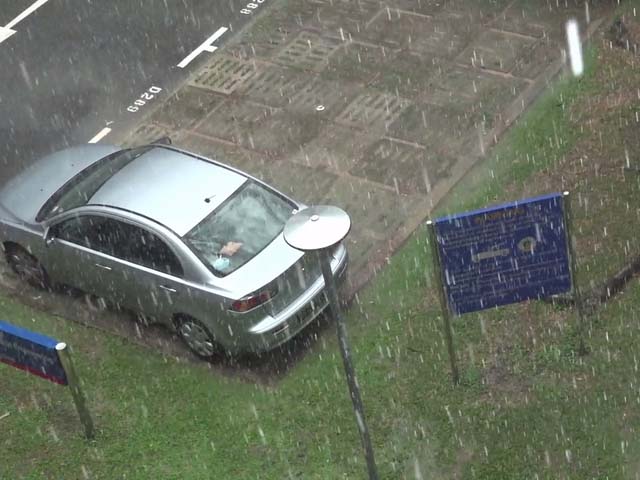}};
\spy [closeup_rain,magnification=3] on ($(Fig2A)+(-0.25,-0.1)$) 
    in node[largewindow,anchor=north west] at ($(Fig2A.south west) + (0.04,0)$);
\node [anchor=north] at ($(Fig2A.south)+(0,-0.23)$) {\small Input};

\node[anchor=south] (Fig2B) at (1,-1) {\includegraphics[width=0.23\textwidth]{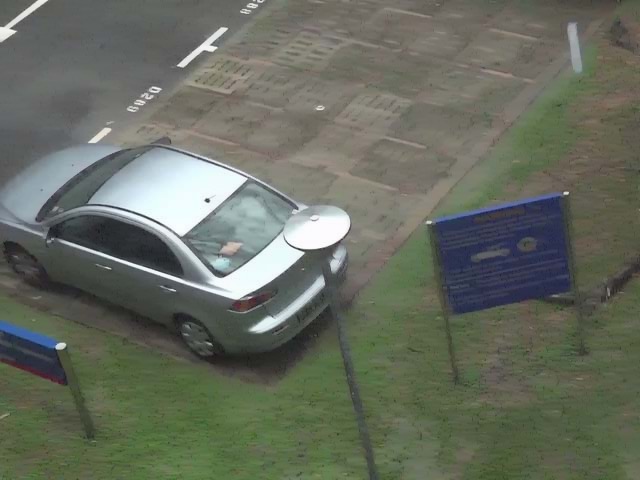}};
\spy [closeup_rain,magnification=3] on ($(Fig2B)+(-0.25,-0.1)$) 
    in node[largewindow,anchor=north west] at ($(Fig2B.south west) + (0.04,0)$);
\node [anchor=north] at ($(Fig2B.south)+(0,-0.23)$) {\small FastDeRain~\cite{Jiang:2018:FastDeRain}};

\node[anchor=south] (Fig2C) at (2,-1) {\includegraphics[width=0.23\textwidth]{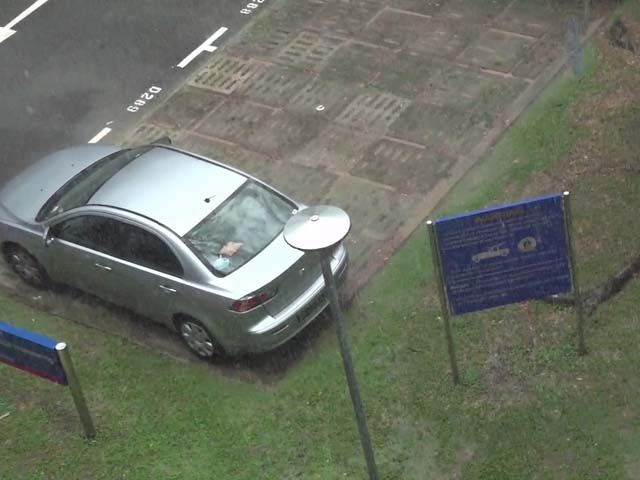}};
\spy [closeup_rain,magnification=3] on ($(Fig2C)+(-0.25,-0.1)$) 
    in node[largewindow,anchor=north west] at ($(Fig2C.south west) + (0.04,0)$);
\node [anchor=north] at ($(Fig2C.south)+(0,-0.23)$) {\small SpacCNN~\cite{Chen:2018:NTURain}};
    
\node[anchor=south] (Fig2D) at (3,-1) {\includegraphics[width=0.23\textwidth]{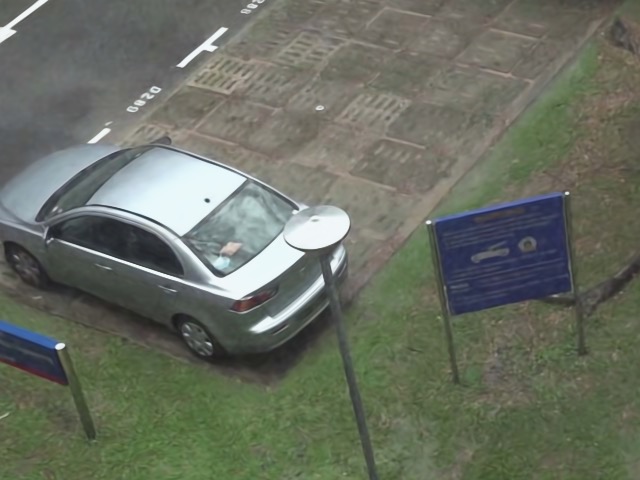}};
\spy [closeup_rain,magnification=3] on ($(Fig2D)+(-0.25,-0.1)$) 
    in node[largewindow,anchor=north west] at ($(Fig2D.south west) + (0.04,0)$);
\node [anchor=north] at ($(Fig2D.south)+(0,-0.23)$) {\small Ours};
\end{tikzpicture}
\caption{Qualitative comparison of rain removal on real images in NTURain~\cite{Chen:2018:NTURain}.}
\label{fig:rain}
\vspace{-5.0pt}
\end{figure*}

%% file: figures/ablation_combined.tex
\begin{figure}
\begin{minipage}[b]{0.48\textwidth}
    \centering
    {
    \setlength{\tabcolsep}{3.5pt}
    \renewcommand{\arraystretch}{0.5}
    \begin{tabular}{ccc}
    \multirow{2}{*}[0.43in]{\includegraphics[width=0.28\textwidth]{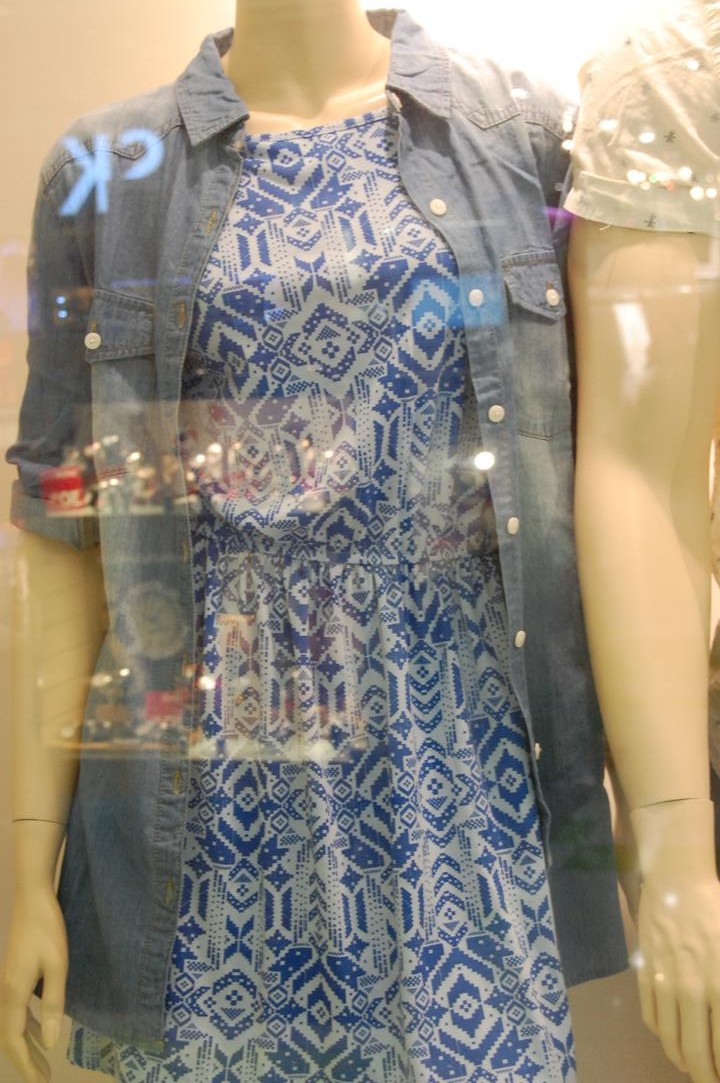}} & \includegraphics[trim=0 550 0 0,clip,width=0.28\textwidth]{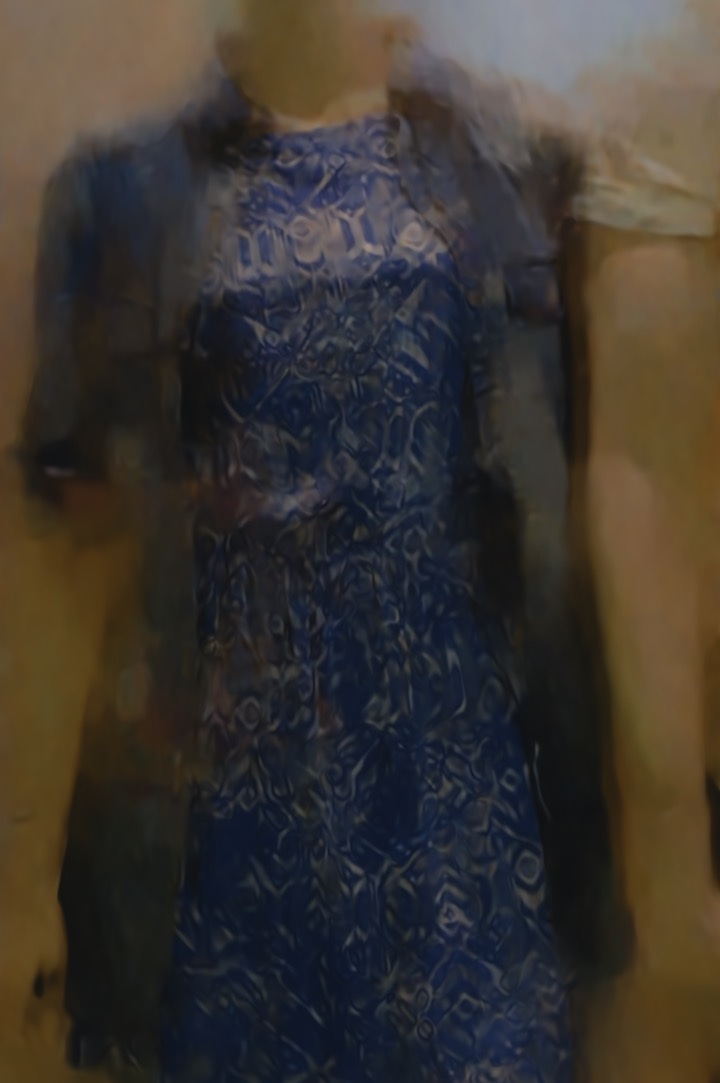} & \includegraphics[trim=0 550 0 0,clip,width=0.28\textwidth]{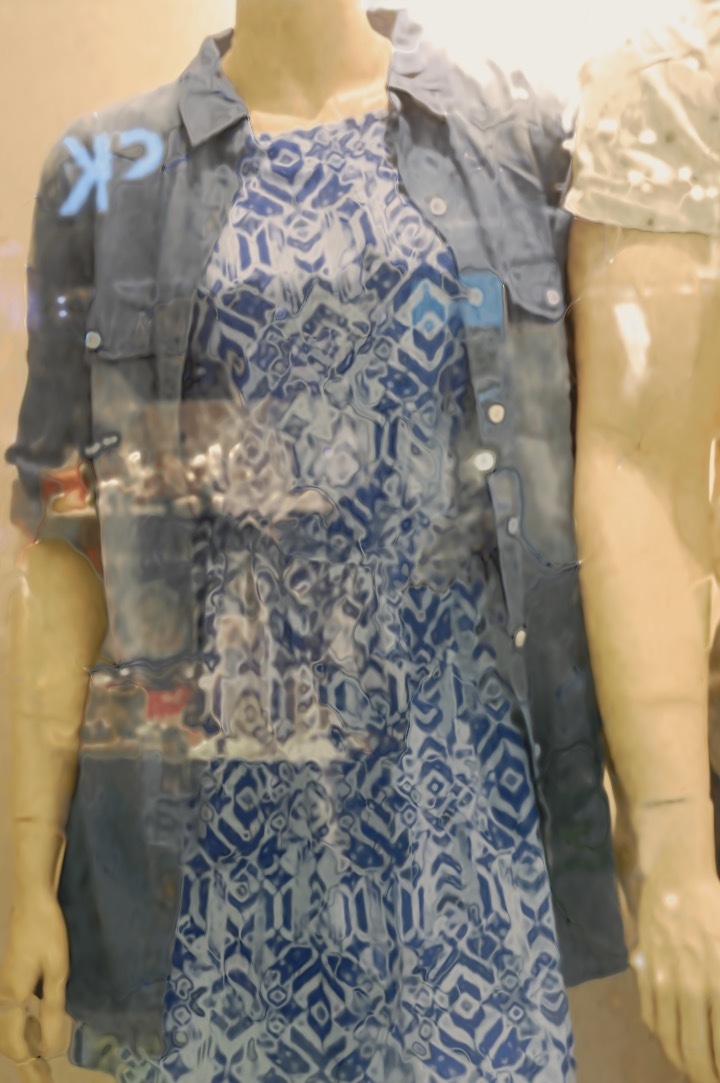} \\
    & \includegraphics[trim=0 0 0 550,clip,width=0.28\textwidth]{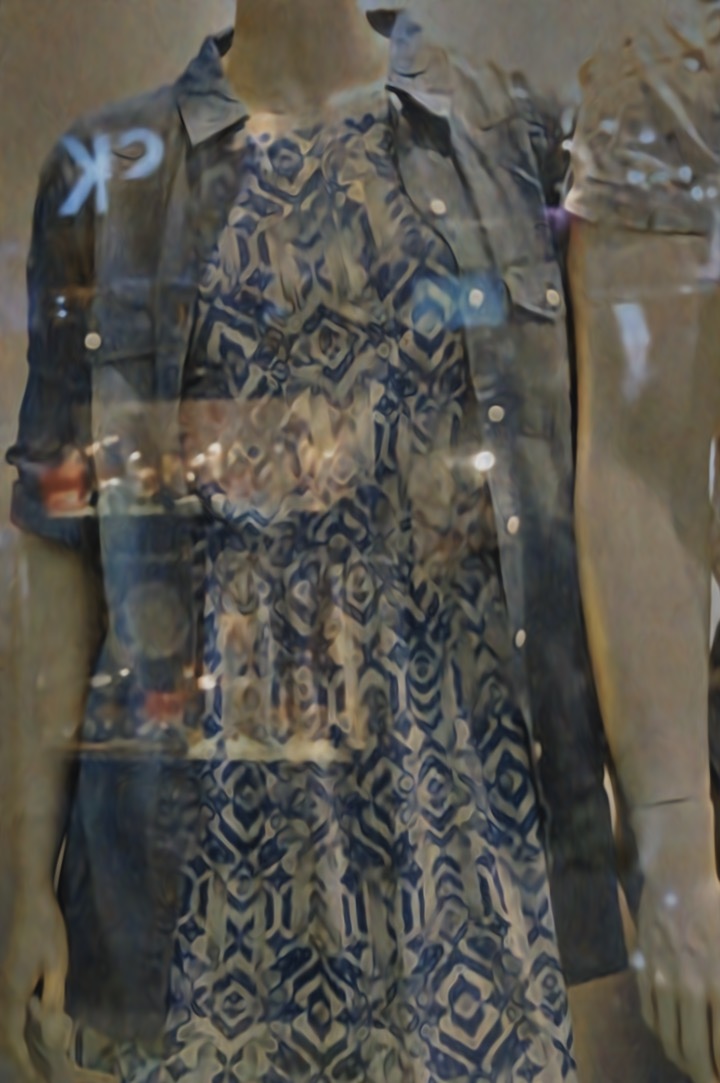} & \includegraphics[trim=0 0 0 550,clip,width=0.28\textwidth]{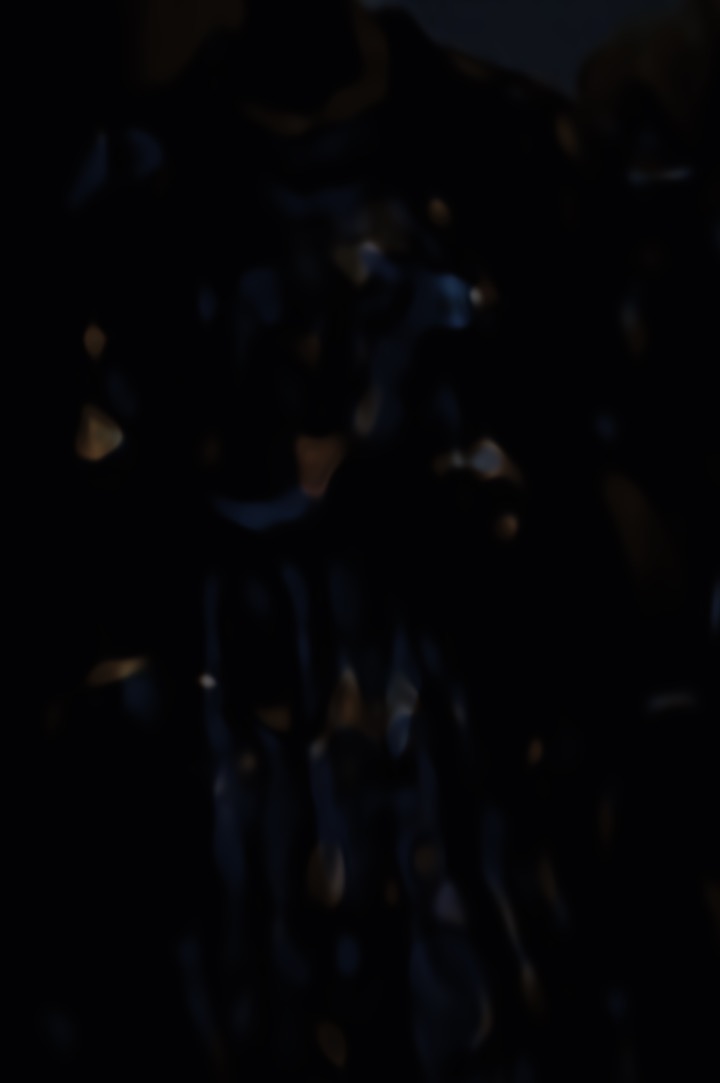} \\
    {\small Input} & {\small W/o $\lossftwo$} & {\small W/o $\lossTVFlow$} \\
    & \includegraphics[trim=0 550 0 0,clip,width=0.28\textwidth]{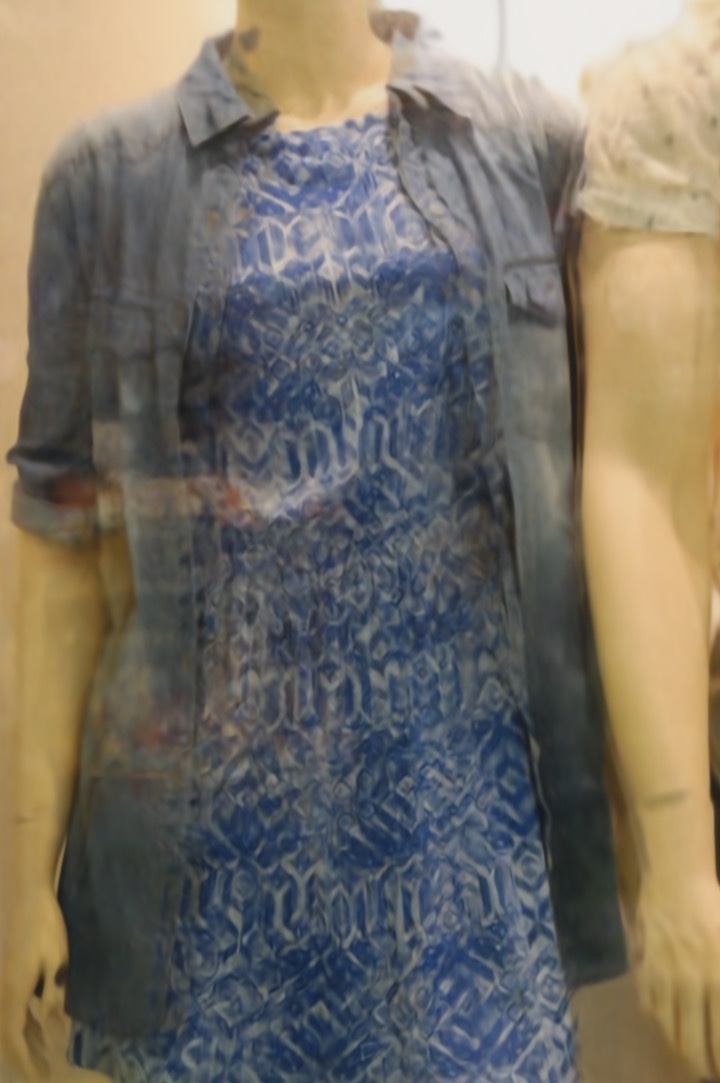} & \includegraphics[trim=0 550 0 0,clip,width=0.28\textwidth]{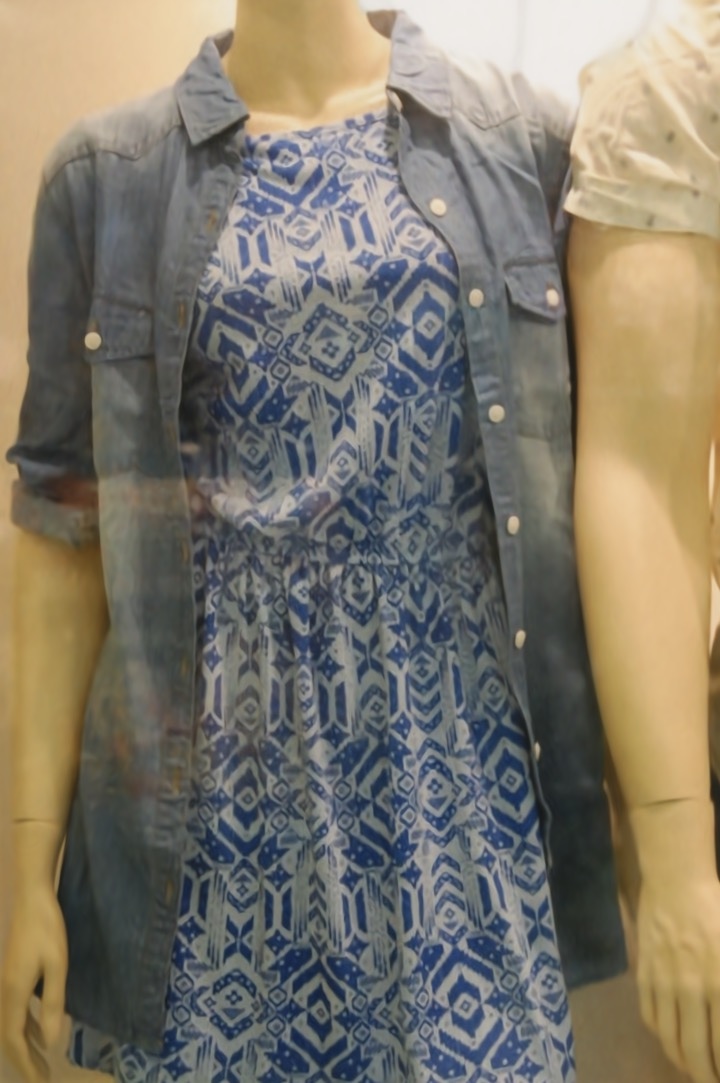} \\
    & \includegraphics[trim=0 0 0 550,clip,width=0.28\textwidth]{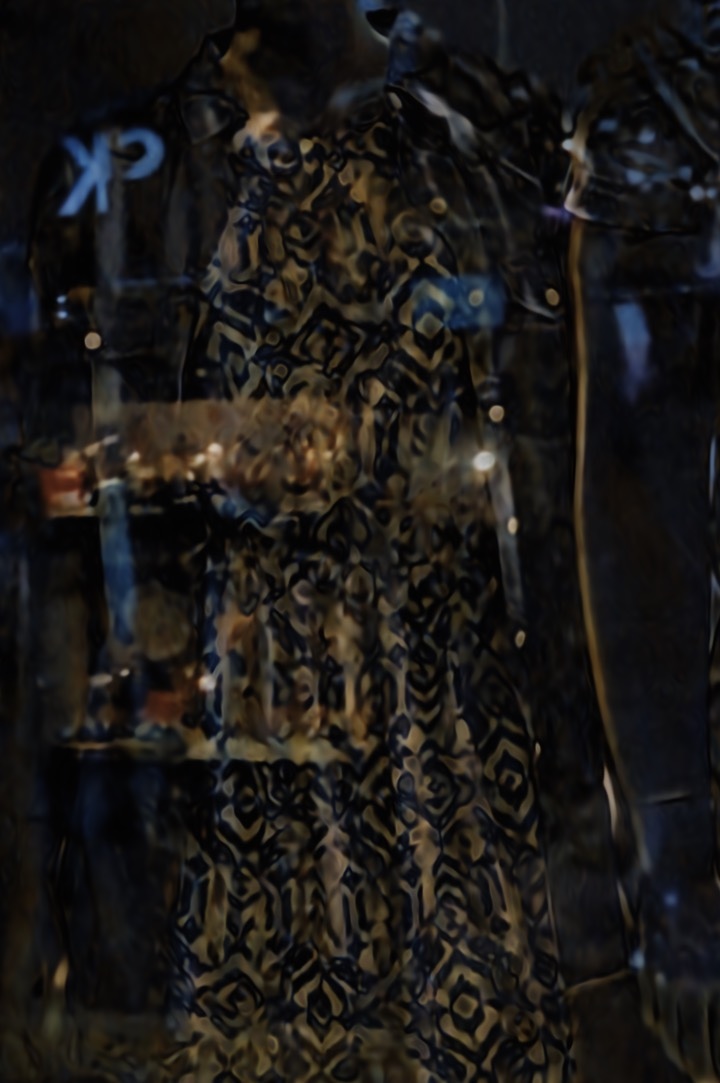} & \includegraphics[trim=0 0 0 550,clip,width=0.28\textwidth]{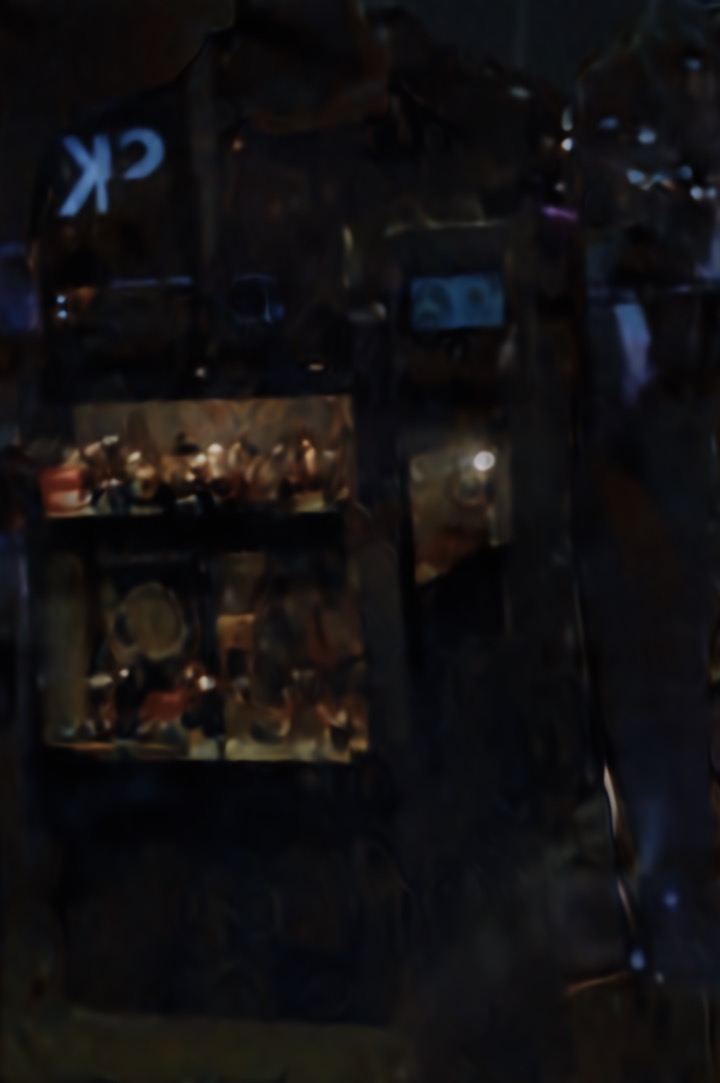} \\
    & {\small W/o $\lossExcl$} & {\small Full objective}
    \end{tabular}
    }
    \captionof{figure}{Ablation study of loss functions on reflection removal. The top and bottom images show the background and reflection layer, respectively.}
    \label{fig:ablation_reflection}
\end{minipage}
\hfill
\begin{minipage}[b]{0.48\textwidth}
    \centering
    \setlength{\tabcolsep}{1pt}
    \begin{tabular}{cc}
    \includegraphics[width=0.45\textwidth]{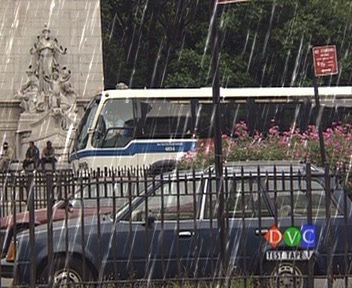} & \includegraphics[width=0.45\textwidth]{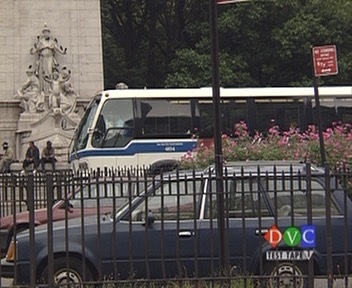} \\
    {\small Input} & {\small Ground Truth} \\
    \includegraphics[width=0.45\textwidth]{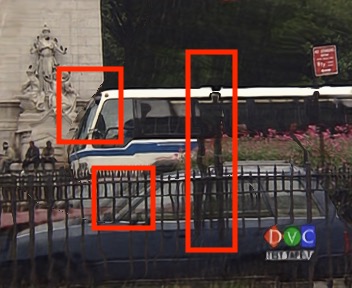} & \includegraphics[width=0.45\textwidth]{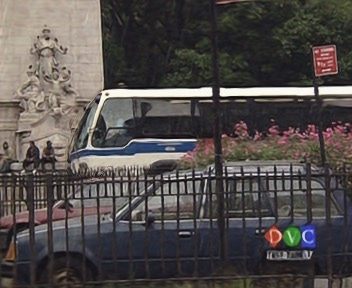} \\
    {\small W/o $w$} & {\small W/ $w$} \\
    {\small PSNR: 20.42} & {\small PSNR: 24.15}
    \end{tabular}
    \captionof{figure}{Ablation study on $w$. We show PSNRs of two outputs using the synthetic dataset RainSynLight25~\cite{Liu:2018:RainSyn}.}
    \label{fig:ablation_w}
\end{minipage}
\end{figure}

%% file: figures/further_discussion.tex
\begin{figure}[t]
\begin{minipage}[b]{0.48\textwidth}
    \setlength{\tabcolsep}{1pt}
    \centering
    \begin{tabular}{ccc}
        \includegraphics[width=0.32\linewidth,clip,trim=150 210 0 0]{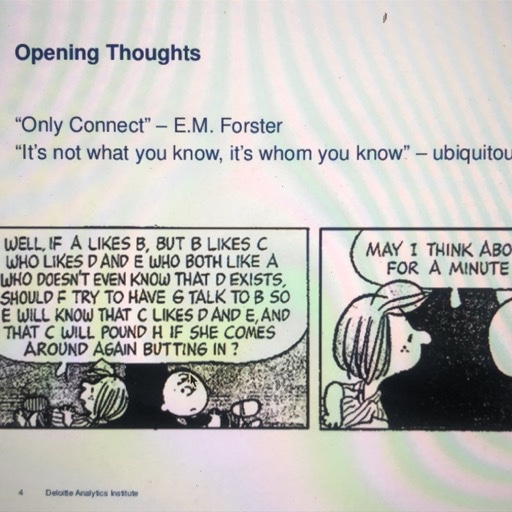} & \includegraphics[width=0.32\linewidth,clip,trim=150 210 0 0]{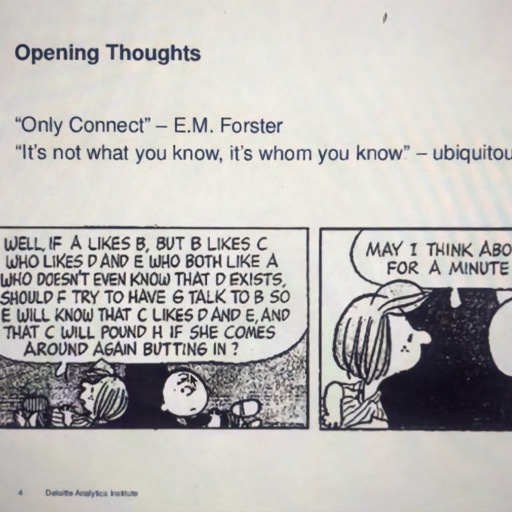} & \includegraphics[width=0.32\linewidth,clip,trim=150 210 0 0]{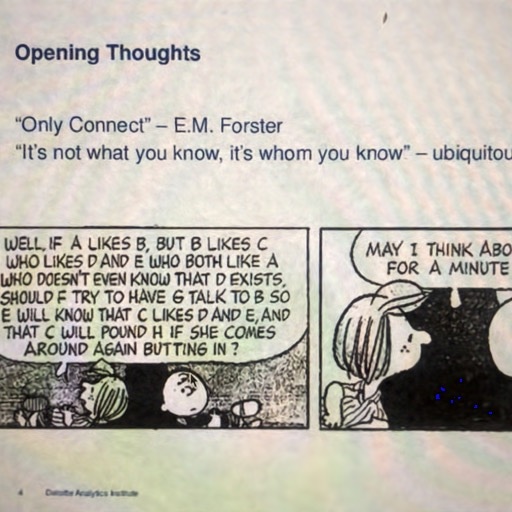} \\
        {\small Input} & {\small Homography} & {\small Optical flow} \\
    \end{tabular}
    \captionof{figure}{Analysis on different motion models. We apply homography-based (center) and occlusion-aware flow-based (right) models to demoir\'eing.}
    \label{fig:model_experiment}
\end{minipage}
\hfill
\begin{minipage}[b]{0.48\textwidth}
    \setlength{\tabcolsep}{1pt}
    \centering
    \begin{tabular}{ccc}
        \includegraphics[width=0.32\linewidth,clip,trim=200 100 250 180]{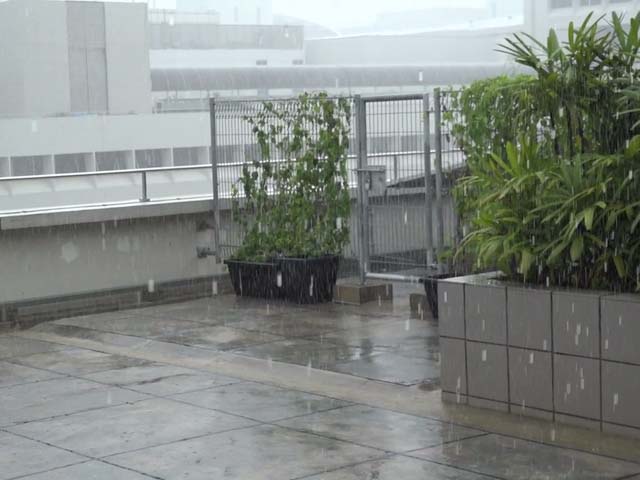} & \includegraphics[width=0.32\linewidth,clip,trim=200 100 250 180]{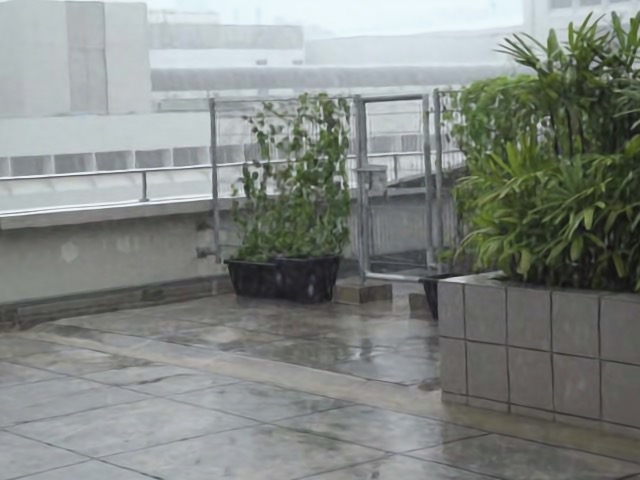} & \includegraphics[width=0.32\linewidth,clip,trim=200 100 250 180]{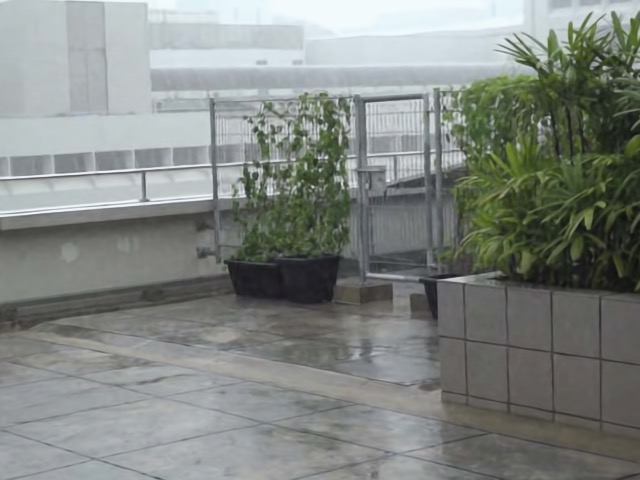} \\
        {\small Input} & {\small 2 images} & {\small 5 images} \\
    \end{tabular}
    \captionof{figure}{Analysis on the number of input images. We test 2 and 5 input images on a rain removal task.}
    \label{fig:number_of_images}
\end{minipage}
\end{figure}

%% file: figures/analysis_supp.tex
\begin{figure}[t]
\begin{minipage}[b]{0.48\textwidth}
    \centering
    \setlength{\tabcolsep}{5pt}
    \begin{tabular}{cc}
    \includegraphics[width=0.4\linewidth]{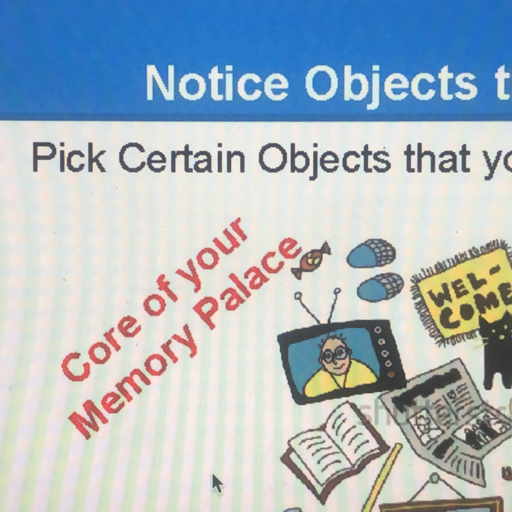} & \\
    {\small Input} & \\
    \includegraphics[width=0.4\linewidth]{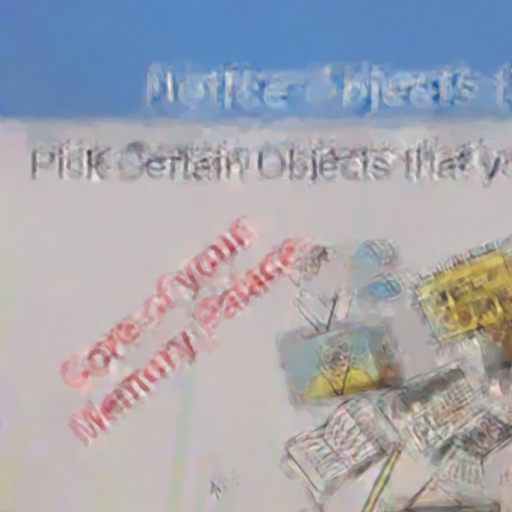} & \includegraphics[width=0.4\linewidth]{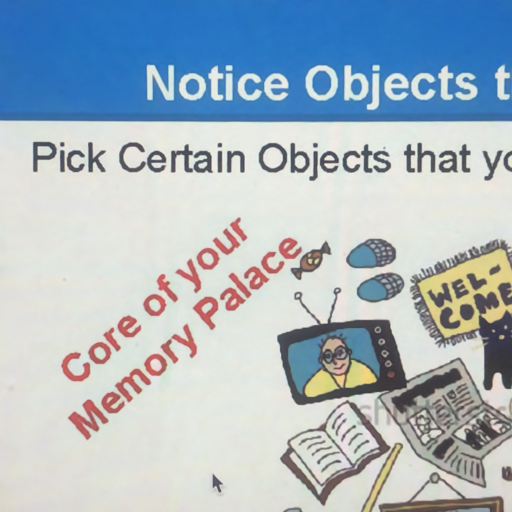} \\
    {\small Homography~\cite{Hartley:2003:Homography}} & {\small Ours}
    \end{tabular}
    \captionof{figure}{Comparison with conventional homography estimation. We adopt a homography estimation method~\cite{Hartley:2003:Homography} in a homography-based NIR.}
    \label{fig:analysis_motion}
\end{minipage}
\hfill
\begin{minipage}[b]{0.48\textwidth}
    \centering
    \setlength{\tabcolsep}{1pt}
    \begin{tabular}{cccccccc}
    \multicolumn{4}{c}{\includegraphics[width=0.44\textwidth]{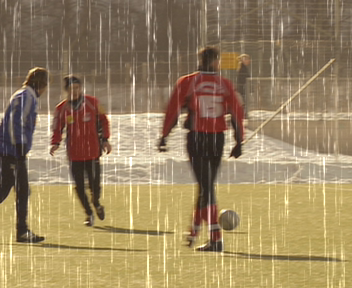}} & \multicolumn{2}{c}{\includegraphics[trim=0 0 176 0,clip,width=0.22\textwidth]{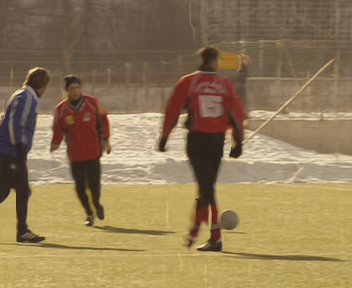}} & \multicolumn{2}{c}{\includegraphics[trim=176 0 0 0,clip,width=0.22\textwidth]{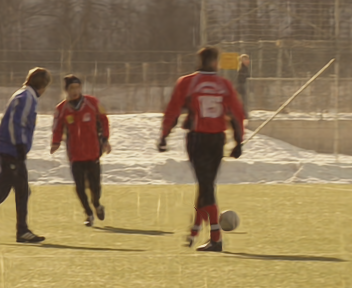}} \\
    \multicolumn{4}{c}{\small Input} & \multicolumn{2}{c}{\small W/ $\lossW$} & \multicolumn{2}{c}{\small W/o $\lossW$} \\
    \multicolumn{3}{c}{\includegraphics[width=0.356\textwidth]{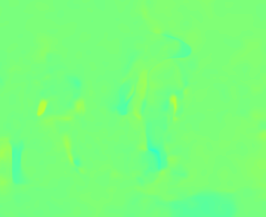}} & \includegraphics[width=0.092\textwidth]{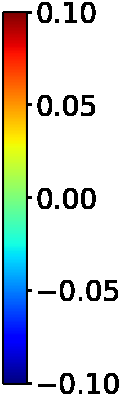} & \multicolumn{3}{c}{\includegraphics[width=0.356\textwidth]{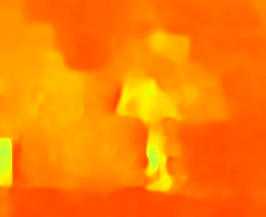}} & \includegraphics[width=0.092\textwidth]{figures/analysis_w/w_tick.png} \\
    \multicolumn{4}{c}{\small W/ $\lossW$} & \multicolumn{4}{c}{\small W/o $\lossW$} \\
    \multicolumn{4}{c}{\small PSNR: 30.31 $\pm$ 0.53} & \multicolumn{4}{c}{\small PSNR: 31.26 $\pm$ 0.69}
    \end{tabular}
    \captionof{figure}{Analysis on $w$. We apply an additional loss $\lossW = \sum~\lVert w \rVert_1$ and compare it with the original model. For PSNRs, we train each model 5 times and aggregates results.}
    \label{fig:analysis_reg_w}
\end{minipage}
\end{figure}

%% file: figures/result_denoising.tex
\tikzstyle{closeup_denoising} = [
  opacity=1.0,          
  height=0.9cm,         
  width=0.236\textwidth, 
  connect spies, red  
]

\begin{figure*}[t]
\centering
\begin{tikzpicture}[x=0.25\textwidth, y=0.2\textheight, spy using outlines={every spy on node/.append style={smallwindow}}]
\node[anchor=south] (Fig1A) at (0,0) {\includegraphics[width=0.24\textwidth]{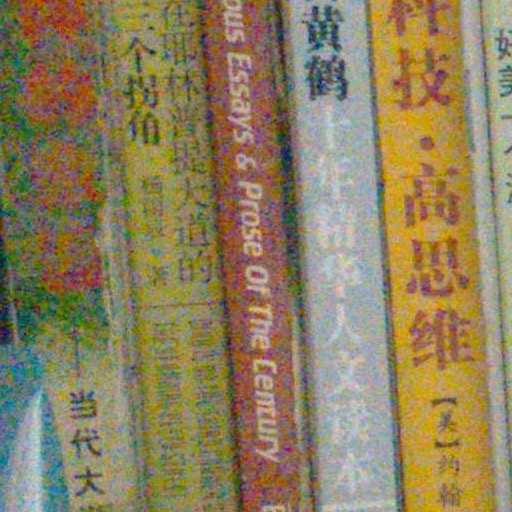}};
\spy [closeup_denoising,magnification=2] on ($(Fig1A)+(-0.15,0.25)$) 
    in node[largewindow,anchor=north west] at ($(Fig1A.south west) + (0.04,0)$);
\node [anchor=north] at ($(Fig1A.south)+(0,-0.25)$) {\small Input};

\node[anchor=south] (Fig1B) at (1,0) {\includegraphics[width=0.24\textwidth]{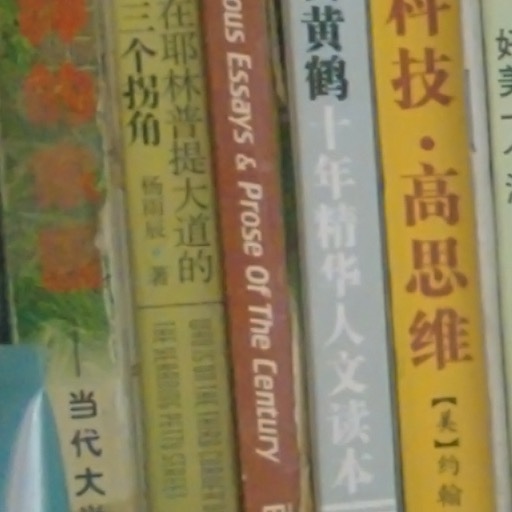}};
\spy [closeup_denoising,magnification=2] on ($(Fig1B)+(-0.15,0.25)$) 
    in node[largewindow,anchor=north west] at ($(Fig1B.south west) + (0.04,0)$);
\node [anchor=north] at ($(Fig1B.south)+(0,-0.25)$) {\small FBID~\cite{Liu:2014:FBID}};

\node[anchor=south] (Fig1C) at (2,0) {\includegraphics[width=0.24\textwidth]{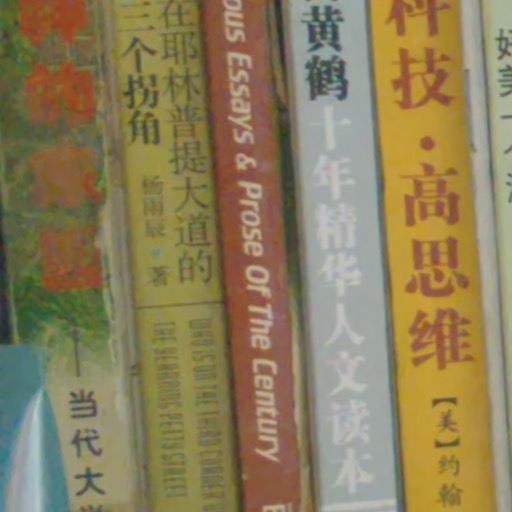}};
\spy [closeup_denoising,magnification=2] on ($(Fig1C)+(-0.15,0.25)$) 
    in node[largewindow,anchor=north west] at ($(Fig1C.south west) + (0.04,0)$);
\node [anchor=north] at ($(Fig1C.south)+(0,-0.25)$) {\small Ours (Scene)};
    
\node[anchor=south] (Fig1D) at (3,0) {\includegraphics[width=0.24\textwidth]{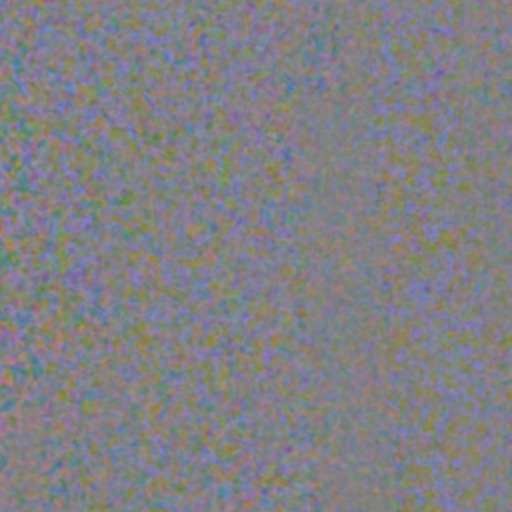}};
\spy [closeup_denoising,magnification=2] on ($(Fig1D)+(-0.15,0.25)$) 
    in node[largewindow,anchor=north west] at ($(Fig1D.south west) + (0.04,0)$);
\node [anchor=north] at ($(Fig1D.south)+(0,-0.25)$) {\small Ours (Noise)};
\end{tikzpicture}
\caption{Additional application to burst image denoising on the images in~\cite{Liu:2014:FBID}.}
\label{fig:denoising}
\end{figure*}

%% file: figures/result_sr.tex
\tikzstyle{closeup_sr} = [
  opacity=1.0,          
  height=0.9cm,         
  width=0.236\textwidth, 
  connect spies, red  
]

\begin{figure*}[t]
\centering
\begin{tikzpicture}[x=0.25\textwidth, y=0.2\textheight, spy using outlines={every spy on node/.append style={smallwindow}}]
\node[anchor=south] (Fig1A) at (0,0) {\includegraphics[interpolate=false,width=0.24\textwidth]{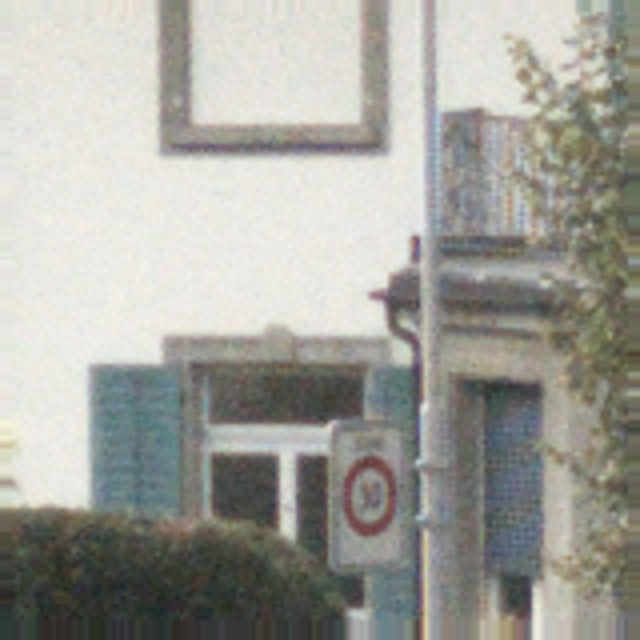}};
\spy [closeup_sr,magnification=2] on ($(Fig1A)+(0.15,-0.25)$) 
    in node[largewindow,anchor=north west] at ($(Fig1A.south west) + (0.04,0)$);
\node [anchor=north] at ($(Fig1A.south)+(0,-0.25)$) {\small Original (Bicubic)};

\node[anchor=south] (Fig1B) at (1,0) {\includegraphics[width=0.24\textwidth]{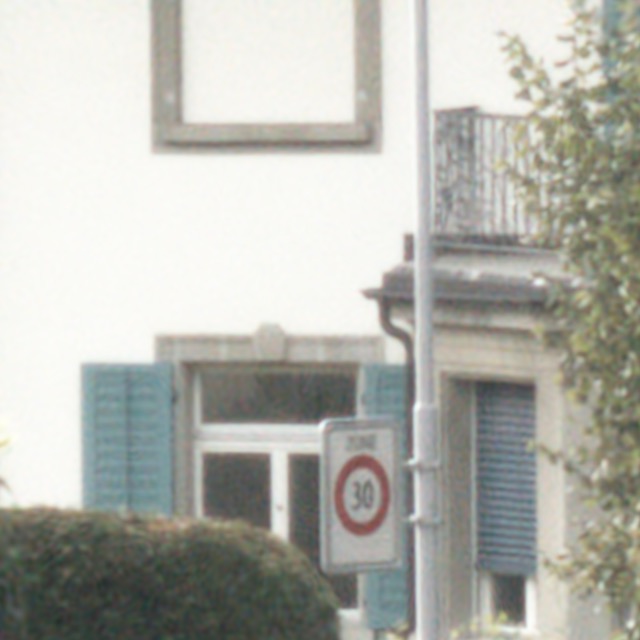}};
\spy [closeup_sr,magnification=2] on ($(Fig1B)+(0.15,-0.25)$) 
    in node[largewindow,anchor=north west] at ($(Fig1B.south west) + (0.04,0)$);
\node [anchor=north] at ($(Fig1B.south)+(0,-0.25)$) {\small Ours};

\node[anchor=south] (Fig1C) at (2,0) {\includegraphics[width=0.24\textwidth]{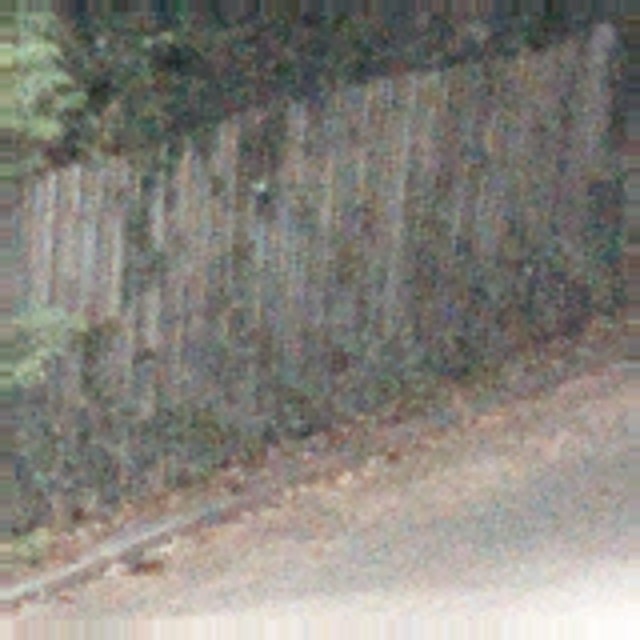}};
\spy [closeup_sr,magnification=2] on ($(Fig1C)+(0,0.10)$) 
    in node[largewindow,anchor=north west] at ($(Fig1C.south west) + (0.04,0)$);
\node [anchor=north] at ($(Fig1C.south)+(0,-0.25)$) {\small Original (Bicubic)};
    
\node[anchor=south] (Fig1D) at (3,0) {\includegraphics[width=0.24\textwidth]{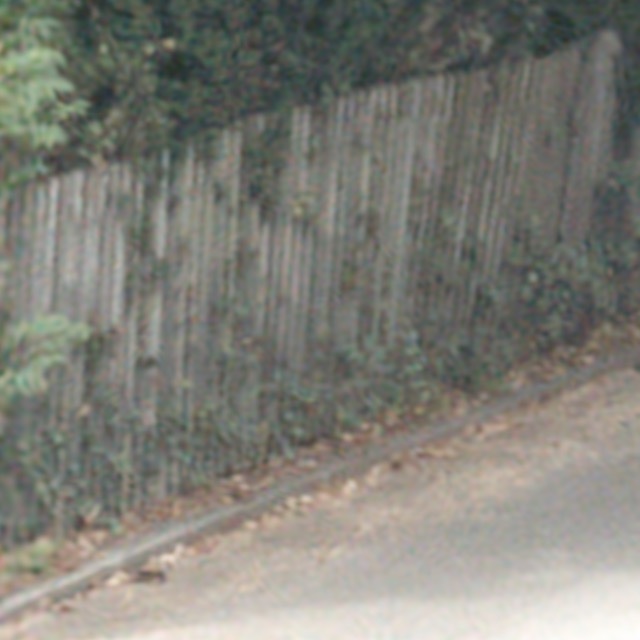}};
\spy [closeup_sr,magnification=2] on ($(Fig1D)+(0,0.10)$) 
    in node[largewindow,anchor=north west] at ($(Fig1D.south west) + (0.04,0)$);
\node [anchor=north] at ($(Fig1D.south)+(0,-0.25)$) {\small Ours};
\end{tikzpicture}
\caption{Additional application to joint image demosaicing and burst super-resolution on the images in~\cite{Bhat:2021:DeepBurstSR}.}
\label{fig:sr}
\end{figure*}

%% file: tables/obstruction.tex
\begin{table}
\begin{center}
\setlength{\tabcolsep}{10pt}
\begin{tabular}{ccccccc}
\toprule
\multirow{2}{*}{Method} & \multicolumn{2}{c}{Stone} & \multicolumn{2}{c}{Toy} & \multicolumn{2}{c}{Hanoi} \\
\cline{2-7}
& SSIM & NCC & SSIM & NCC & SSIM & NCC \\
\hline
~\cite{Li:2013:LiandBrown} & 0.7993 & 0.9334 & 0.6877 & 0.7068 & N/A & N/A \\
~\cite{Xue:2015:Computational} & N/A & 0.9738 & N/A & 0.8985 & N/A & 0.9921 \\
~\cite{Alayrac:2019:VisualCentrifuge} & 0.7942 & 0.9351 & 0.7569 & 0.7972 & N/A & N/A \\
~\cite{Liu:2020:LearningToSee} & 0.8598 & 0.9632 & 0.7696 & 0.9477 & 0.9238 & 0.9929 \\
~\cite{Liu:2020:LearningToSeeJournal} & 0.8635 & 0.9315 & 0.8494 & 0.9542 & 0.9457 & 0.9938 \\
Ours & 0.8617 & 0.9451 & 0.7700 & 0.8136 & 0.9045 & 0.9840 \\
\bottomrule
\end{tabular}
\end{center}
\caption{Quantitative result of obstruction removal on the data in~\cite{Xue:2015:Computational}}
\label{table:obstruction}
\end{table}

%% file: figures/supp_burst.tex
\begin{figure*}
    \centering
    \setlength{\tabcolsep}{1pt}
    \begin{tabular}{ccccc}
    \includegraphics[width=0.185\textwidth]{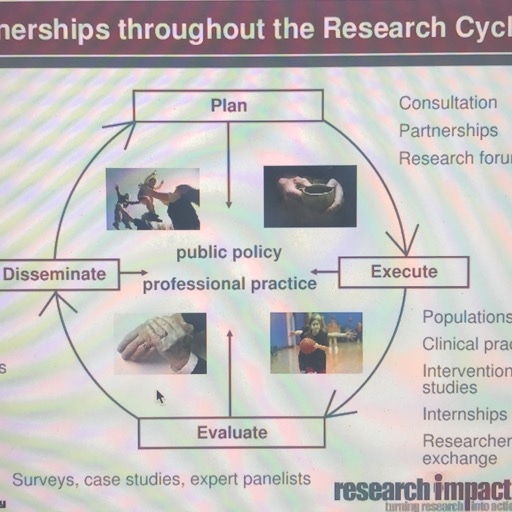} & \includegraphics[width=0.185\textwidth]{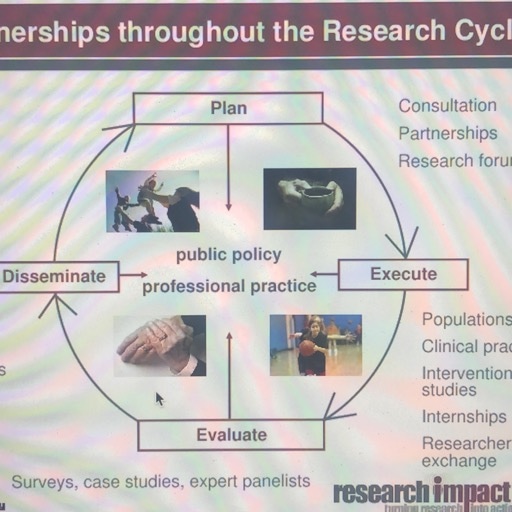} & \includegraphics[width=0.185\textwidth]{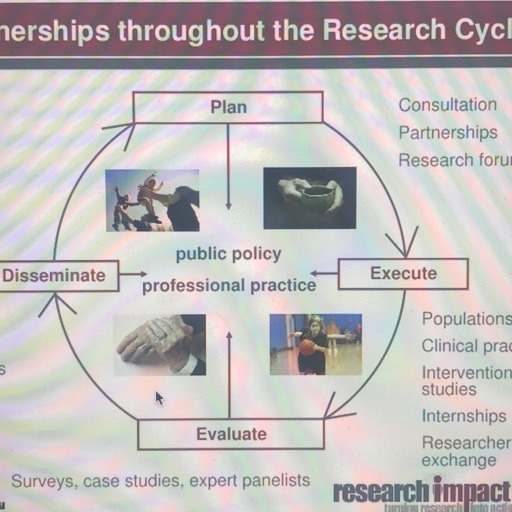} & \includegraphics[width=0.185\textwidth]{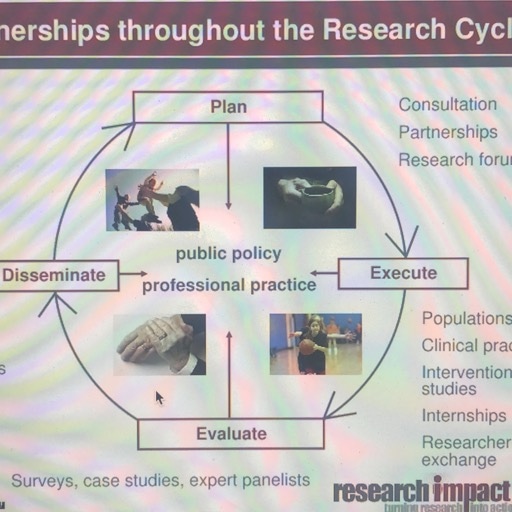} & \includegraphics[width=0.185\textwidth]{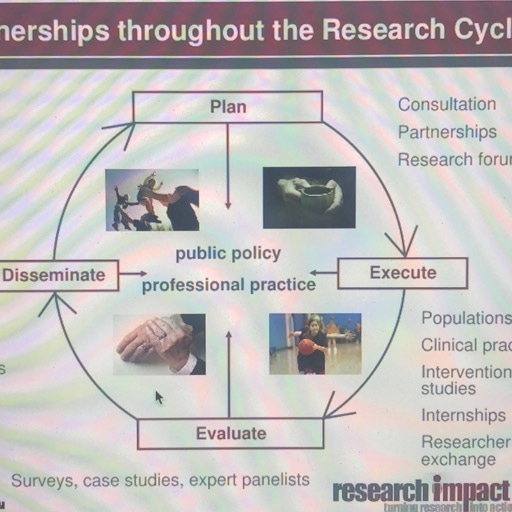} \\
    
    \includegraphics[width=0.185\textwidth]{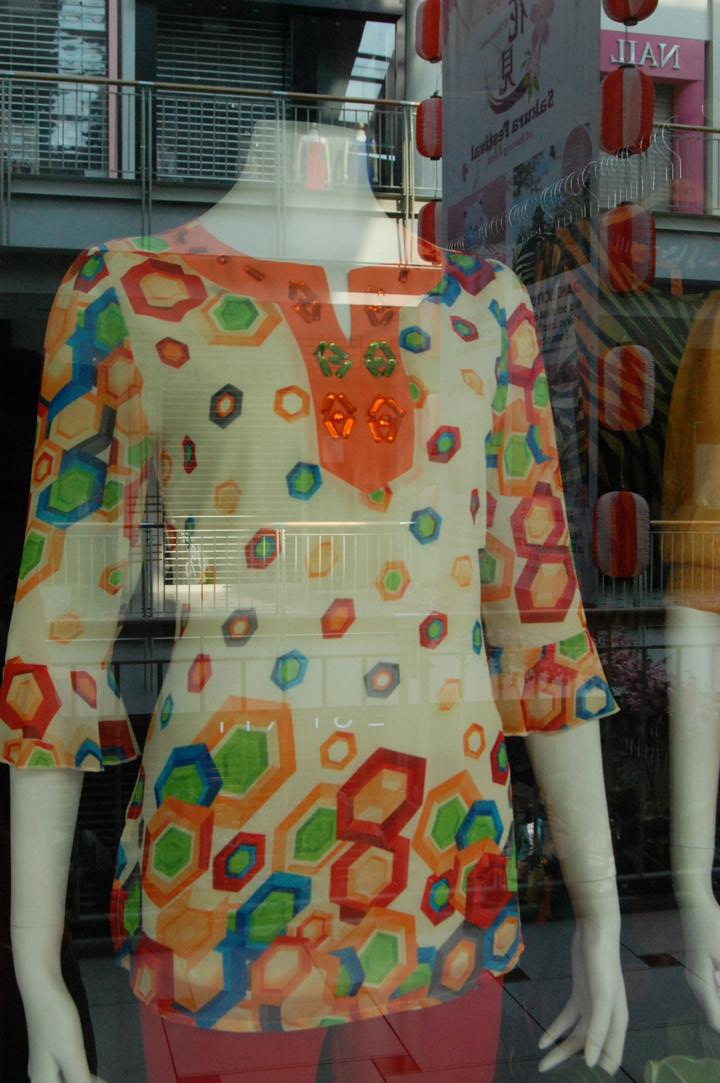} & \includegraphics[width=0.185\textwidth]{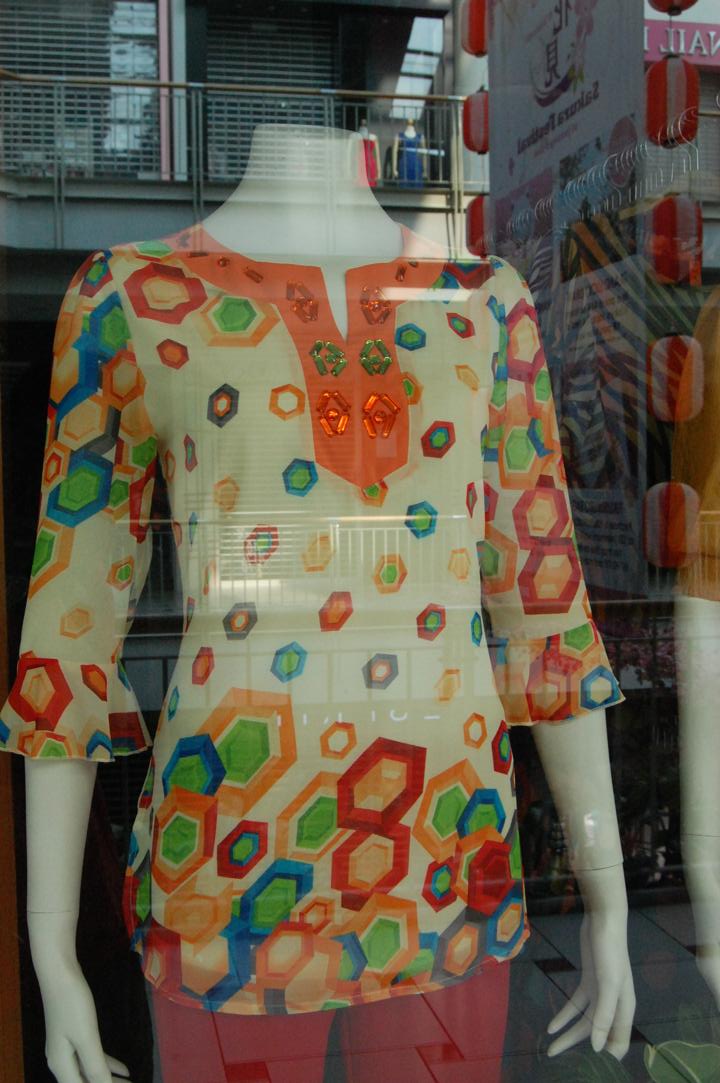} & \includegraphics[width=0.185\textwidth]{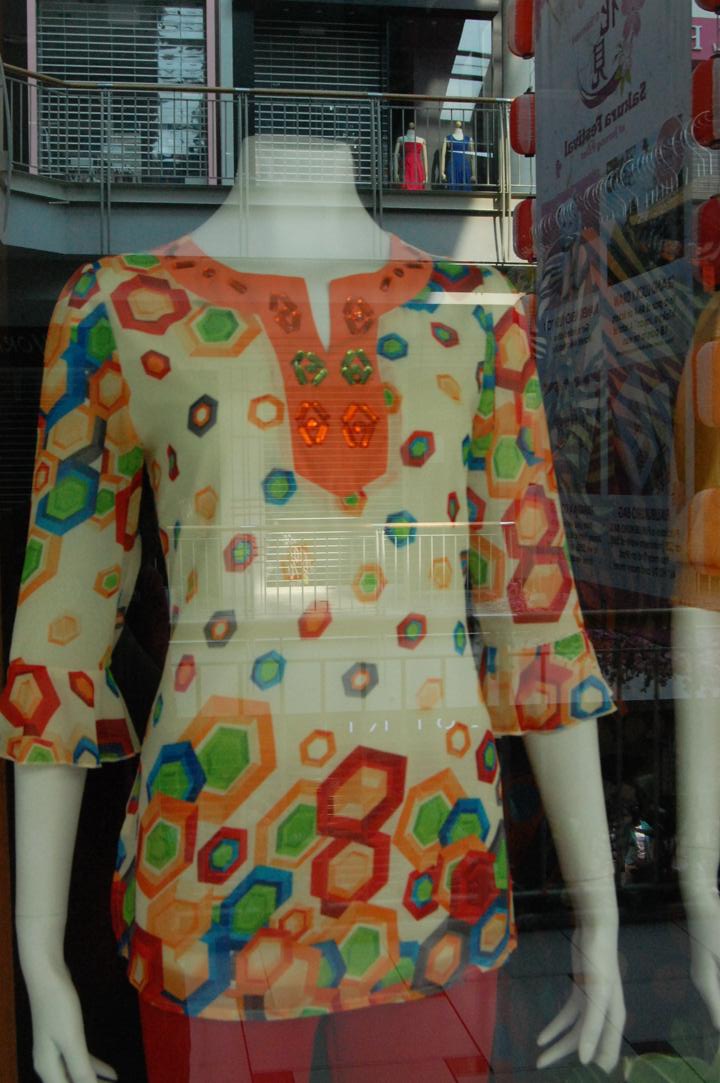} & \includegraphics[width=0.185\textwidth]{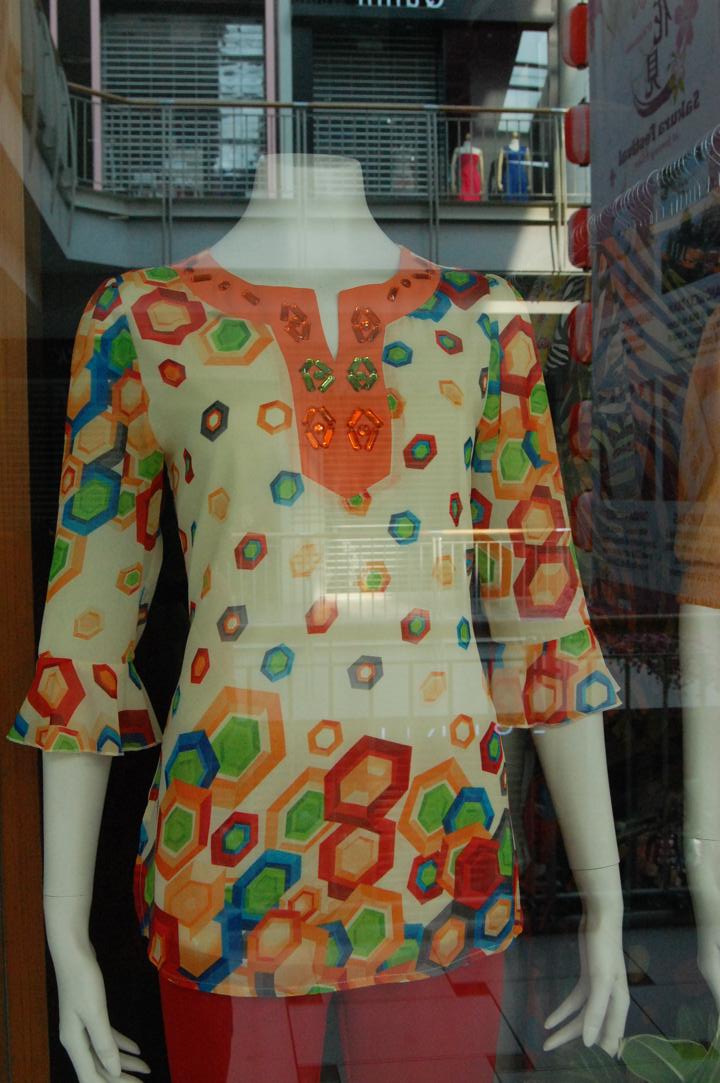} & \includegraphics[width=0.185\textwidth]{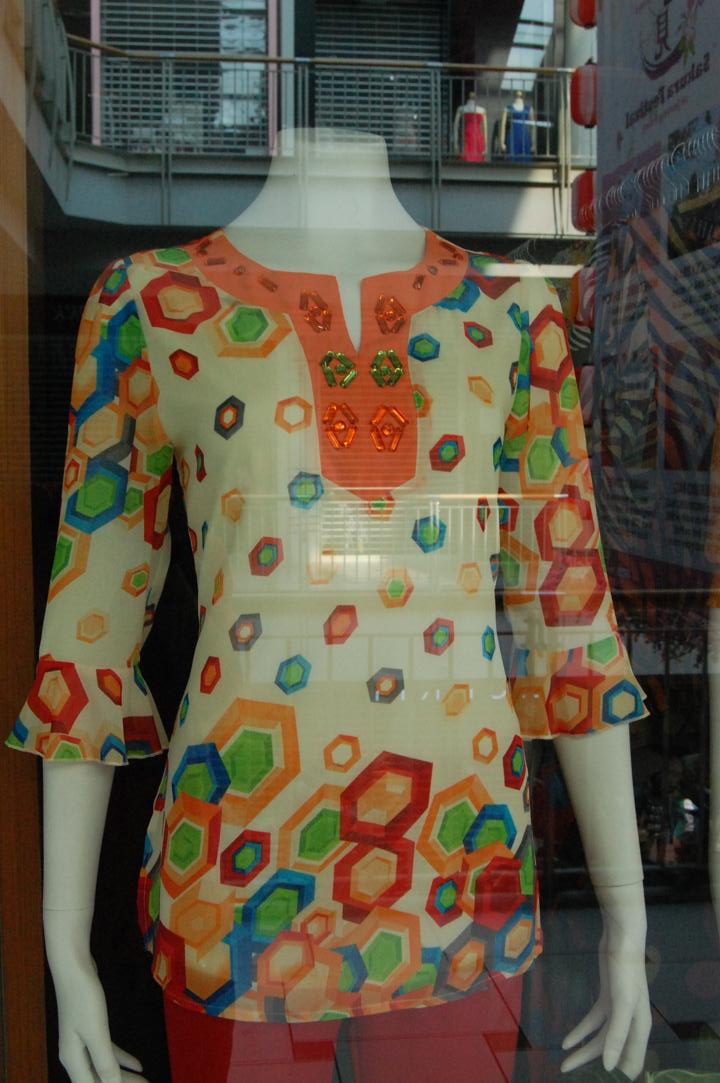} \\
    
    \includegraphics[width=0.185\textwidth]{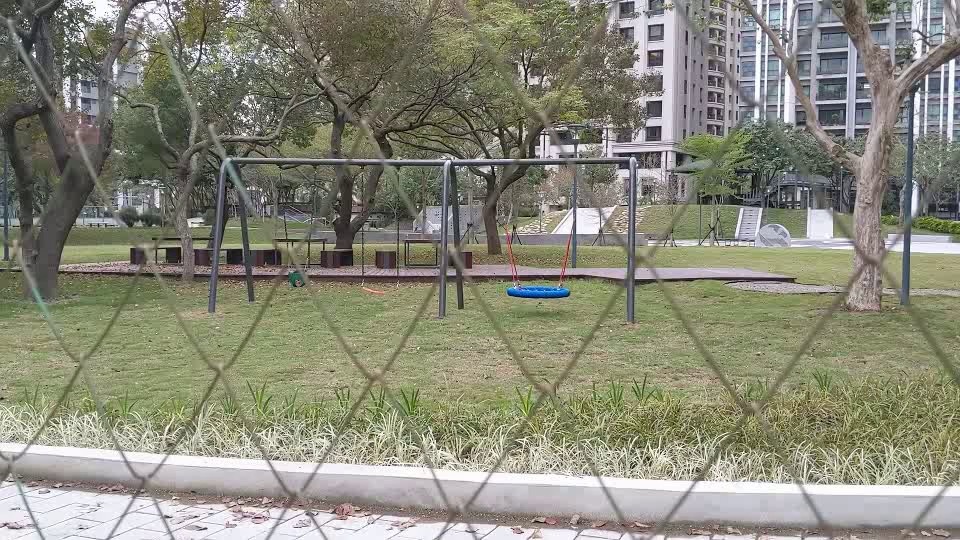} & \includegraphics[width=0.185\textwidth]{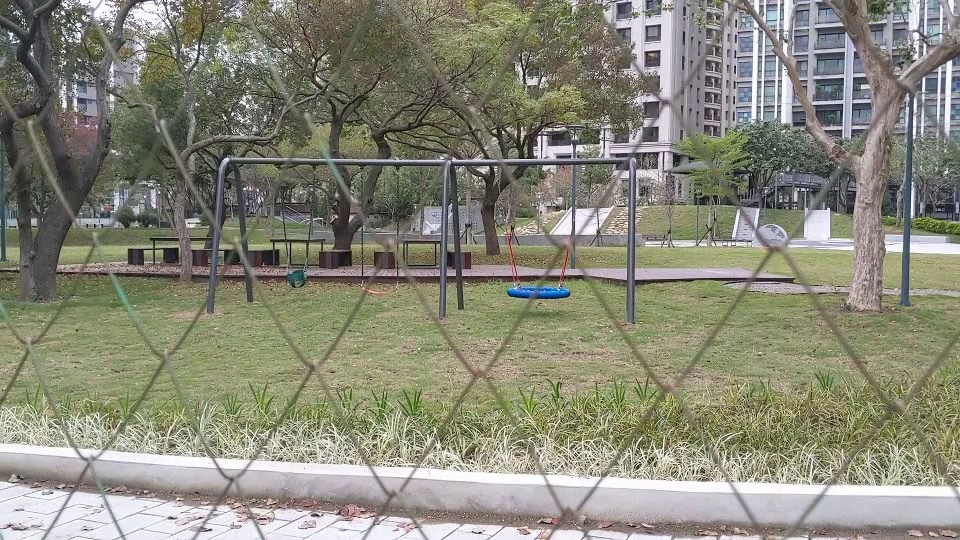} & \includegraphics[width=0.185\textwidth]{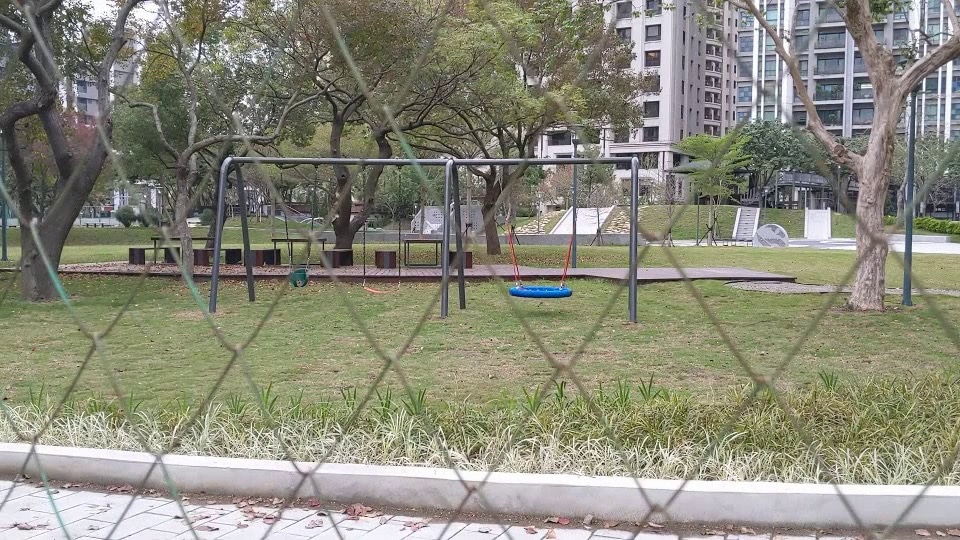} & \includegraphics[width=0.185\textwidth]{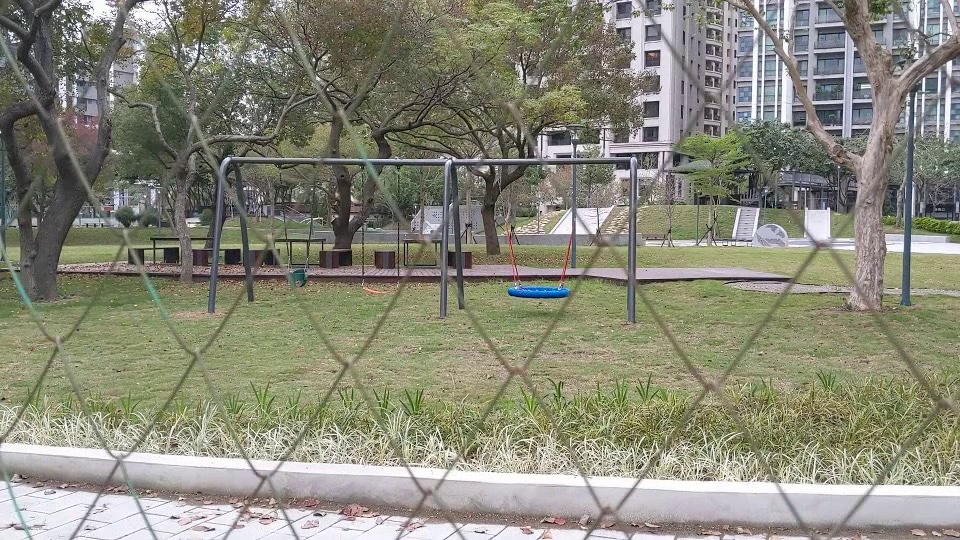} & \includegraphics[width=0.185\textwidth]{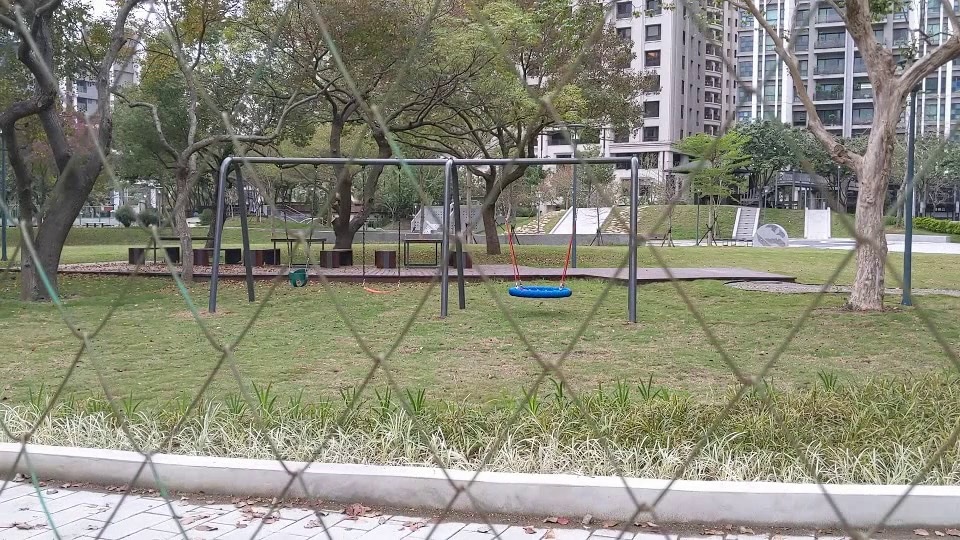} \\
    
    \includegraphics[width=0.185\textwidth]{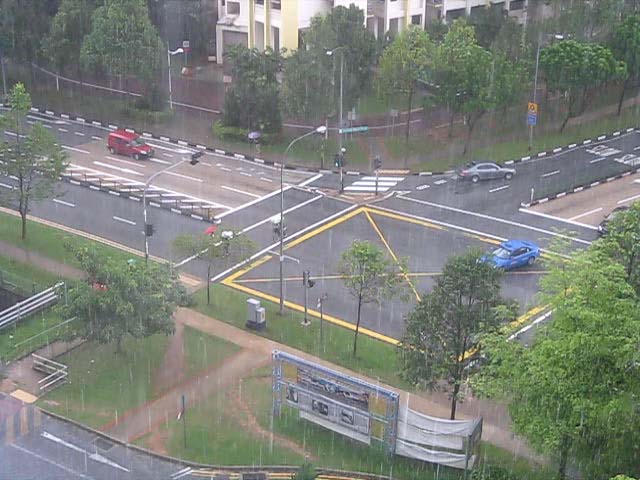} & \includegraphics[width=0.185\textwidth]{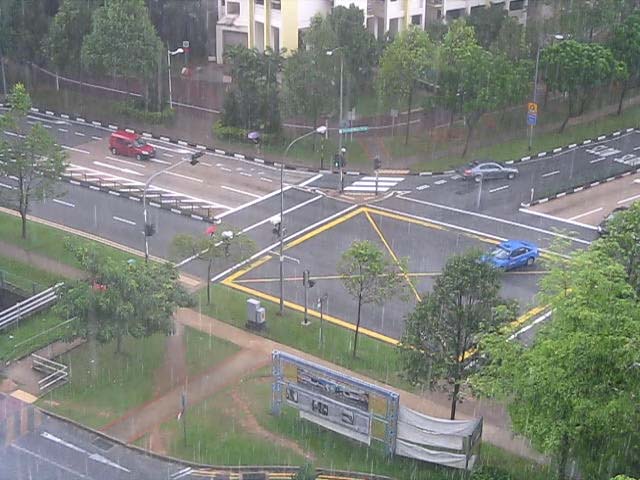} & \includegraphics[width=0.185\textwidth]{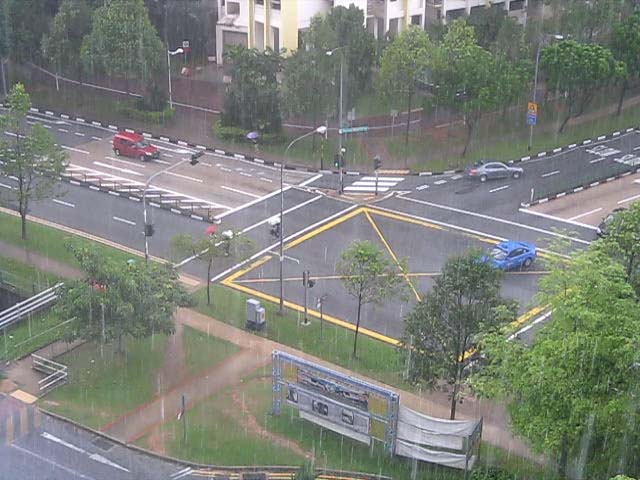} & \includegraphics[width=0.185\textwidth]{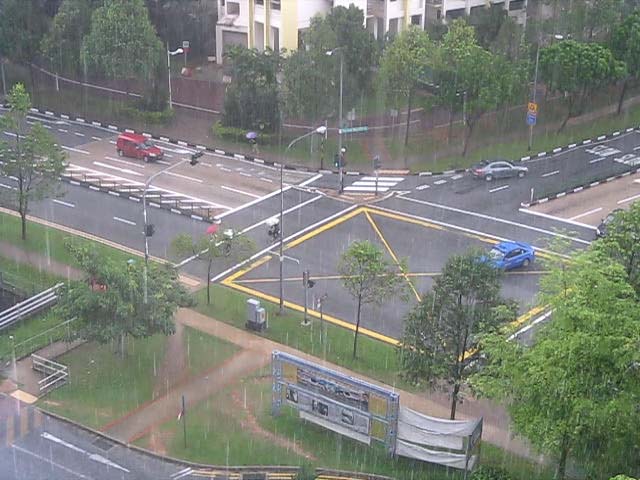} & \includegraphics[width=0.185\textwidth]{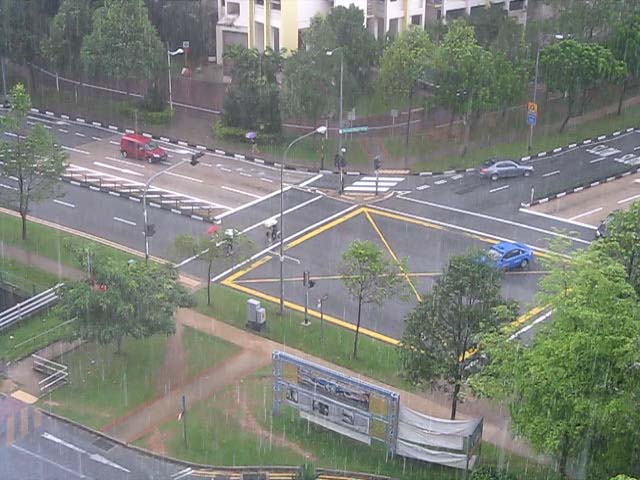} \\
    \end{tabular}
    \caption{Examples of burst images used in our paper. From top to bottom, each row shows burst images for moir\'e, reflection, fence, and rain removal, respectively.}
    \label{fig:supp_burst}
\end{figure*}

%% file: figures/supp_moire.tex
\tikzstyle{closeup} = [
  opacity=1.0,          
  height=0.9cm,         
  width=0.178\textwidth, 
  connect spies, red  
]

\begin{figure*}
\centering


    


\begin{tikzpicture}[x=0.19\textwidth, y=0.2\textheight, spy using outlines={every spy on node/.append style={smallwindow}}]
\node[anchor=south] (Fig1A) at (0,0) {\includegraphics[width=0.18\textwidth]{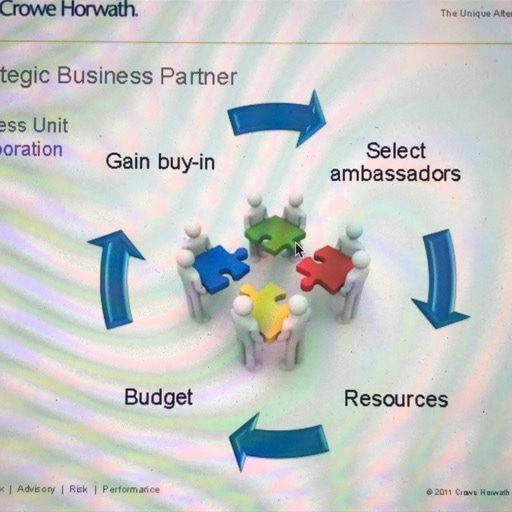}};
\spy [closeup,magnification=2] on ($(Fig1A)+(-0.15,-0.15)$) 
    in node[largewindow,anchor=north west] at ($(Fig1A.south west) + (0.04,0)$);

\node[anchor=south] (Fig1B) at (1,0) {\includegraphics[width=0.18\textwidth]{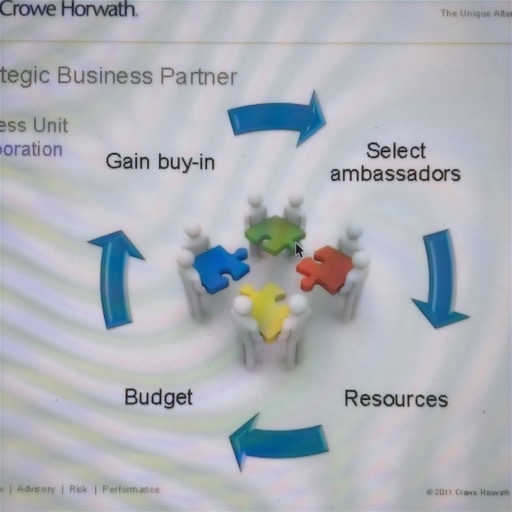}};
\spy [closeup,magnification=2] on ($(Fig1B)+(-0.15,-0.15)$) 
    in node[largewindow,anchor=north west] at ($(Fig1B.south west) + (0.04,0)$);

\node[anchor=south] (Fig1C) at (2,0) {\includegraphics[width=0.18\textwidth]{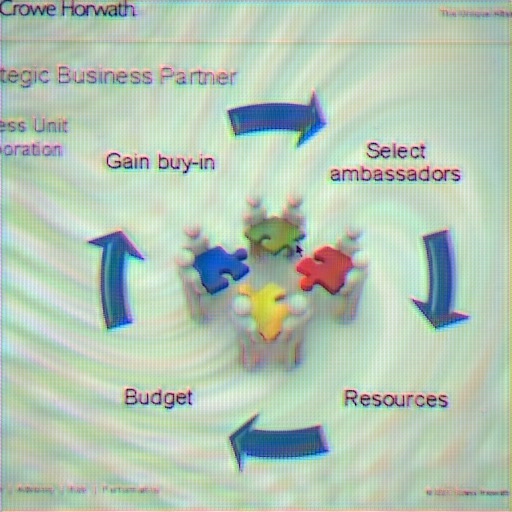}};
\spy [closeup,magnification=2] on ($(Fig1C)+(-0.15,-0.15)$) 
    in node[largewindow,anchor=north west] at ($(Fig1C.south west) + (0.04,0)$);
    
\node[anchor=south] (Fig1D) at (3,0) {\includegraphics[width=0.18\textwidth]{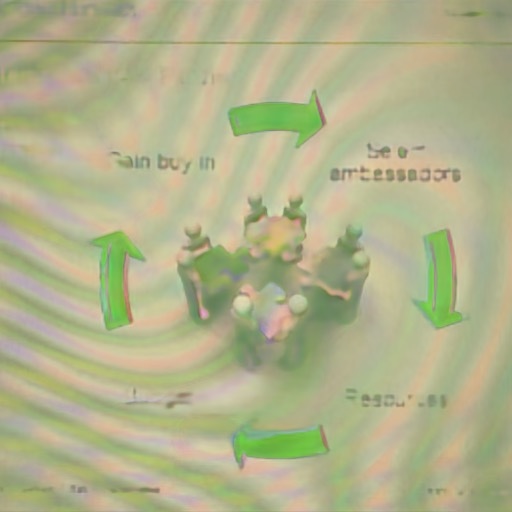}};
\spy [closeup,magnification=2] on ($(Fig1D)+(-0.15,-0.15)$) 
    in node[largewindow,anchor=north west] at ($(Fig1D.south west) + (0.04,0)$);

\node[anchor=south] (Fig1E) at (4,0) {\includegraphics[width=0.18\textwidth]{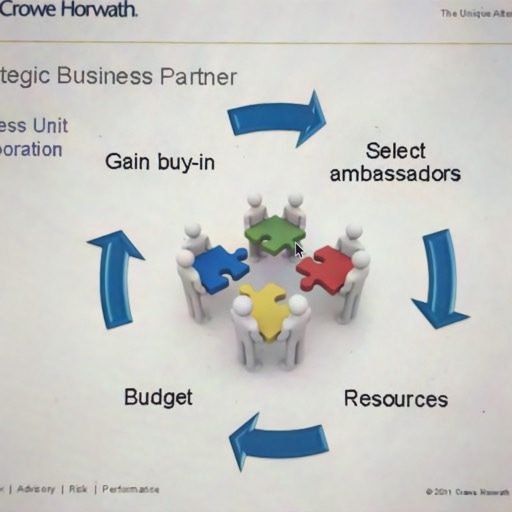}};
\spy [closeup,magnification=2] on ($(Fig1E)+(-0.15,-0.15)$) 
    in node[largewindow,anchor=north west] at ($(Fig1E.south west) + (0.04,0)$);
\end{tikzpicture}

\begin{tikzpicture}[x=0.19\textwidth, y=0.2\textheight, spy using outlines={every spy on node/.append style={smallwindow}}]
\node[anchor=south] (Fig1A) at (0,0) {\includegraphics[width=0.18\textwidth]{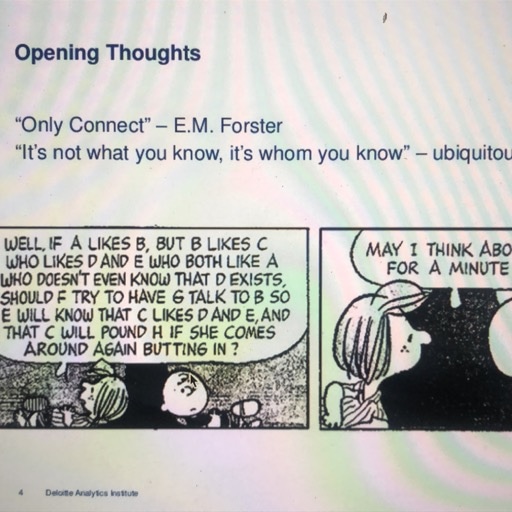}};
\spy [closeup,magnification=2] on ($(Fig1A)+(0,0)$) 
    in node[largewindow,anchor=north west] at ($(Fig1A.south west) + (0.04,0)$);

\node[anchor=south] (Fig1B) at (1,0) {\includegraphics[width=0.18\textwidth]{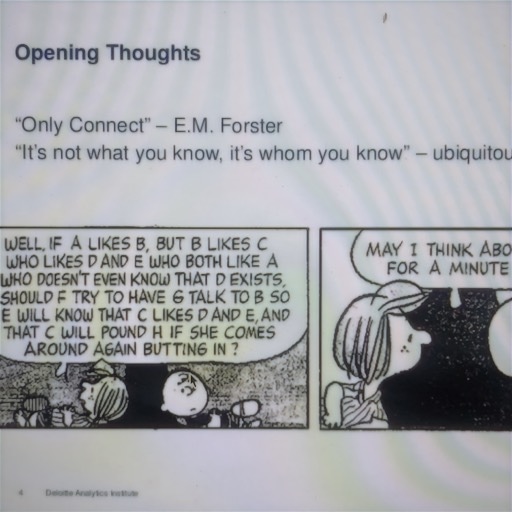}};
\spy [closeup,magnification=2] on ($(Fig1B)+(0,0)$) 
    in node[largewindow,anchor=north west] at ($(Fig1B.south west) + (0.04,0)$);

\node[anchor=south] (Fig1C) at (2,0) {\includegraphics[width=0.18\textwidth]{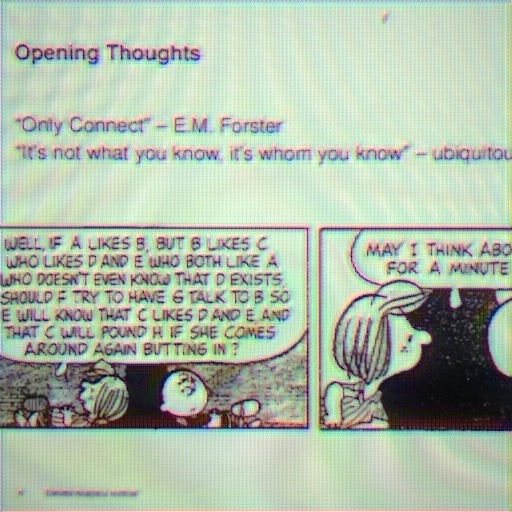}};
\spy [closeup,magnification=2] on ($(Fig1C)+(0,0)$) 
    in node[largewindow,anchor=north west] at ($(Fig1C.south west) + (0.04,0)$);
    
\node[anchor=south] (Fig1D) at (3,0) {\includegraphics[width=0.18\textwidth]{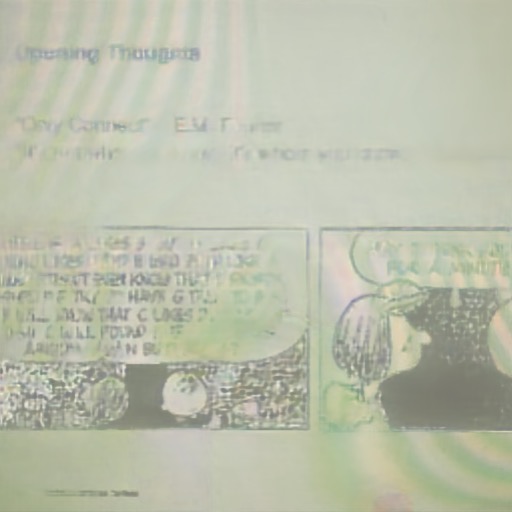}};
\spy [closeup,magnification=2] on ($(Fig1D)+(0,0)$) 
    in node[largewindow,anchor=north west] at ($(Fig1D.south west) + (0.04,0)$);

\node[anchor=south] (Fig1E) at (4,0) {\includegraphics[width=0.18\textwidth]{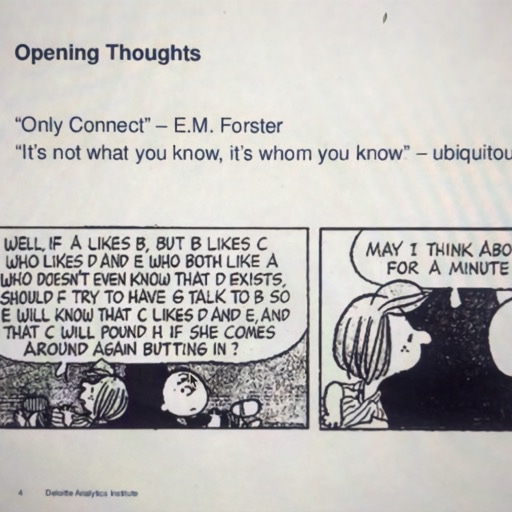}};
\spy [closeup,magnification=2] on ($(Fig1E)+(0,0)$) 
    in node[largewindow,anchor=north west] at ($(Fig1E.south west) + (0.04,0)$);
\end{tikzpicture}

\begin{tikzpicture}[x=0.19\textwidth, y=0.2\textheight, spy using outlines={every spy on node/.append style={smallwindow}}]
\node[anchor=south] (Fig1A) at (0,0) {\includegraphics[width=0.18\textwidth]{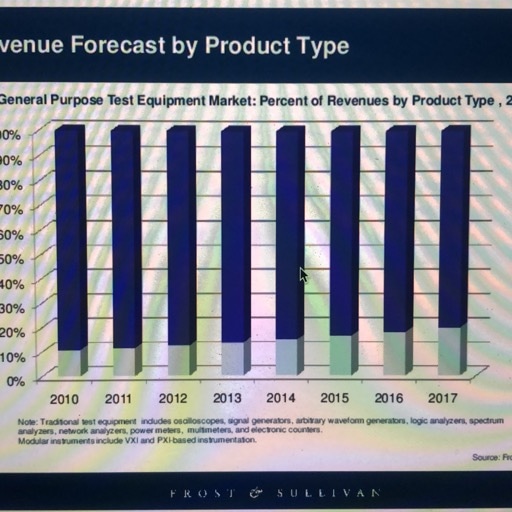}};
\spy [closeup,magnification=2] on ($(Fig1A)+(+0.15,+0.15)$) 
    in node[largewindow,anchor=north west] at ($(Fig1A.south west) + (0.04,0)$);
\node [anchor=north] at ($(Fig1A.south)+(0,-0.23)$) {\small Input};

\node[anchor=south] (Fig1B) at (1,0) {\includegraphics[width=0.18\textwidth]{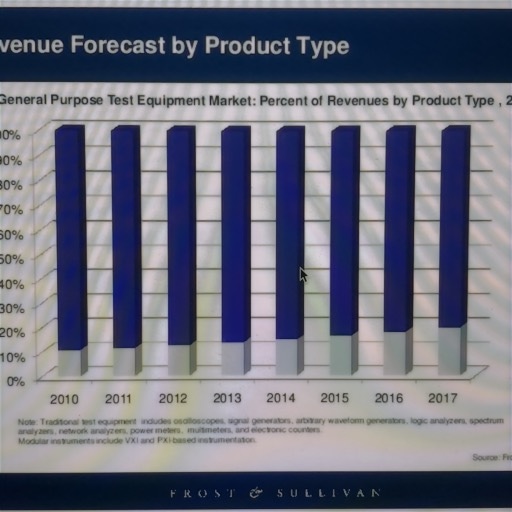}};
\spy [closeup,magnification=2] on ($(Fig1B)+(+0.15,+0.15)$) 
    in node[largewindow,anchor=north west] at ($(Fig1B.south west) + (0.04,0)$);
\node [anchor=north] at ($(Fig1B.south)+(0,-0.23)$) {\small AFN~\cite{Xu:2020:AFN}};

\node[anchor=south] (Fig1C) at (2,0) {\includegraphics[width=0.18\textwidth]{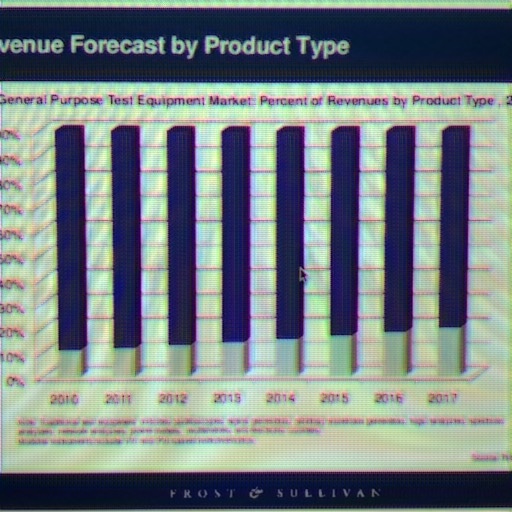}};
\spy [closeup,magnification=2] on ($(Fig1C)+(+0.15,+0.15)$) 
    in node[largewindow,anchor=north west] at ($(Fig1C.south west) + (0.04,0)$);
\node [anchor=north] at ($(Fig1C.south)+(0,-0.23)$) {\small C3Net~\cite{Kim:2020:C3Net}};
    
\node[anchor=south] (Fig1D) at (3,0) {\includegraphics[width=0.18\textwidth]{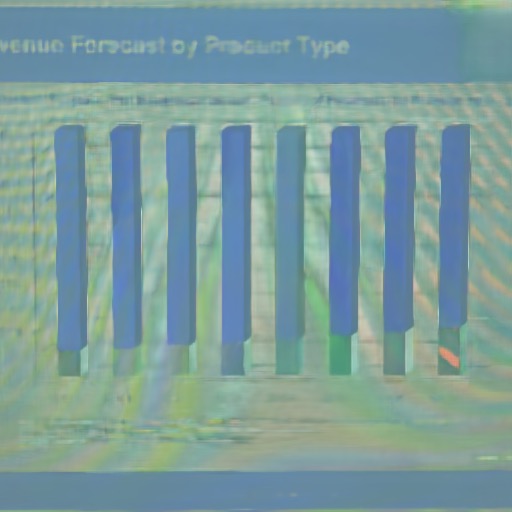}};
\spy [closeup,magnification=2] on ($(Fig1D)+(+0.15,+0.15)$) 
    in node[largewindow,anchor=north west] at ($(Fig1D.south west) + (0.04,0)$);
\node [anchor=north] at ($(Fig1D.south)+(0,-0.23)$) {\small Double DIP~\cite{Gandelsman:2019:DoubleDIP}};
    
\node[anchor=south] (Fig1E) at (4,0) {\includegraphics[width=0.18\textwidth]{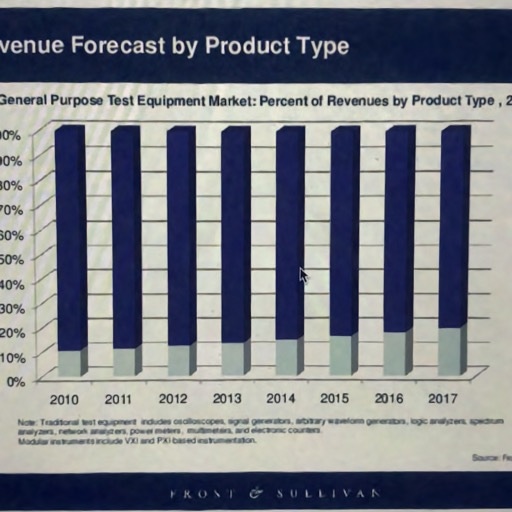}};
\spy [closeup,magnification=2] on ($(Fig1E)+(+0.15,+0.15)$) 
    in node[largewindow,anchor=north west] at ($(Fig1E.south west) + (0.04,0)$);
\node [anchor=north] at ($(Fig1E.south)+(0,-0.23)$) {\small Ours};
\end{tikzpicture}
\caption{Qualitative comparison of moir\'e removal on real images.}
\label{fig:supp_moire}
\end{figure*}

%% file: figures/supp_reflection.tex
\begin{figure*}
    \centering
    \setlength{\tabcolsep}{1.1pt}
    \begin{tabular}{cccccc}
    
    \includegraphics[width=0.16\textwidth]{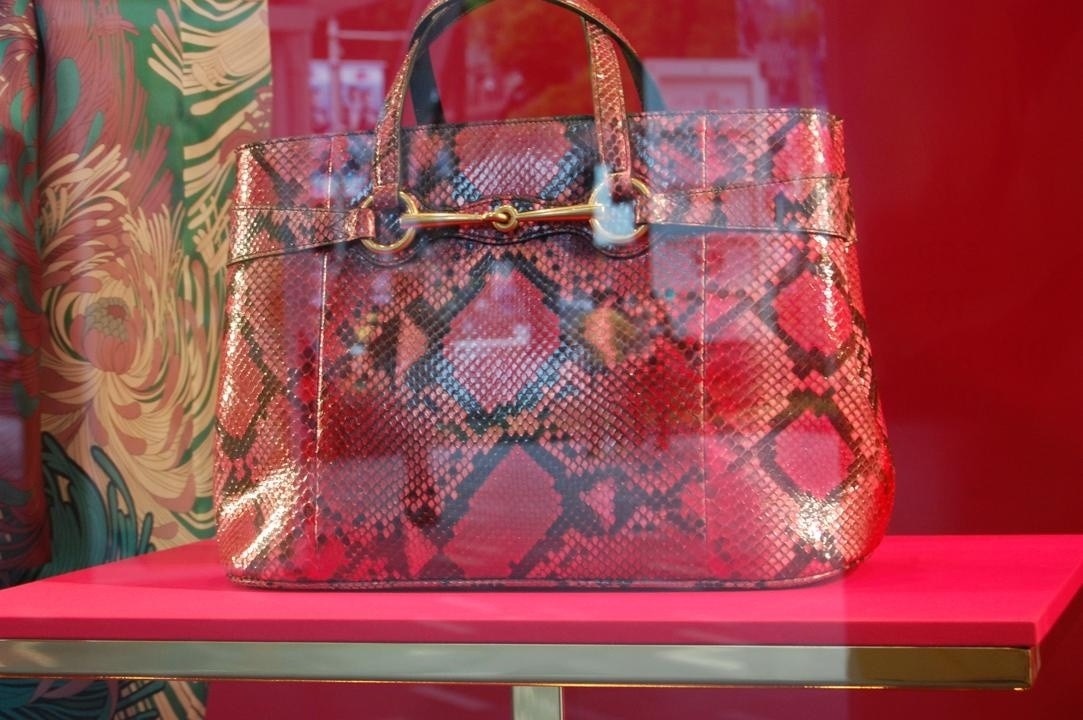} & \includegraphics[width=0.16\textwidth]{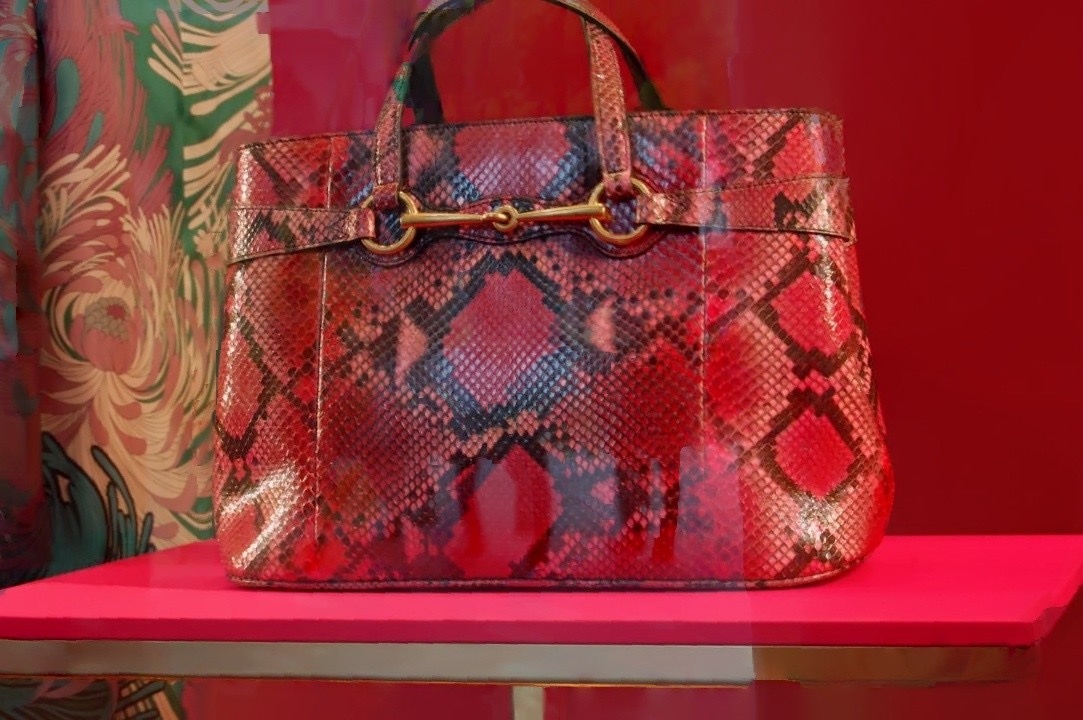} & \includegraphics[width=0.16\textwidth]{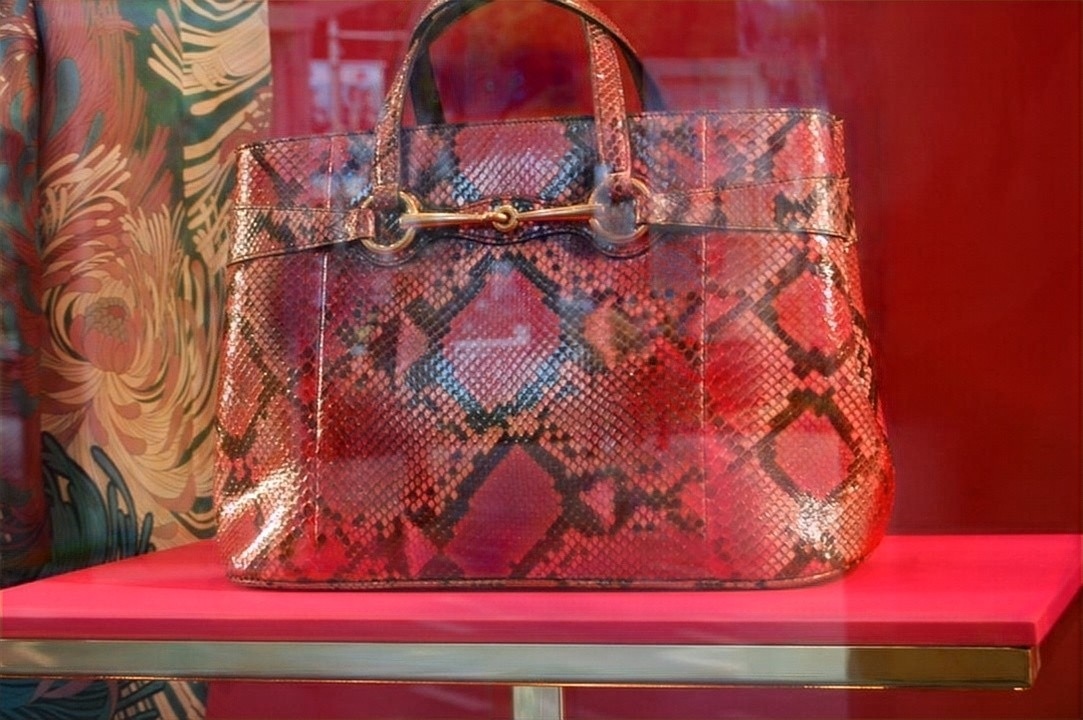} & \includegraphics[width=0.16\textwidth]{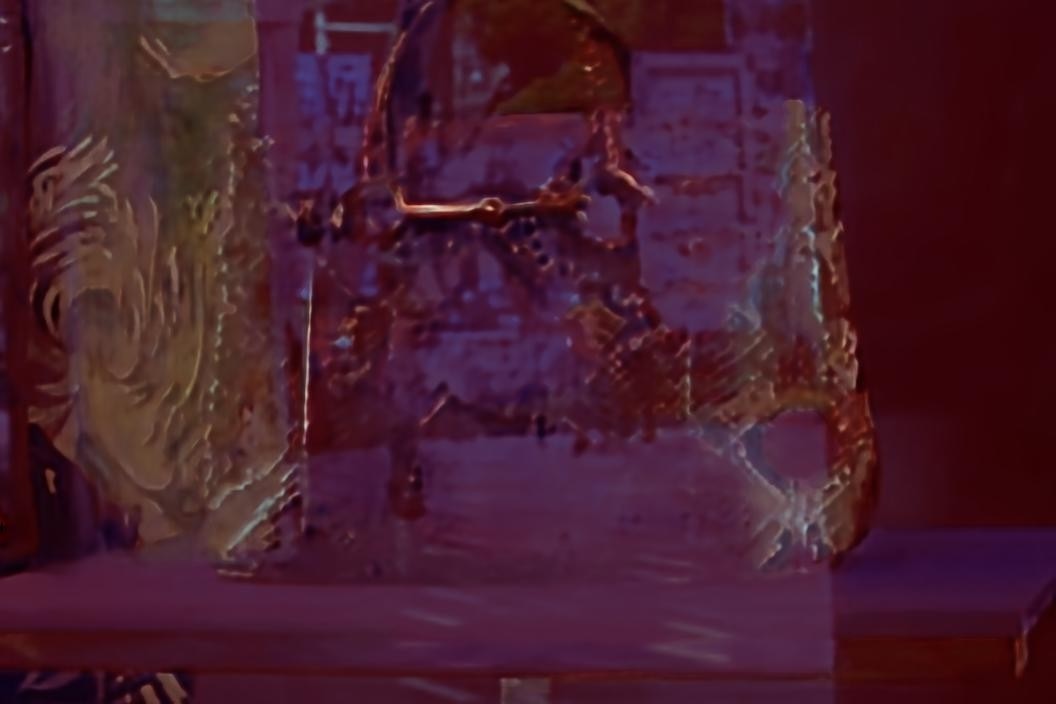} & \includegraphics[width=0.16\textwidth]{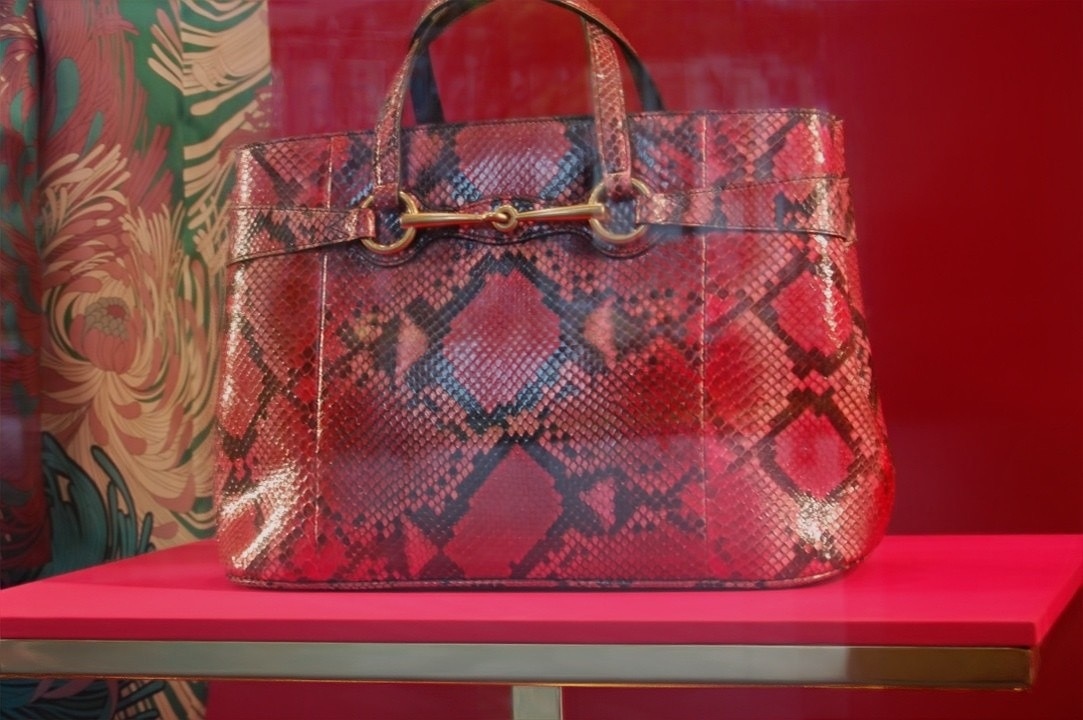} & \includegraphics[width=0.16\textwidth]{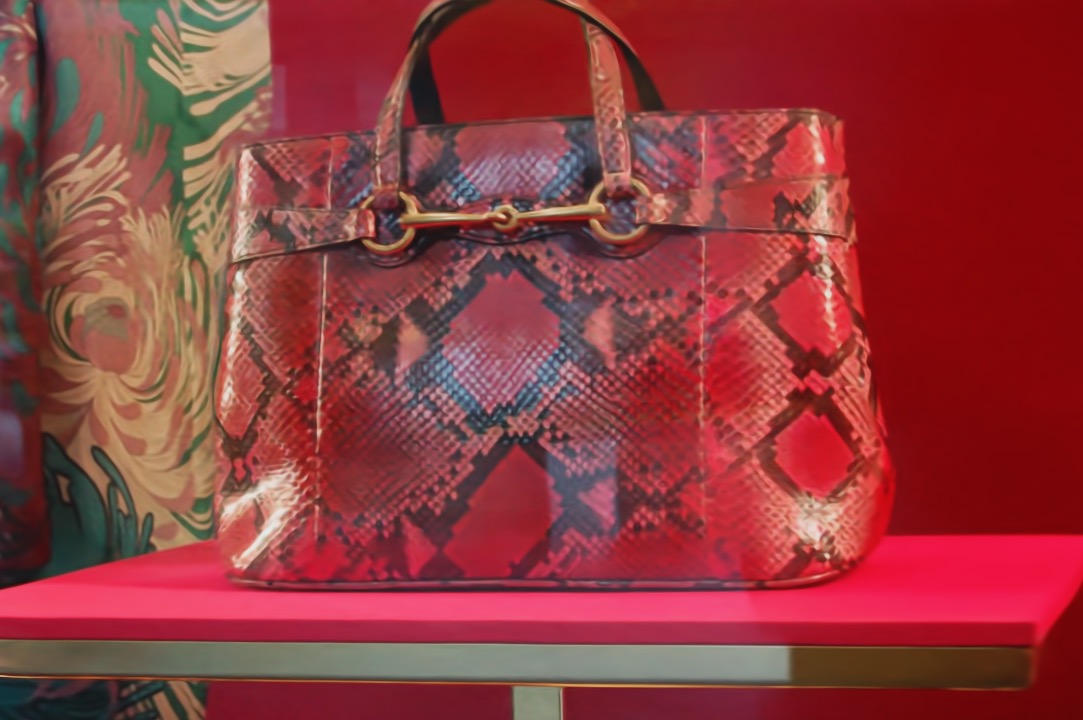} \\
    & \includegraphics[width=0.16\textwidth]{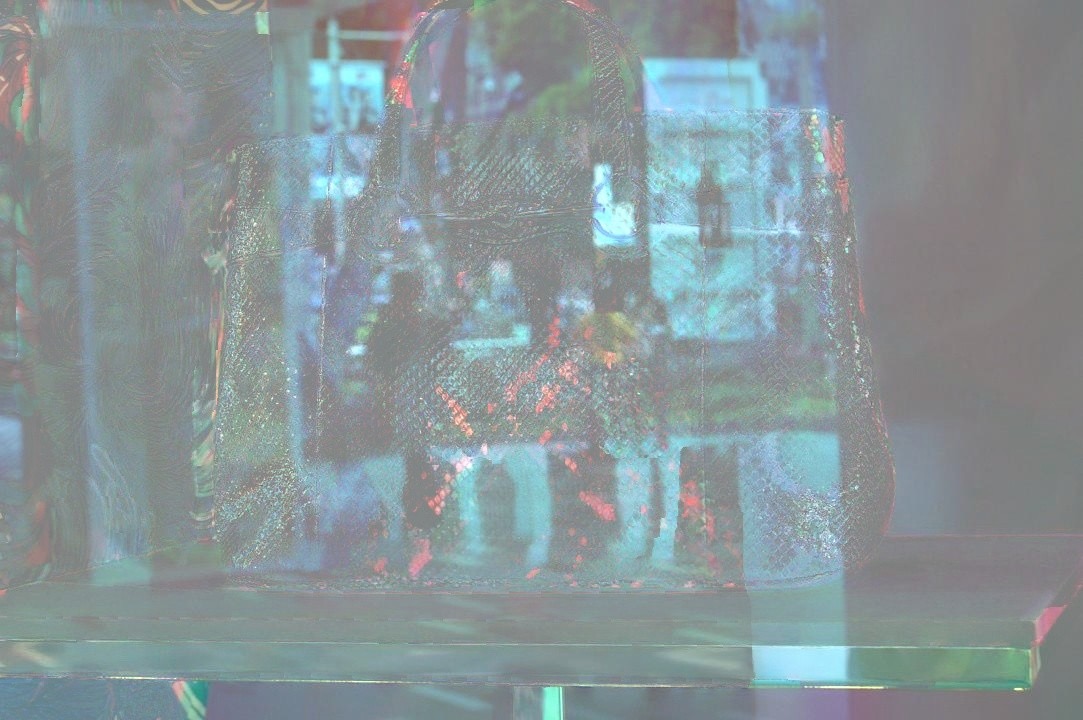} & \includegraphics[width=0.16\textwidth]{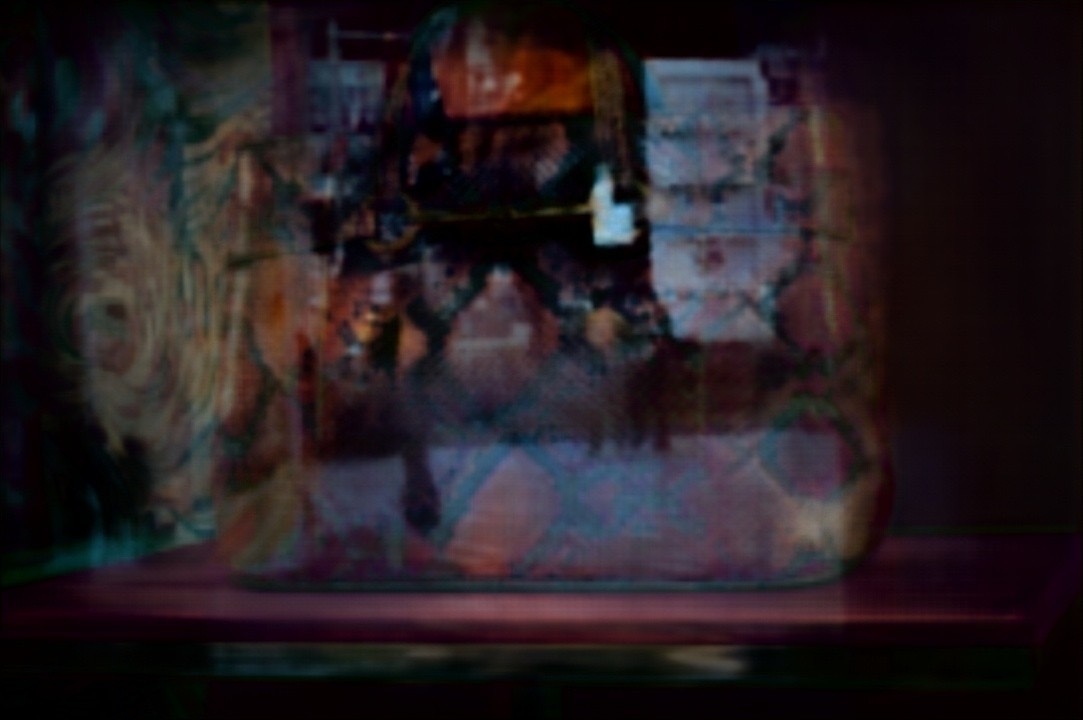} & \includegraphics[width=0.16\textwidth]{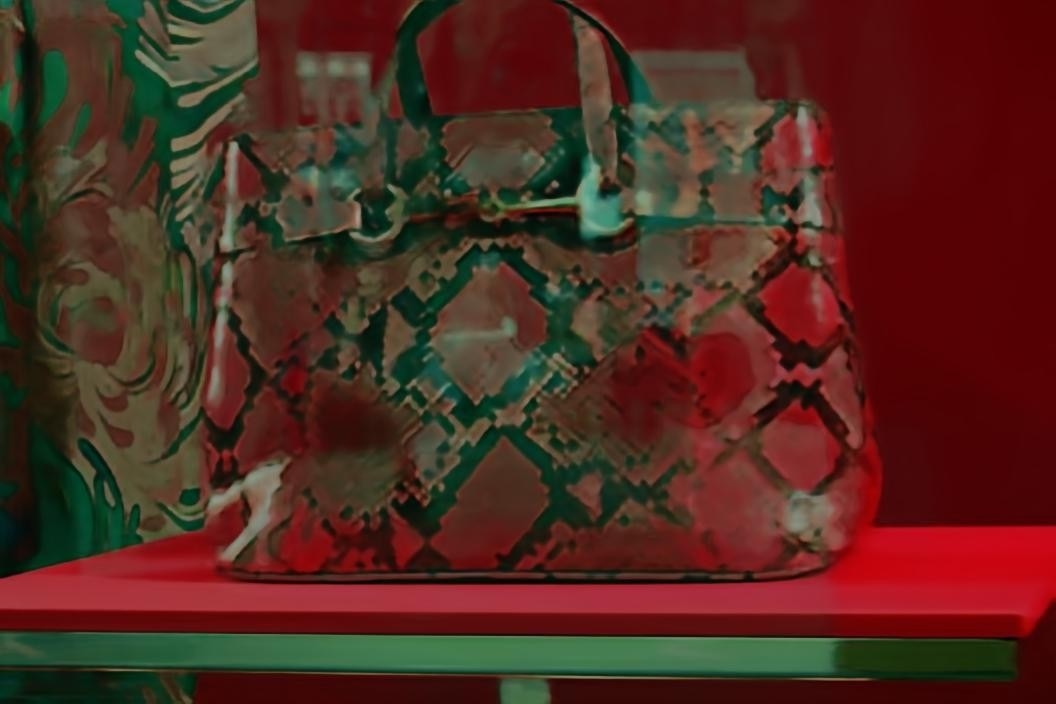} & \includegraphics[width=0.16\textwidth]{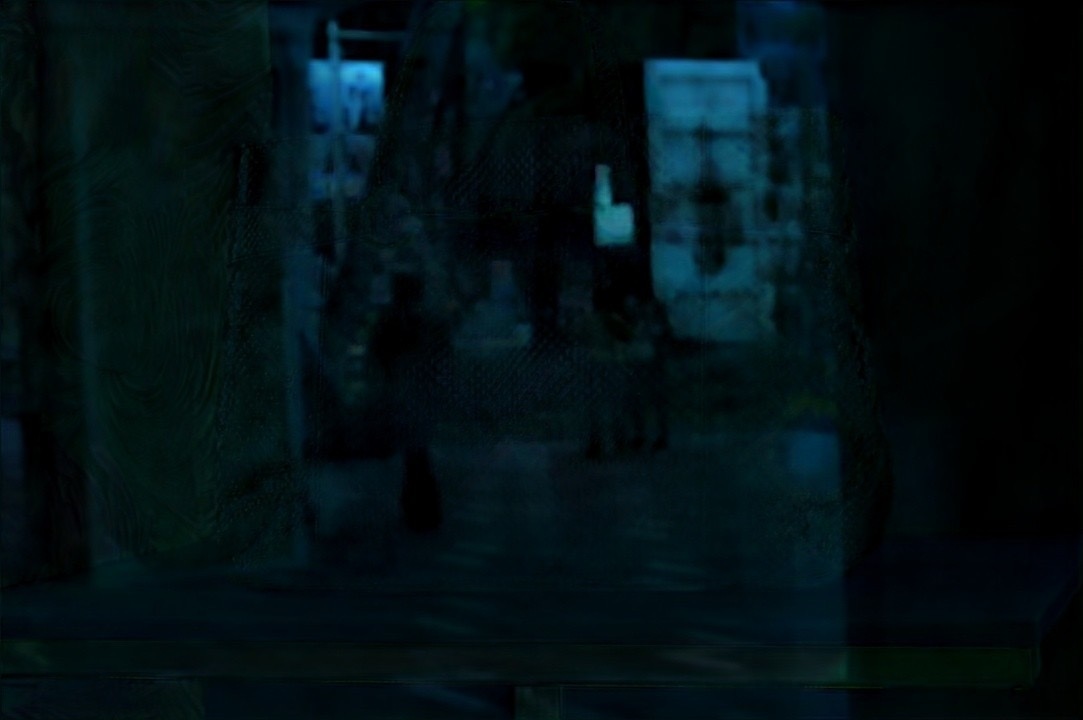} & \includegraphics[width=0.16\textwidth]{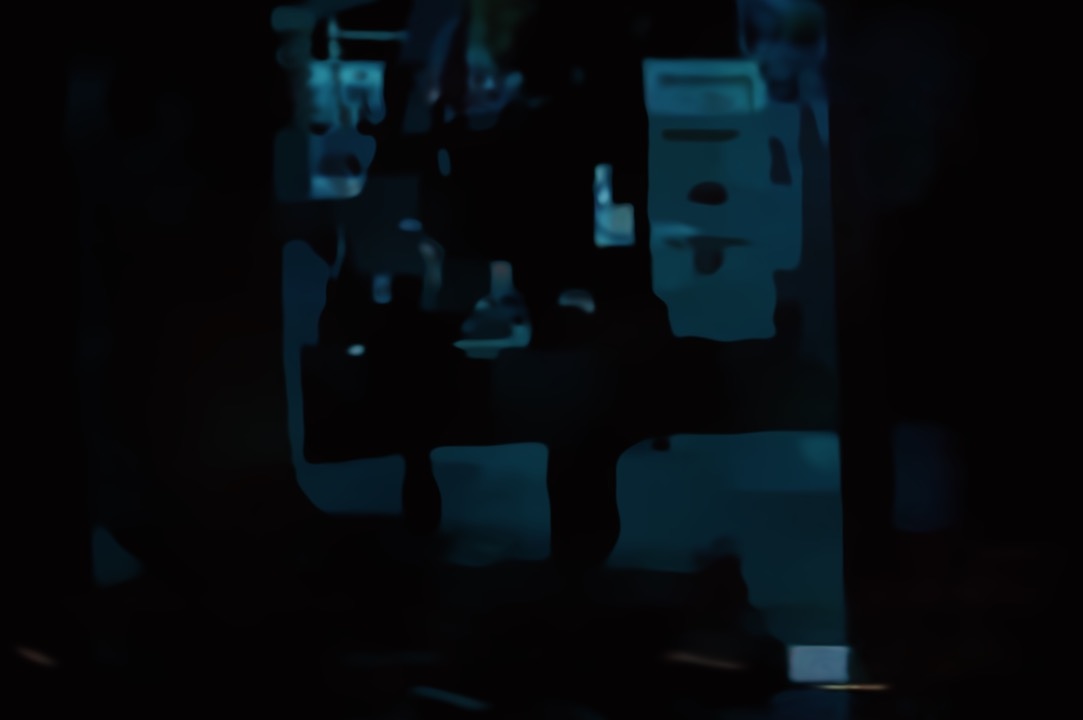} \\
    
    \includegraphics[width=0.16\textwidth]{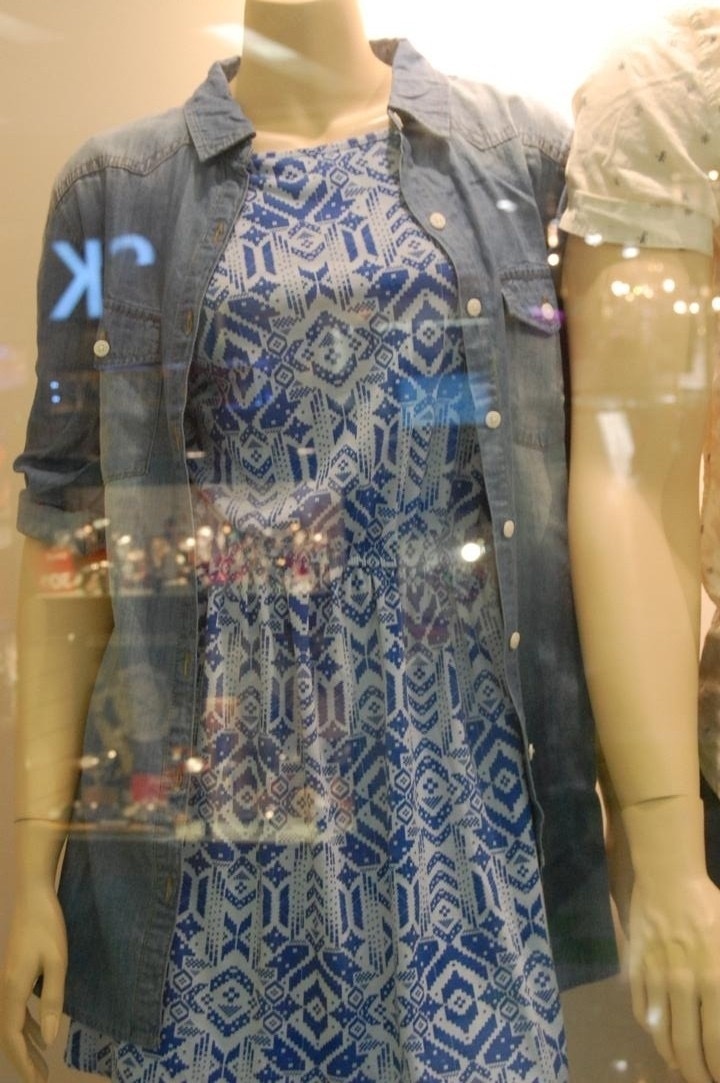} & \includegraphics[width=0.16\textwidth]{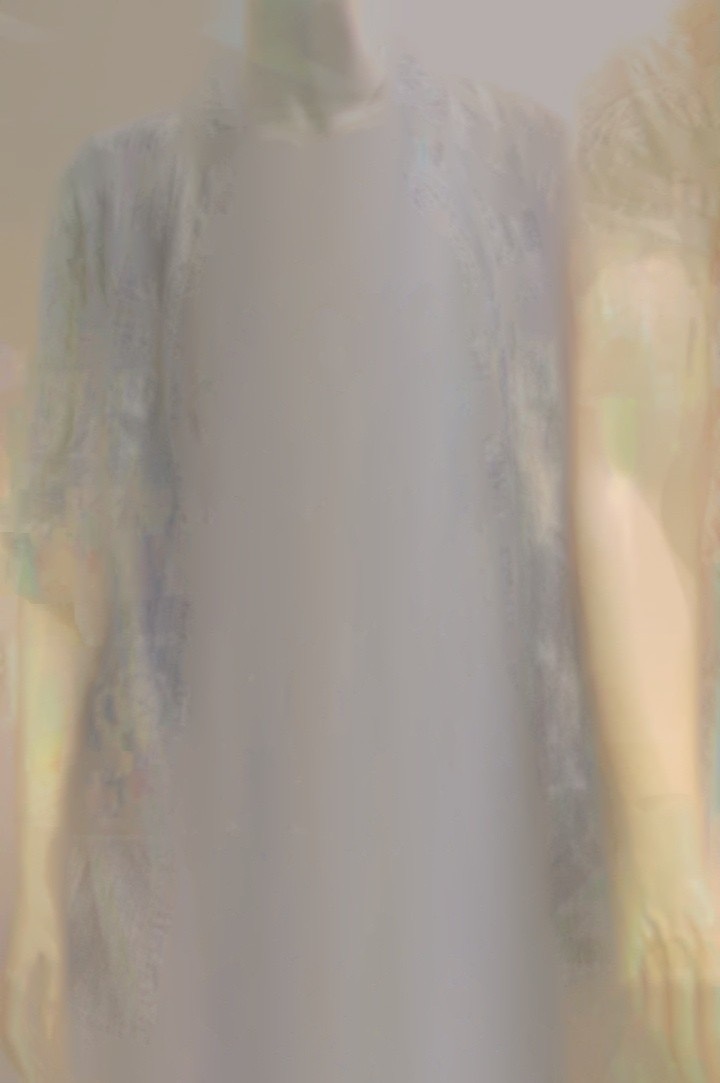} & \includegraphics[width=0.16\textwidth]{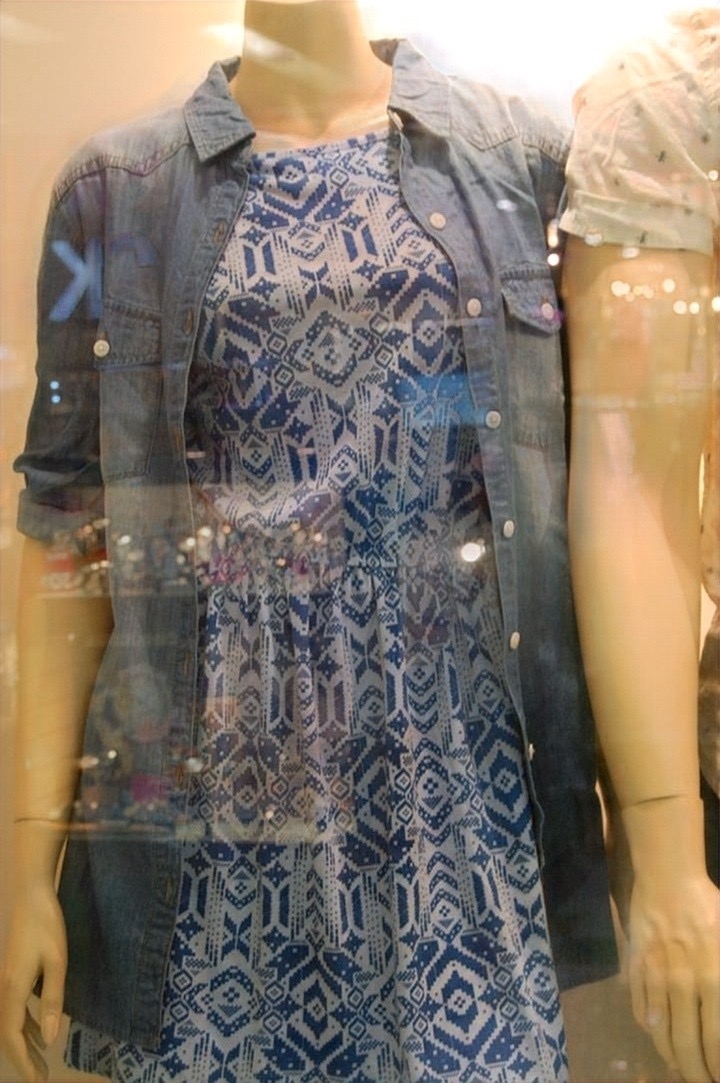} & \includegraphics[width=0.16\textwidth]{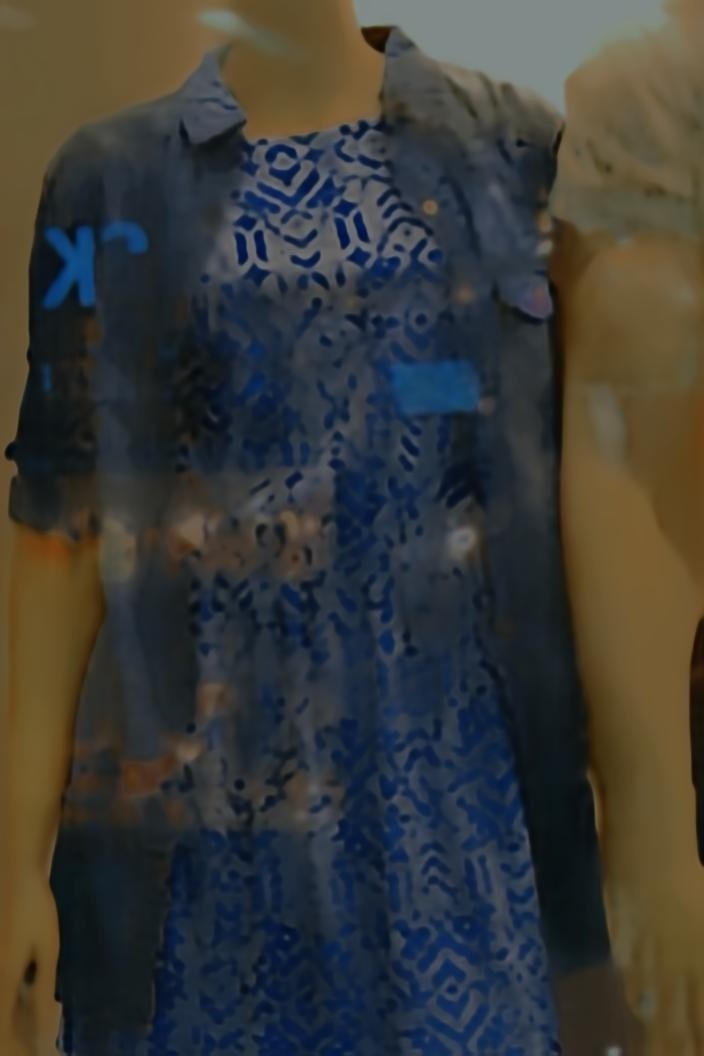} & \includegraphics[width=0.16\textwidth]{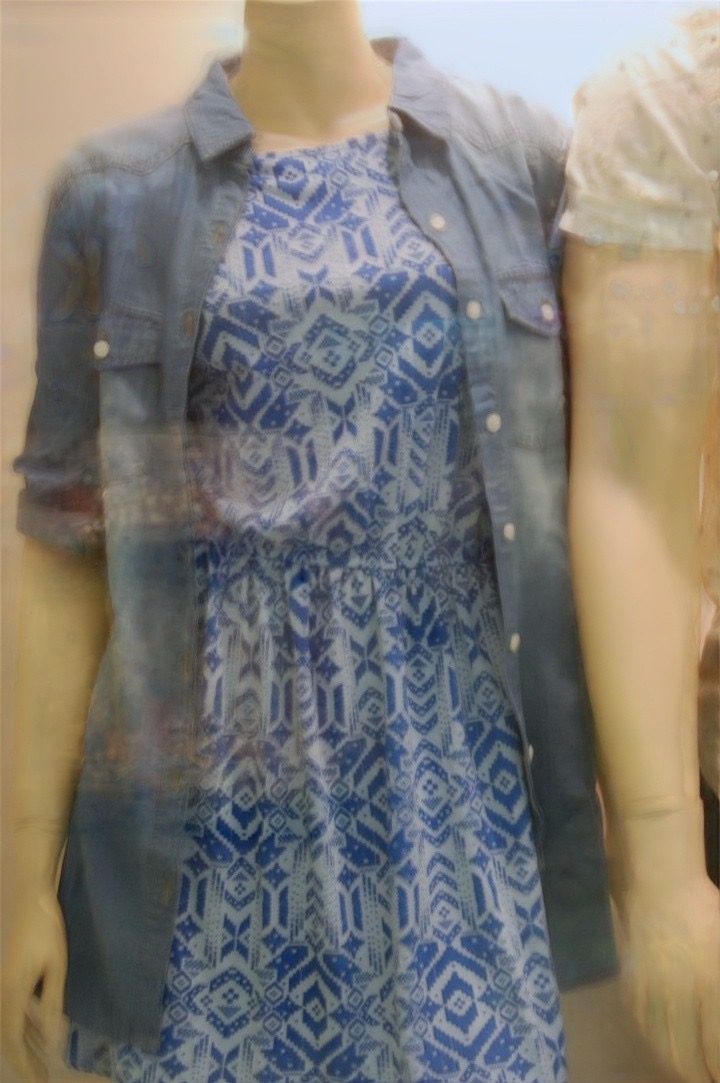} & \includegraphics[width=0.16\textwidth]{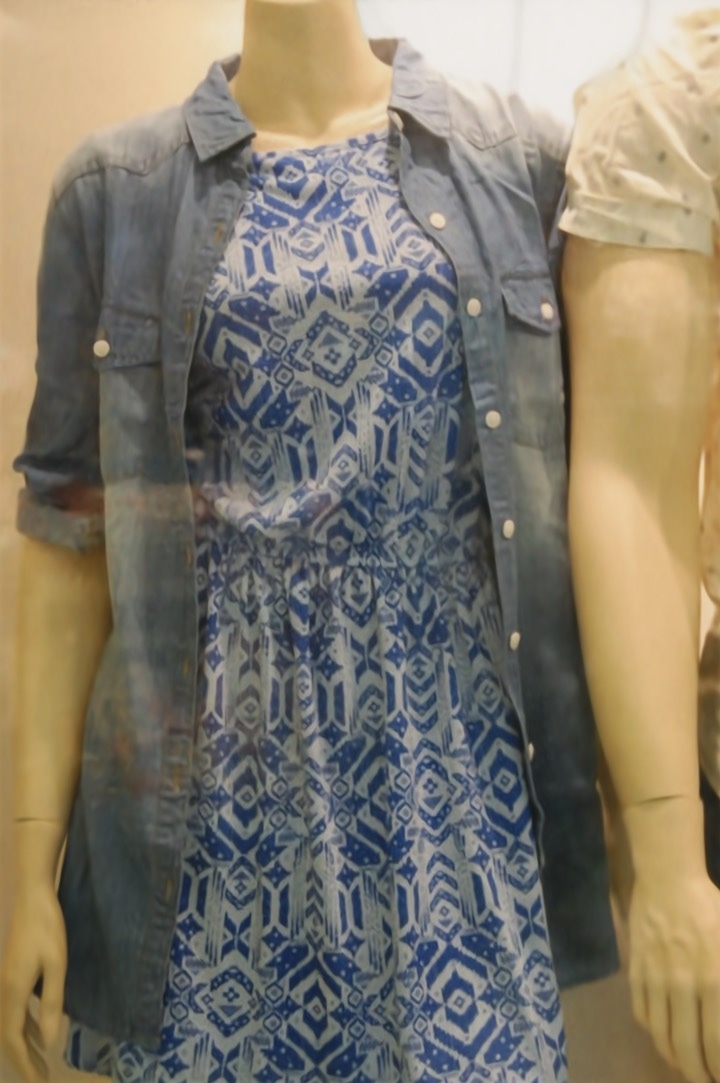} \\
    & \includegraphics[width=0.16\textwidth]{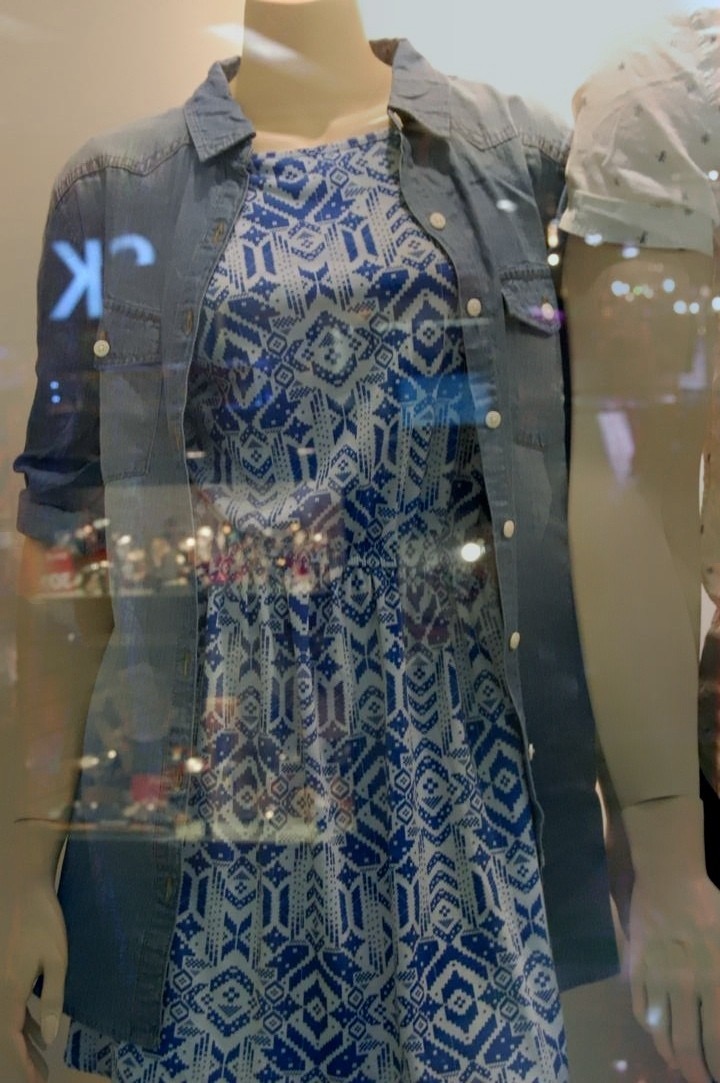} & \includegraphics[width=0.16\textwidth]{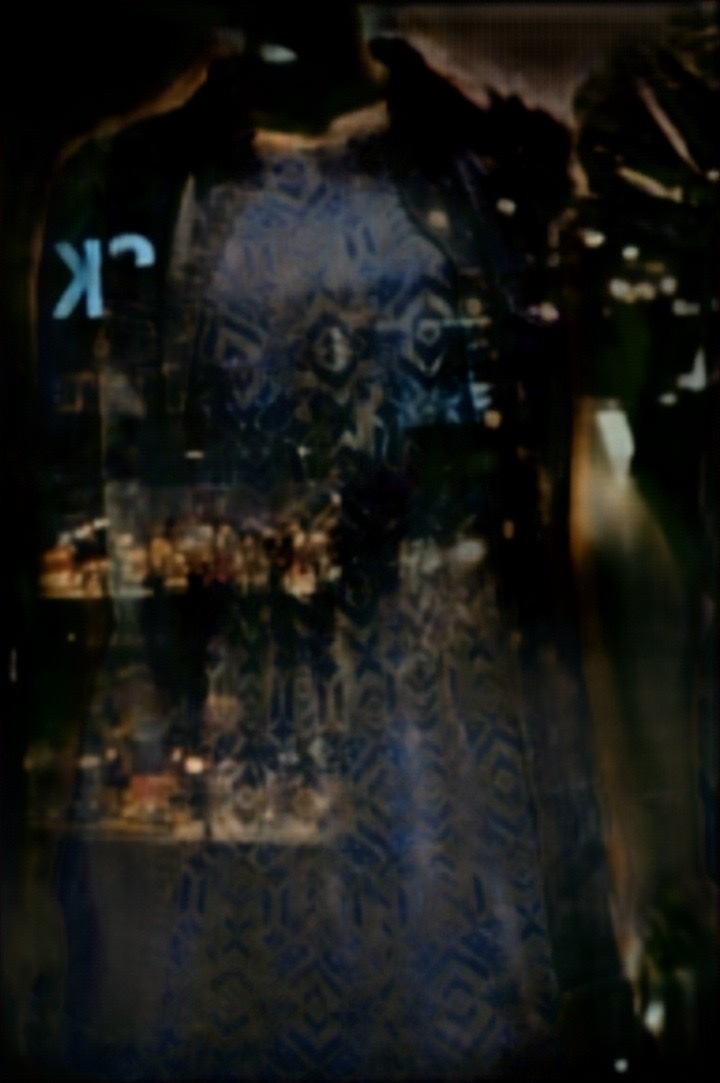} & \includegraphics[width=0.16\textwidth]{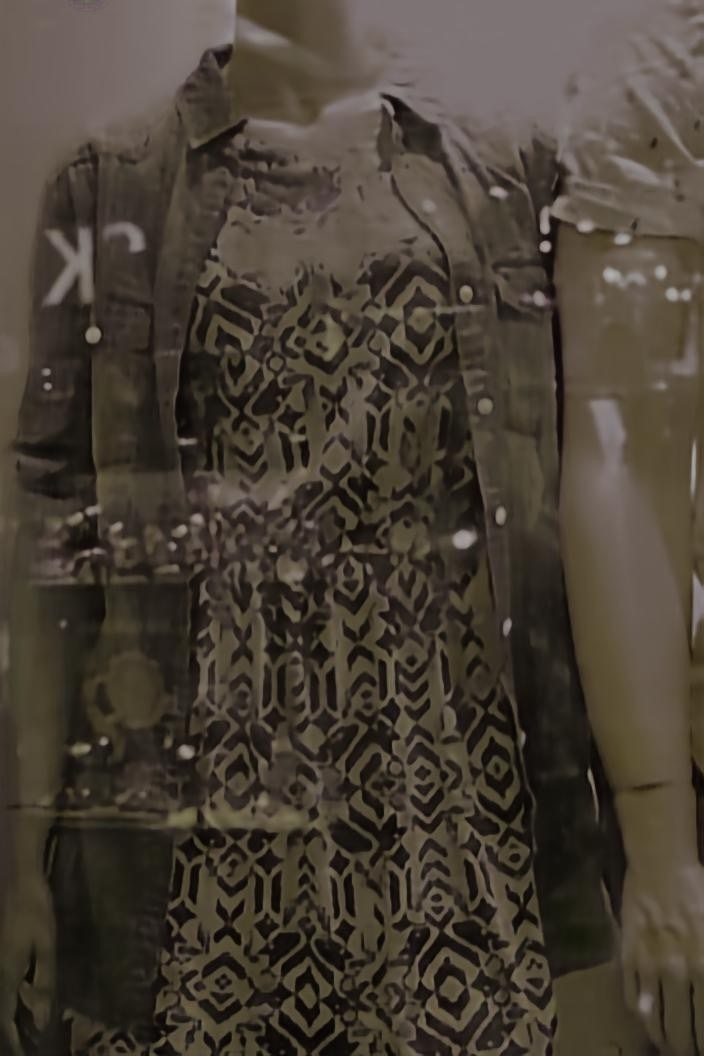} & \includegraphics[width=0.16\textwidth]{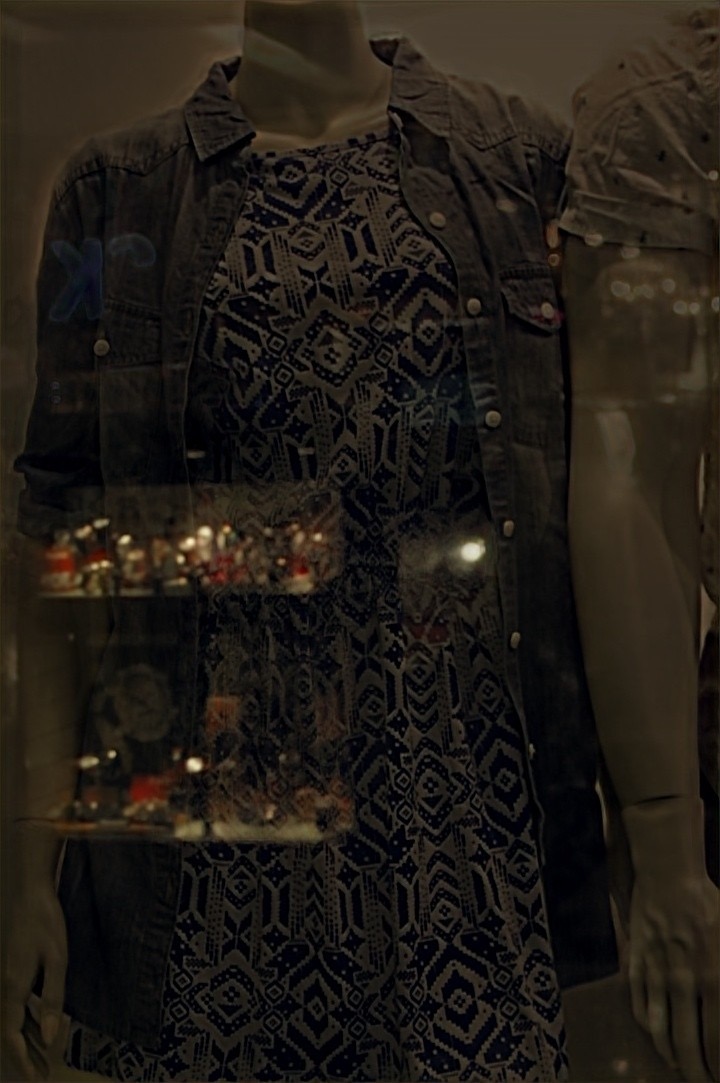} & \includegraphics[width=0.16\textwidth]{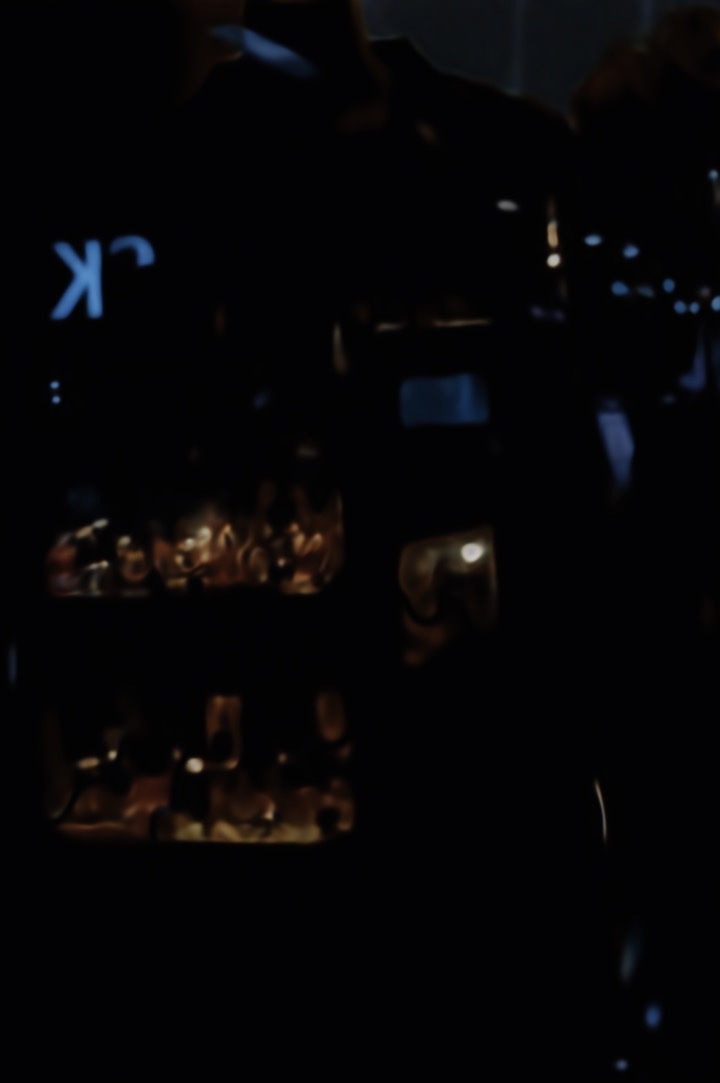} \\
    
    \includegraphics[width=0.16\textwidth]{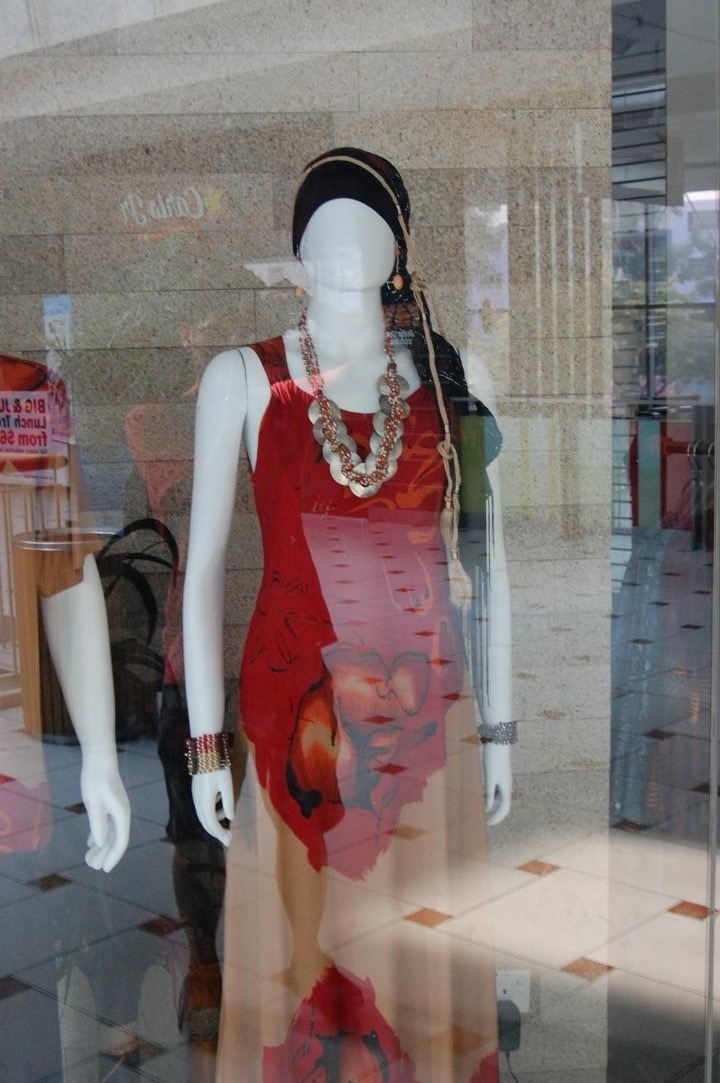} & \includegraphics[width=0.16\textwidth]{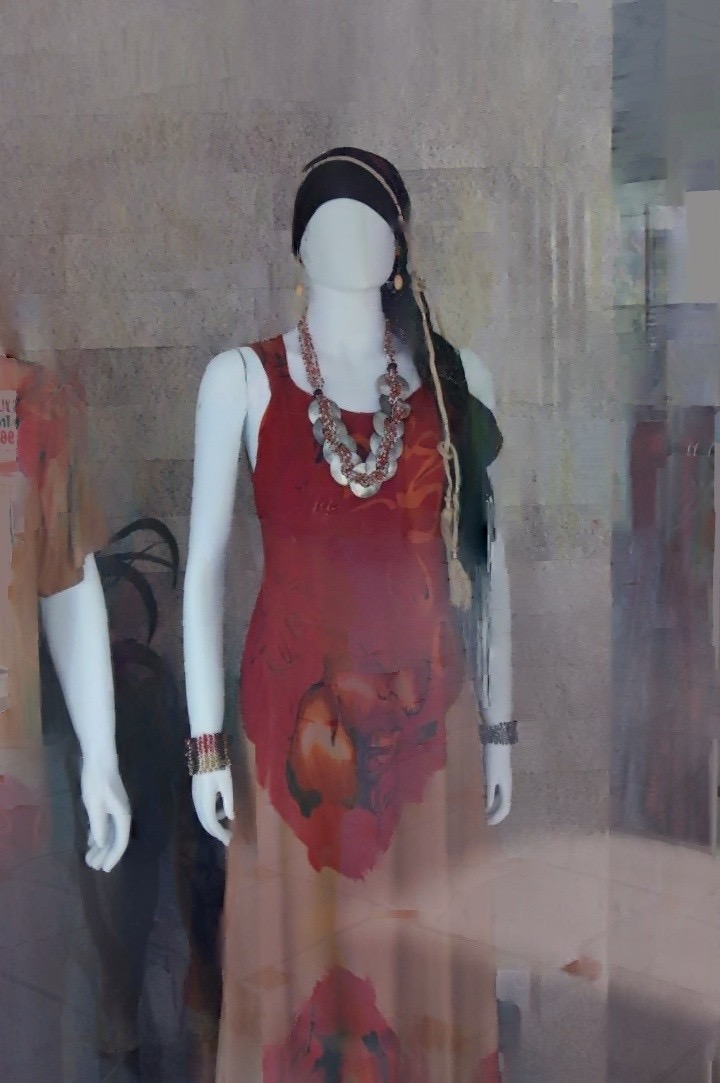} & \includegraphics[width=0.16\textwidth]{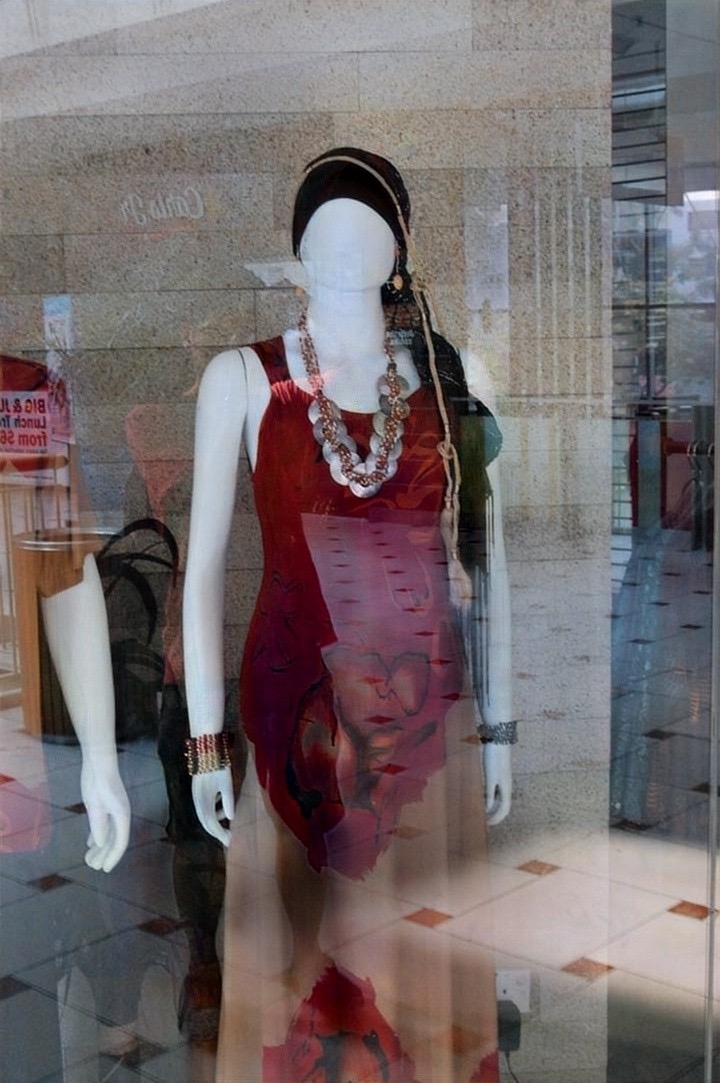} & \includegraphics[width=0.16\textwidth]{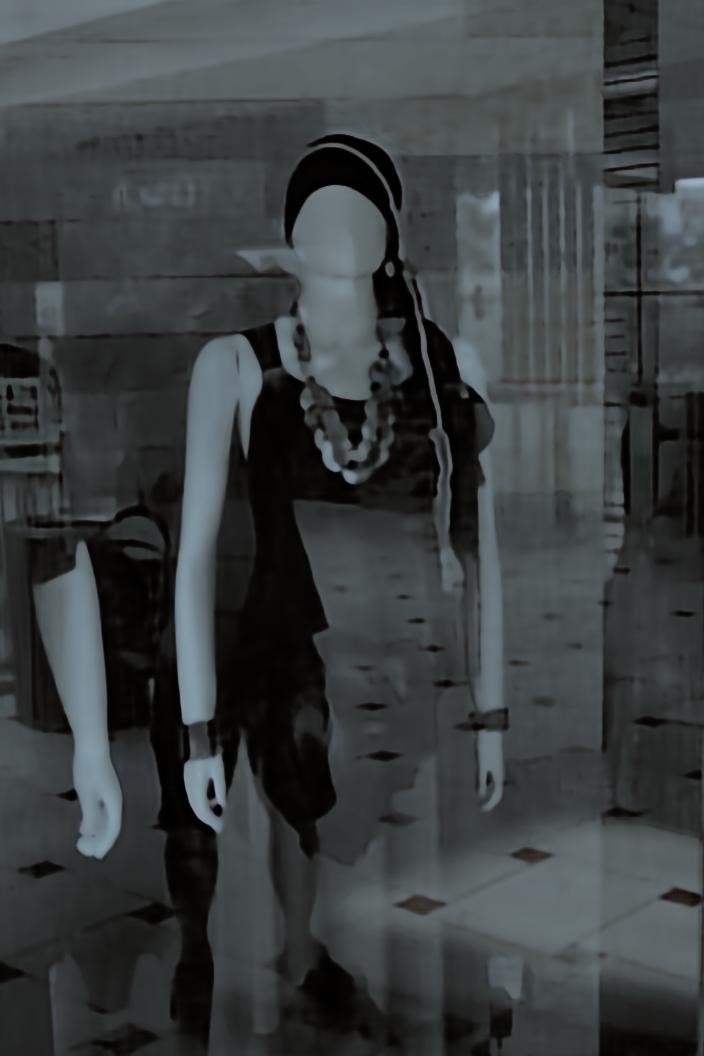} & \includegraphics[width=0.16\textwidth]{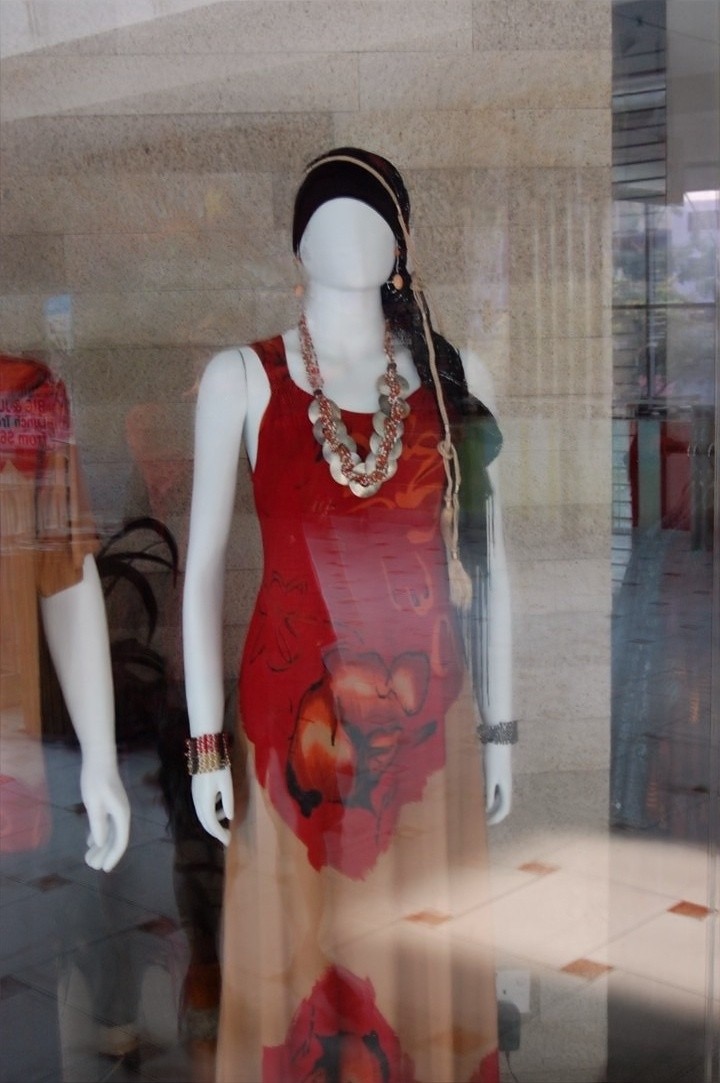} & \includegraphics[width=0.16\textwidth]{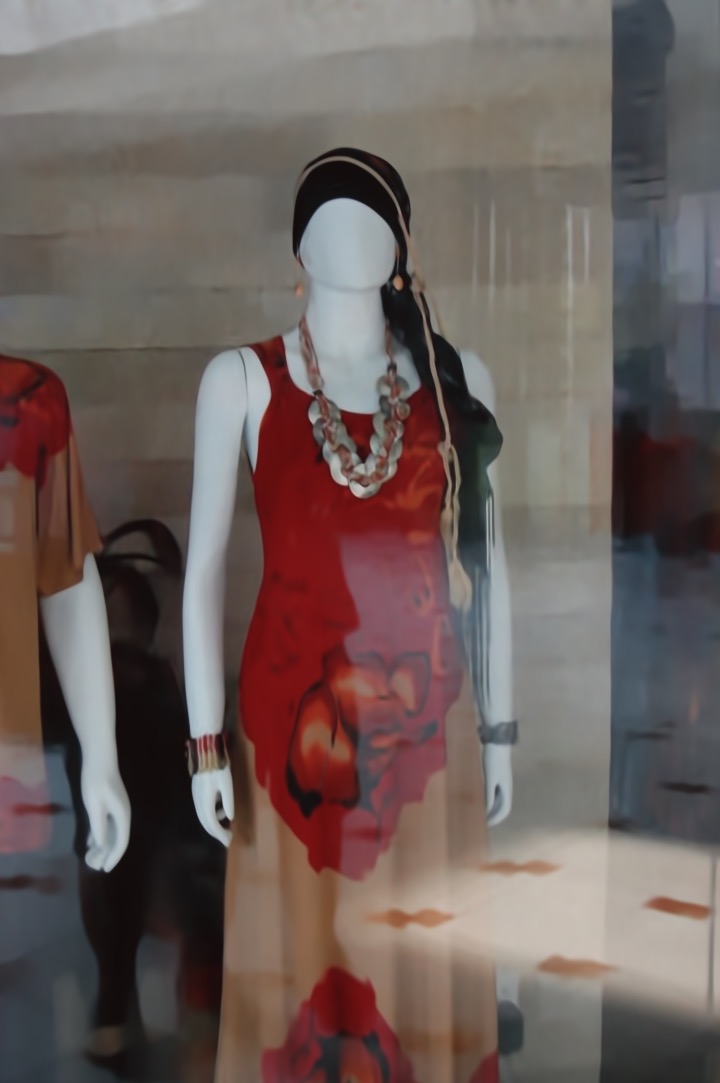} \\
    & \includegraphics[width=0.16\textwidth]{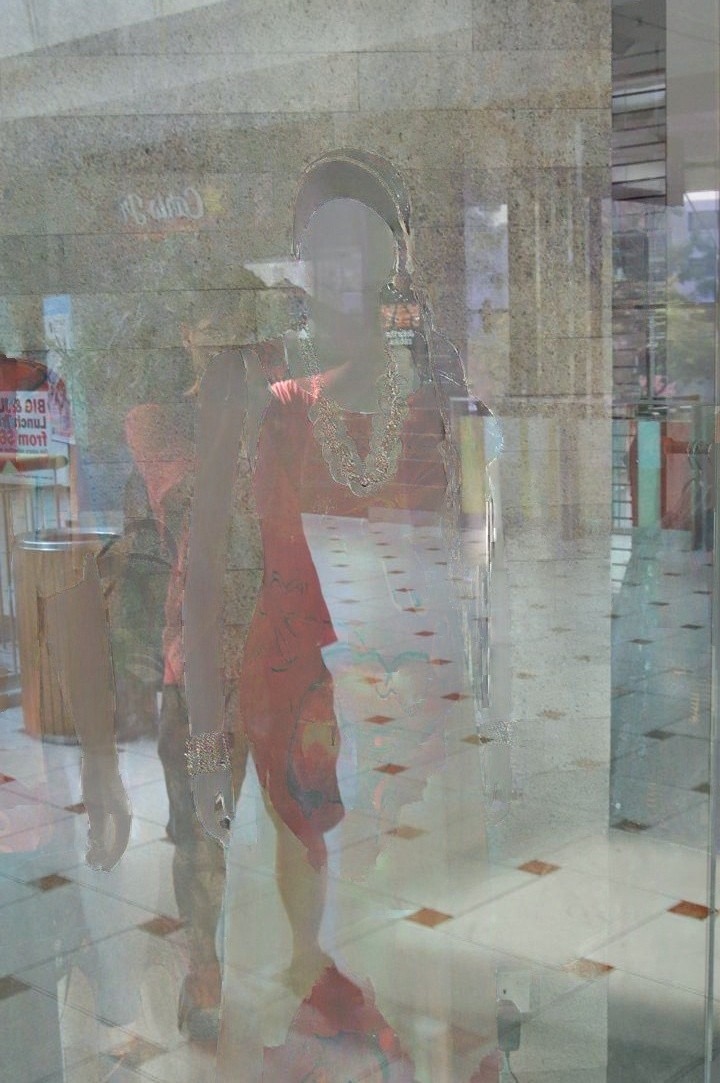} & \includegraphics[width=0.16\textwidth]{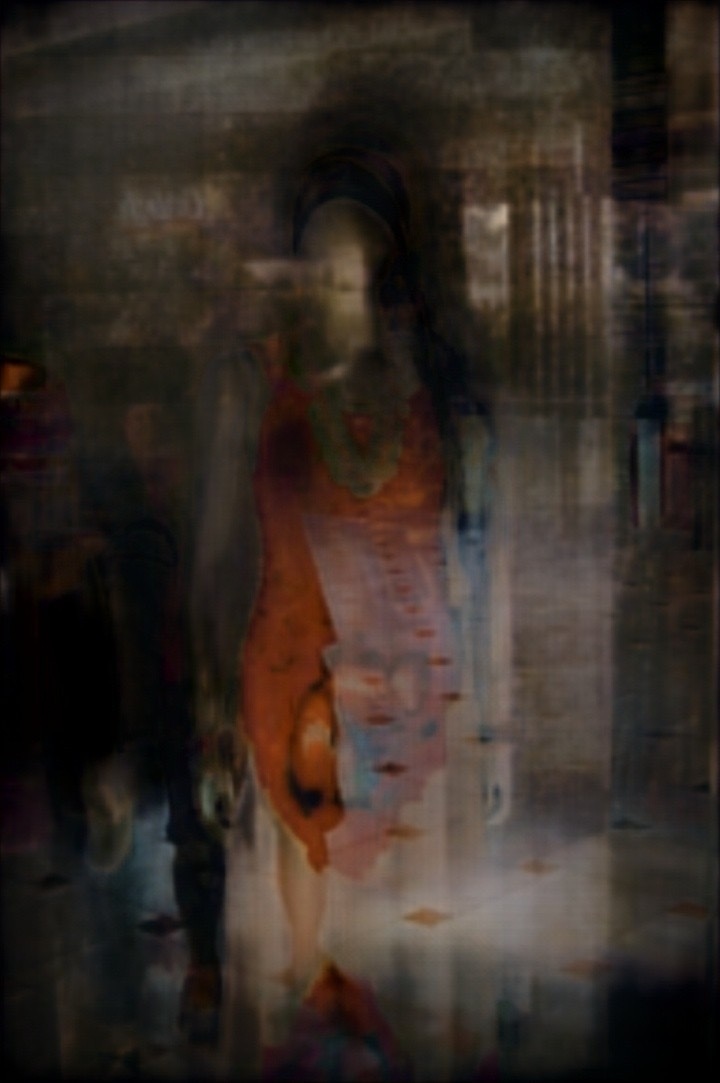} & \includegraphics[width=0.16\textwidth]{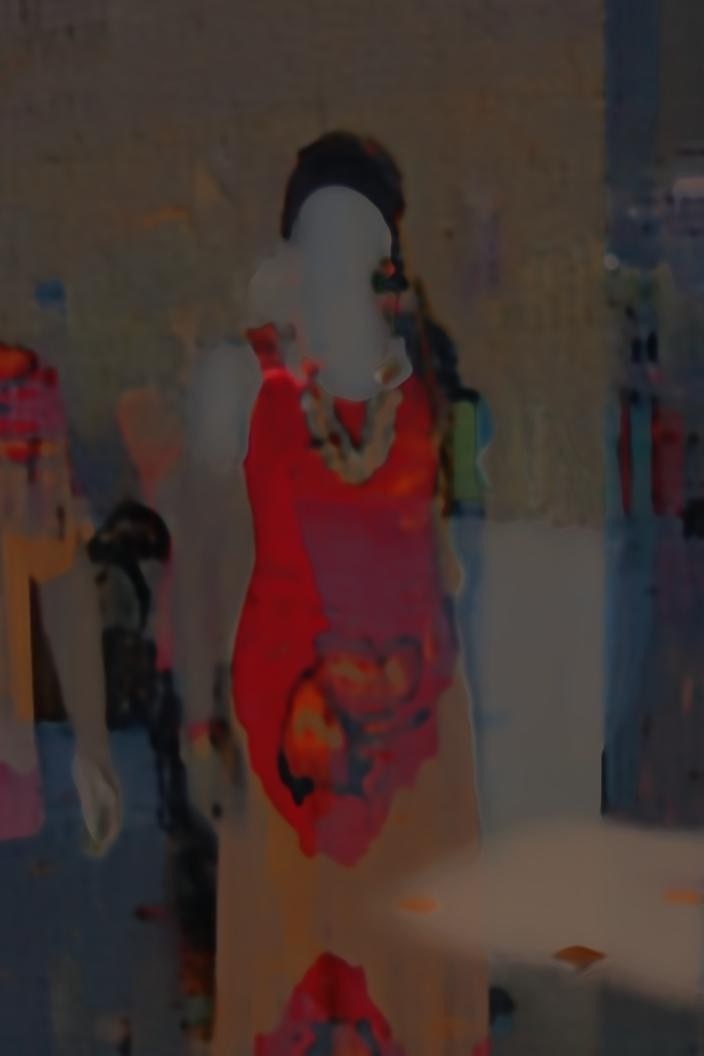} & \includegraphics[width=0.16\textwidth]{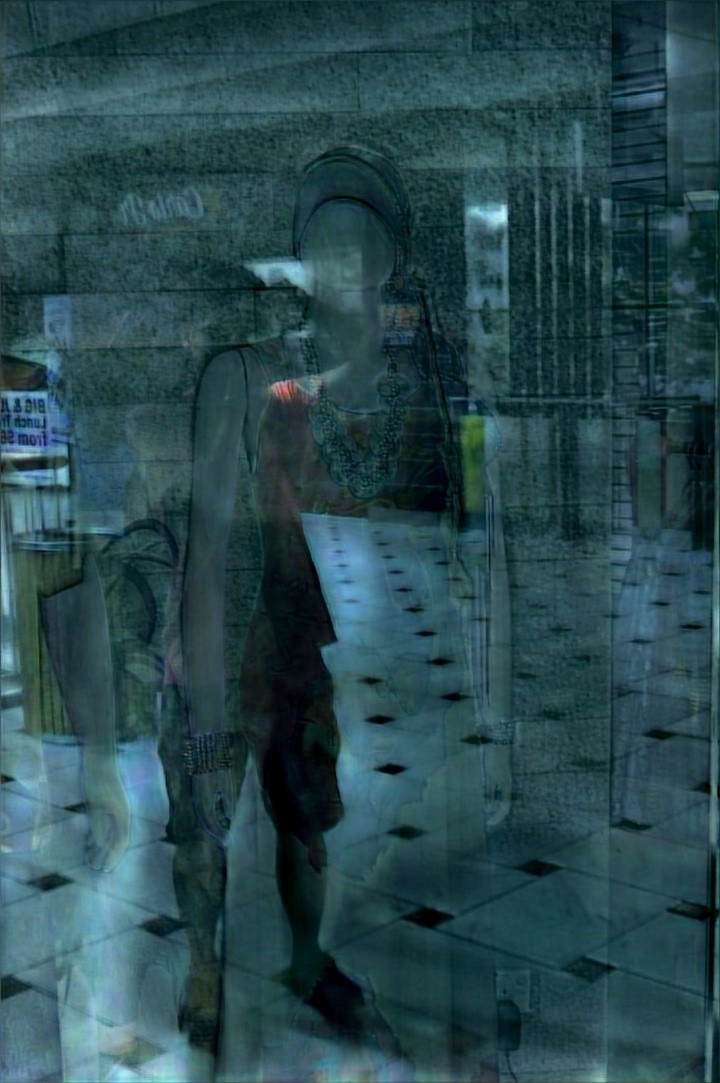} & \includegraphics[width=0.16\textwidth]{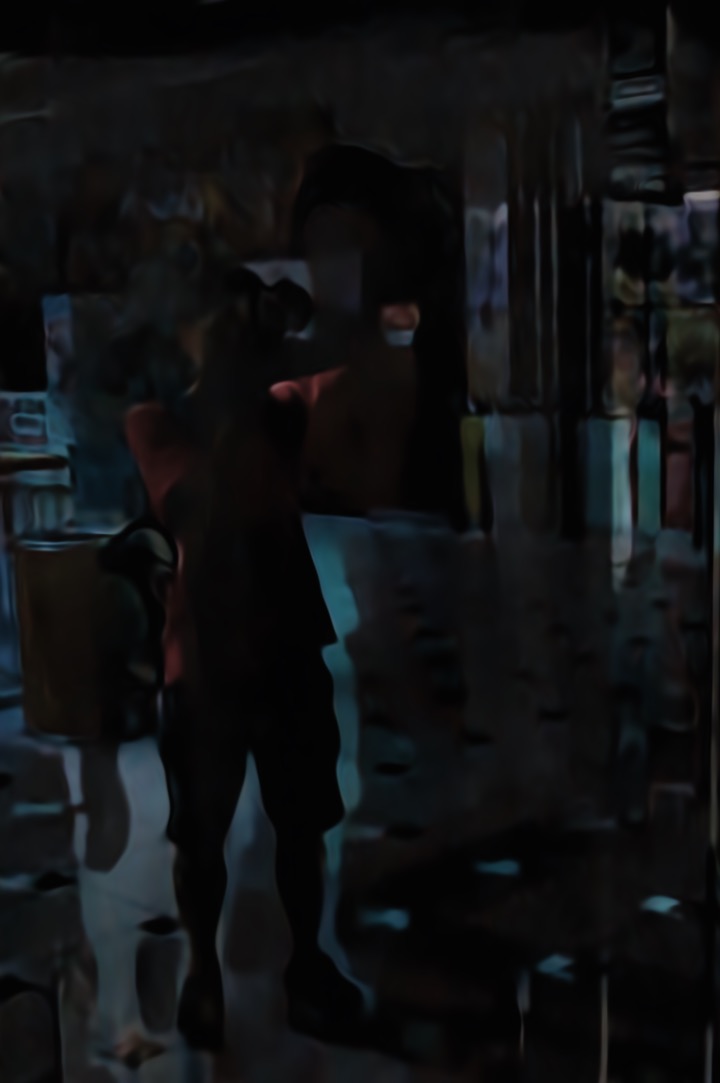} \\

    \multirow{2}{*}{\small Input} & {\small Li and Brown} & {\small Alayrac~\etal} & {\small Double DIP} & {\small Liu~\etal} & \multirow{2}{*}{\small Ours} \\
    & {\small \cite{Li:2013:LiandBrown}} & {\small \cite{Alayrac:2019:VisualCentrifuge}} & {\small \cite{Gandelsman:2019:DoubleDIP}} & {\small \cite{Liu:2020:LearningToSeeJournal}} &
    \end{tabular}
    \caption{Qualitative results of reflection removal on real images in~\cite{Li:2013:LiandBrown}.}
    \label{fig:supp_reflection}
\end{figure*}

%% file: figures/supp_fence.tex
\tikzstyle{closeup_fence} = [
  opacity=1.0,          
  height=0.9cm,         
  width=0.226\textwidth, 
  connect spies, red  
]

\begin{figure*}
\centering
\begin{tikzpicture}[x=0.24\textwidth, y=0.14\textheight, spy using outlines={every spy on node/.append style={smallwindow}}]
\node[anchor=south] (Fig1A) at (0,0) {\includegraphics[width=0.23\textwidth]{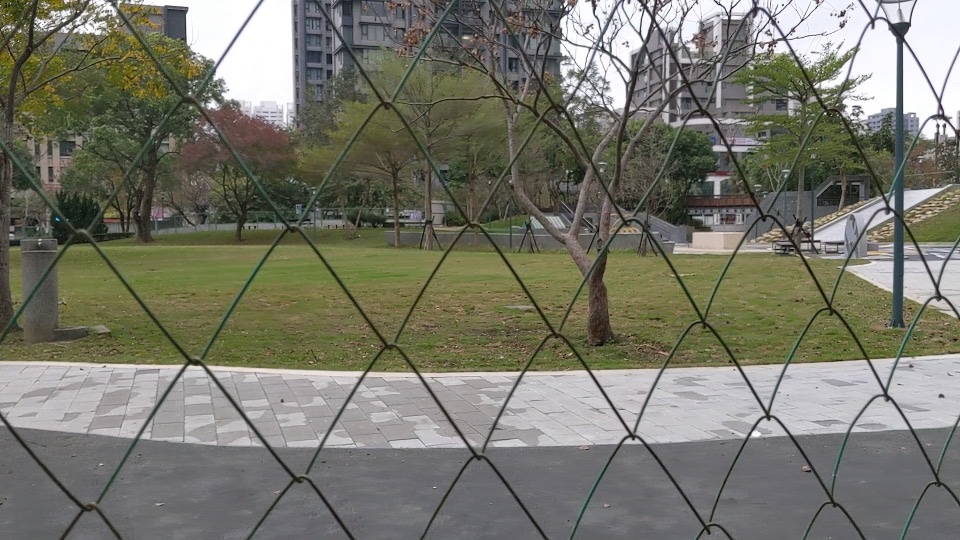}};
\spy [closeup_fence,magnification=2] on ($(Fig1A)+(0.24,0.17)$) 
    in node[largewindow,anchor=north west] at ($(Fig1A.south west) + (0.04,0)$);

\node[anchor=south] (Fig1B) at (1,0) {\includegraphics[width=0.23\textwidth]{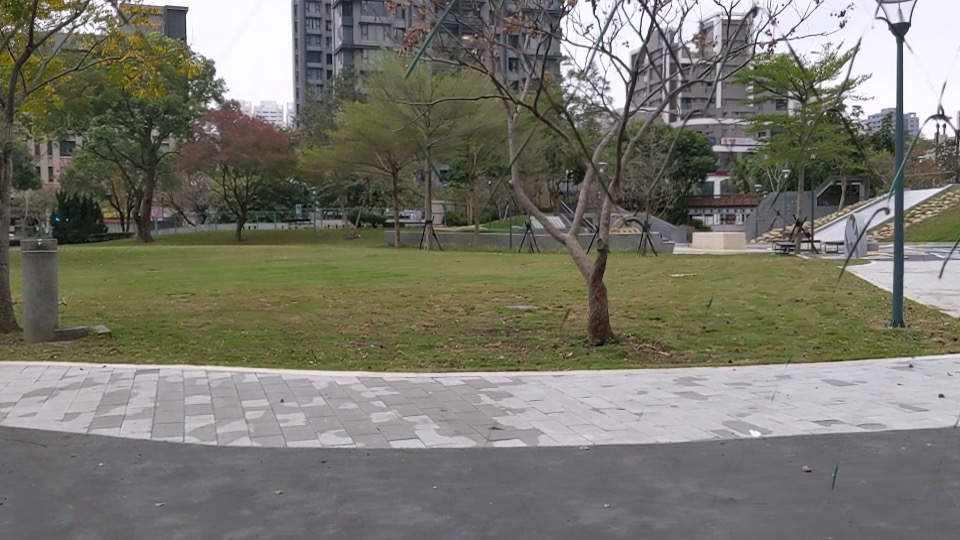}};
\spy [closeup_fence,magnification=2] on ($(Fig1B)+(0.24,0.17)$) 
    in node[largewindow,anchor=north west] at ($(Fig1B.south west) + (0.04,0)$);

\node[anchor=south] (Fig1C) at (2,0) {\includegraphics[width=0.23\textwidth]{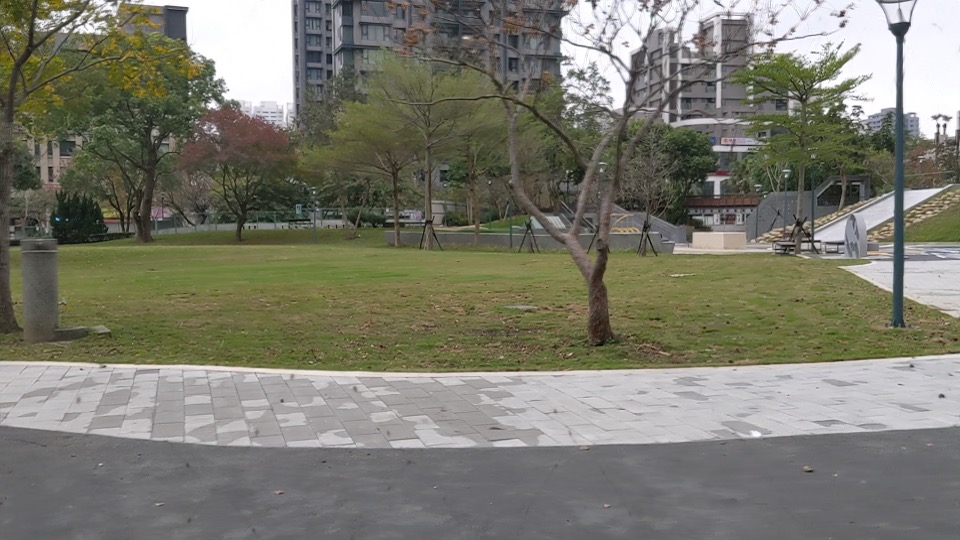}};
\spy [closeup_fence,magnification=2] on ($(Fig1C)+(0.24,0.17)$) 
    in node[largewindow,anchor=north west] at ($(Fig1C.south west) + (0.04,0)$);
    
\node[anchor=south] (Fig1D) at (3,0) {\includegraphics[width=0.23\textwidth]{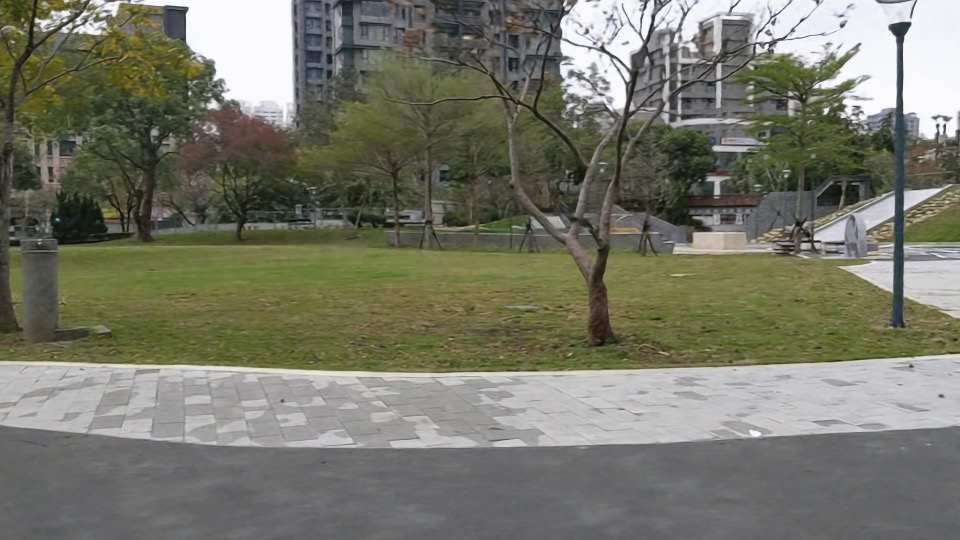}};
\spy [closeup_fence,magnification=2] on ($(Fig1D)+(0.24,0.17)$) 
    in node[largewindow,anchor=north west] at ($(Fig1D.south west) + (0.04,0)$);

\node[anchor=south] (Fig2B) at (1,-1) {\includegraphics[width=0.23\textwidth]{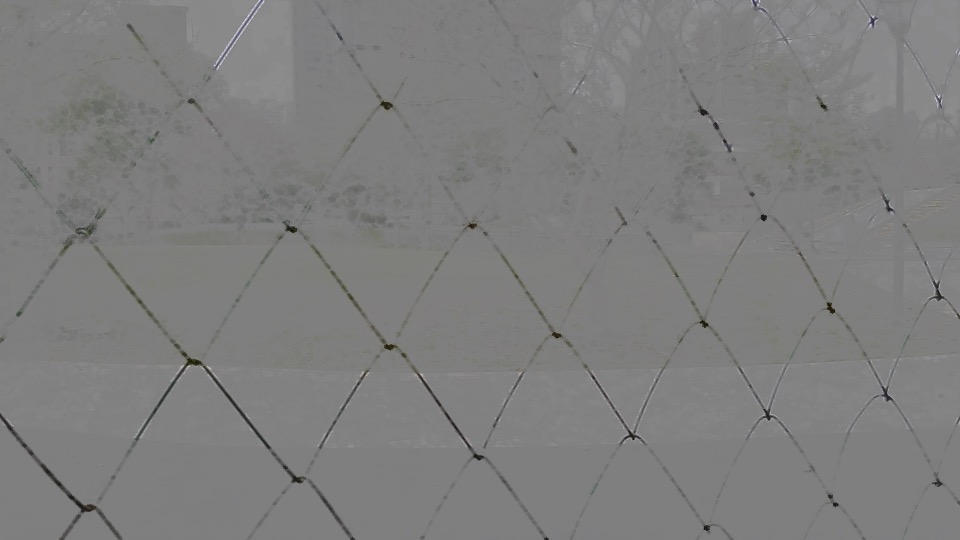}};

\node[anchor=south] (Fig2C) at (2,-1) {\includegraphics[width=0.23\textwidth]{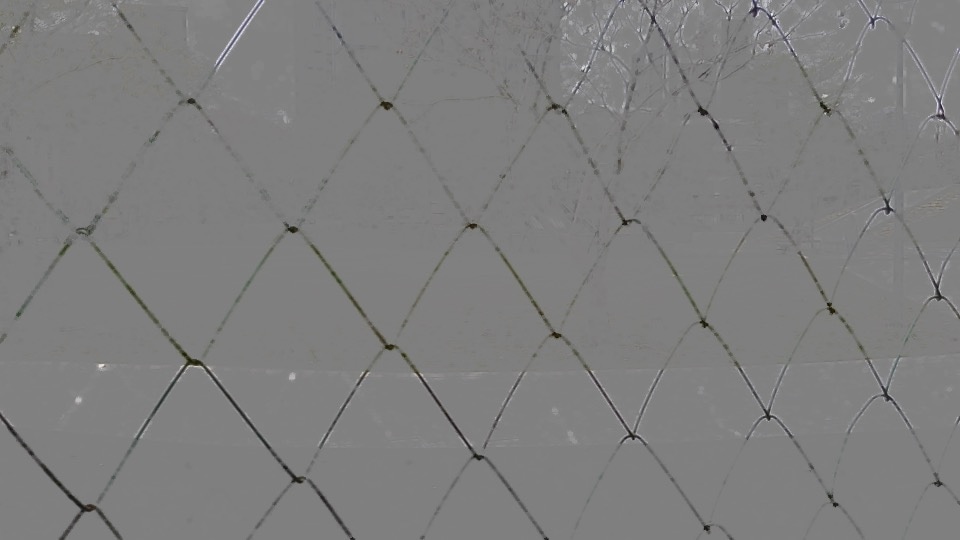}};
    
\node[anchor=south] (Fig2D) at (3,-1) {\includegraphics[width=0.23\textwidth]{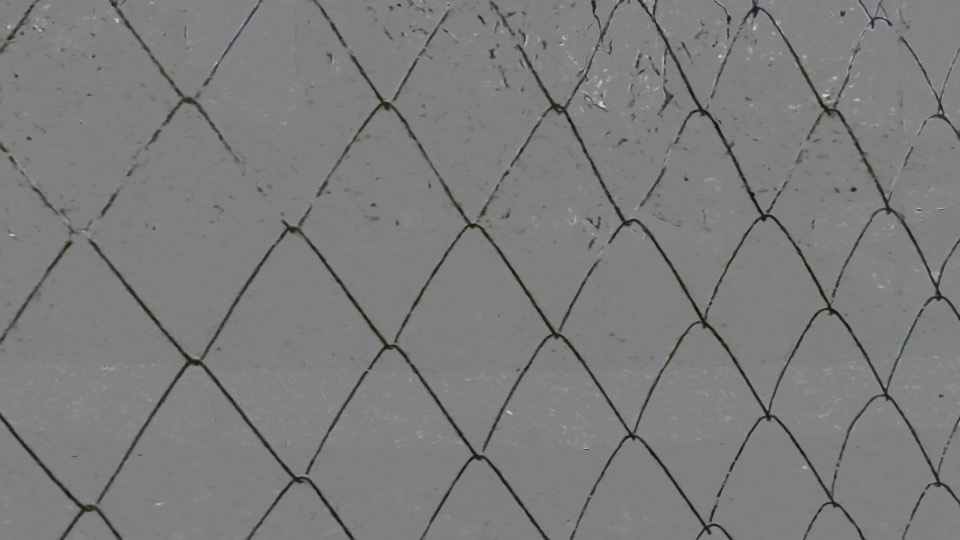}};
\end{tikzpicture}

\begin{tikzpicture}[x=0.24\textwidth, y=0.14\textheight, spy using outlines={every spy on node/.append style={smallwindow}}]
\node[anchor=south] (Fig1A) at (0,0) {\includegraphics[width=0.23\textwidth]{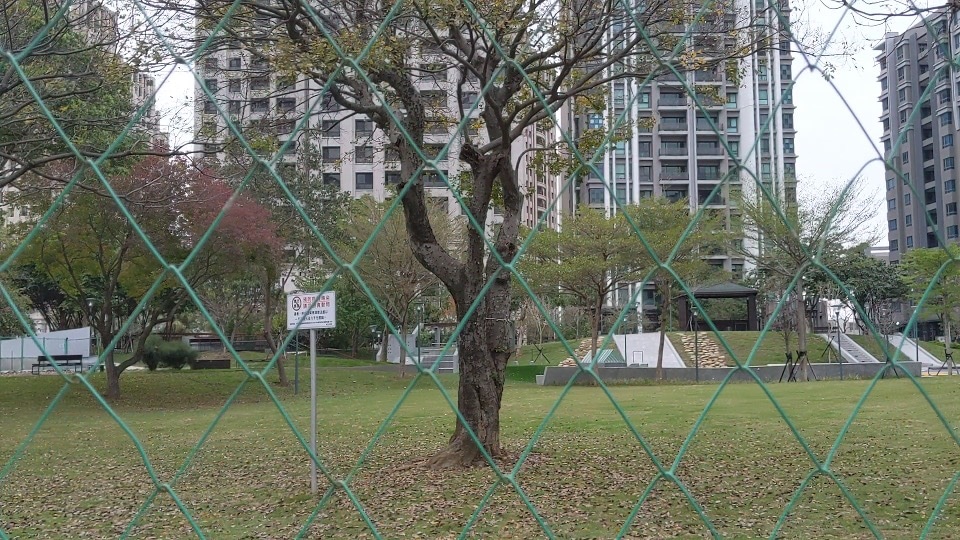}};
\spy [closeup_fence,magnification=2] on ($(Fig1A)+(0.15,0.13)$) 
    in node[largewindow,anchor=north west] at ($(Fig1A.south west) + (0.04,0)$);

\node[anchor=south] (Fig1B) at (1,0) {\includegraphics[width=0.23\textwidth]{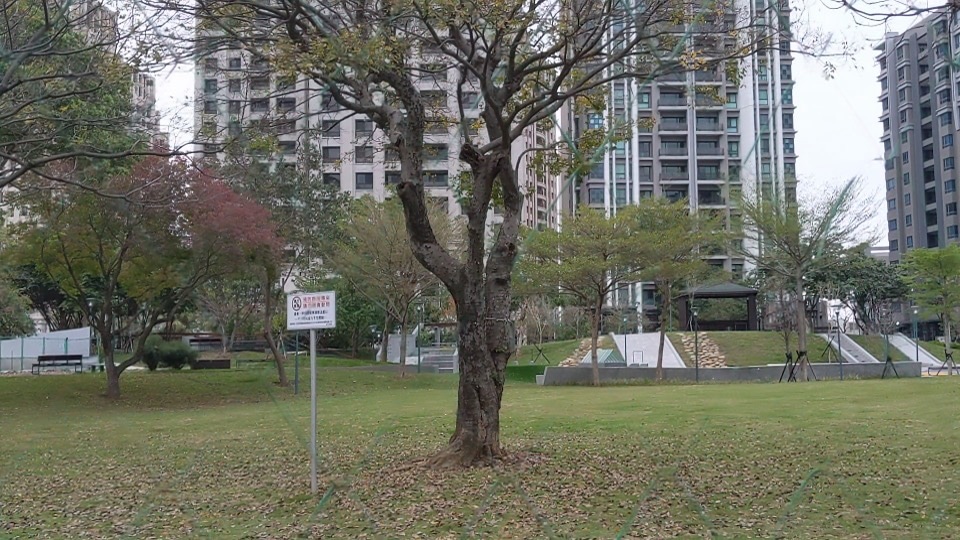}};
\spy [closeup_fence,magnification=2] on ($(Fig1B)+(0.15,0.13)$) 
    in node[largewindow,anchor=north west] at ($(Fig1B.south west) + (0.04,0)$);

\node[anchor=south] (Fig1C) at (2,0) {\includegraphics[width=0.23\textwidth]{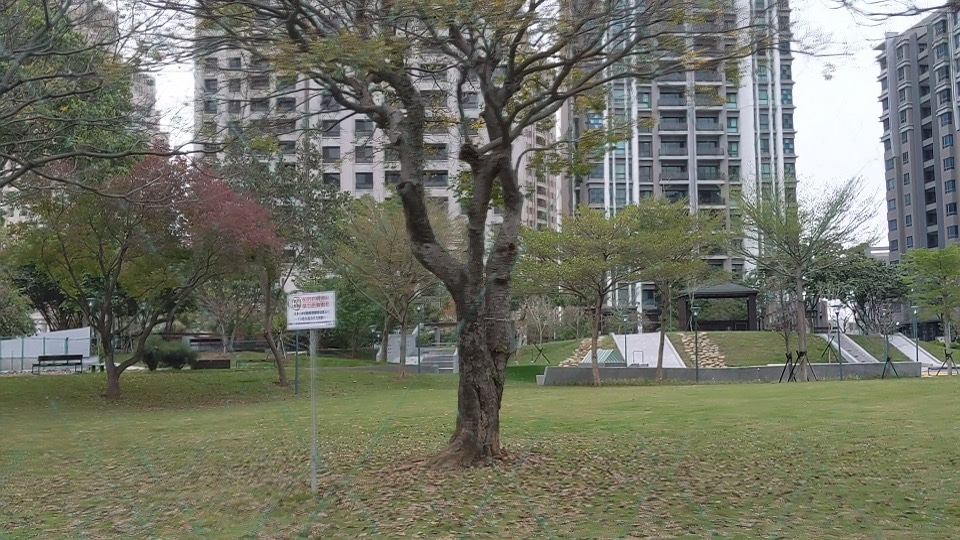}};
\spy [closeup_fence,magnification=2] on ($(Fig1C)+(0.15,0.13)$) 
    in node[largewindow,anchor=north west] at ($(Fig1C.south west) + (0.04,0)$);
    
\node[anchor=south] (Fig1D) at (3,0) {\includegraphics[width=0.23\textwidth]{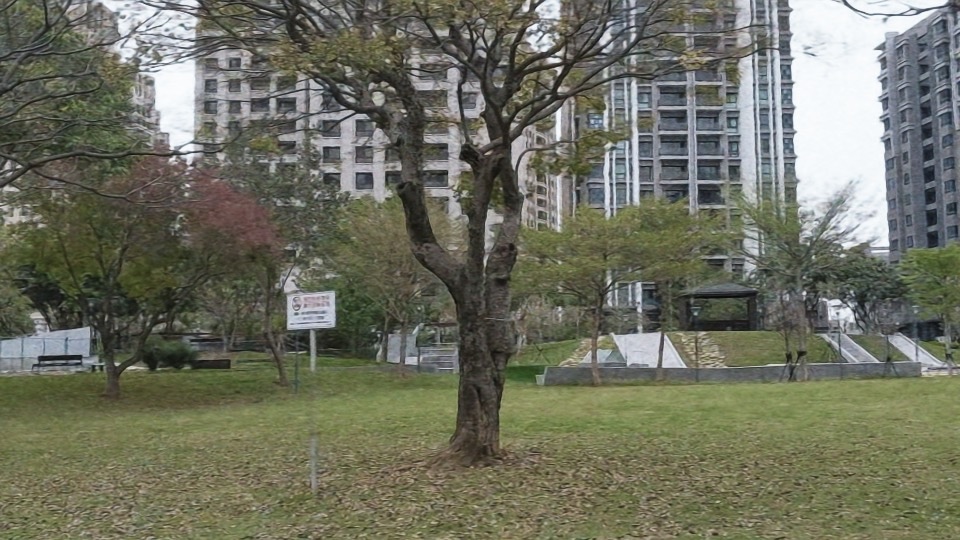}};
\spy [closeup_fence,magnification=2] on ($(Fig1D)+(0.15,0.13)$) 
    in node[largewindow,anchor=north west] at ($(Fig1D.south west) + (0.04,0)$);

\node [anchor=north] at ($(Fig1A.south)+(0,-0.01)$) {\small Input};

\node[anchor=south] (Fig2B) at (1,-1) {\includegraphics[width=0.23\textwidth]{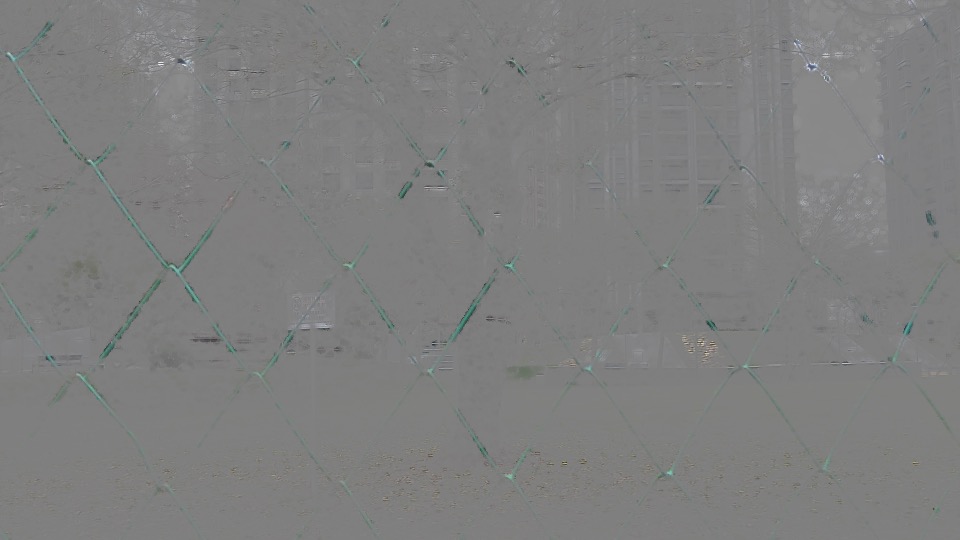}};
\node [anchor=north] at ($(Fig2B.south)+(0,-0.01)$) {\small Liu~\etal~\cite{Liu:2020:LearningToSee}};

\node[anchor=south] (Fig2C) at (2,-1) {\includegraphics[width=0.23\textwidth]{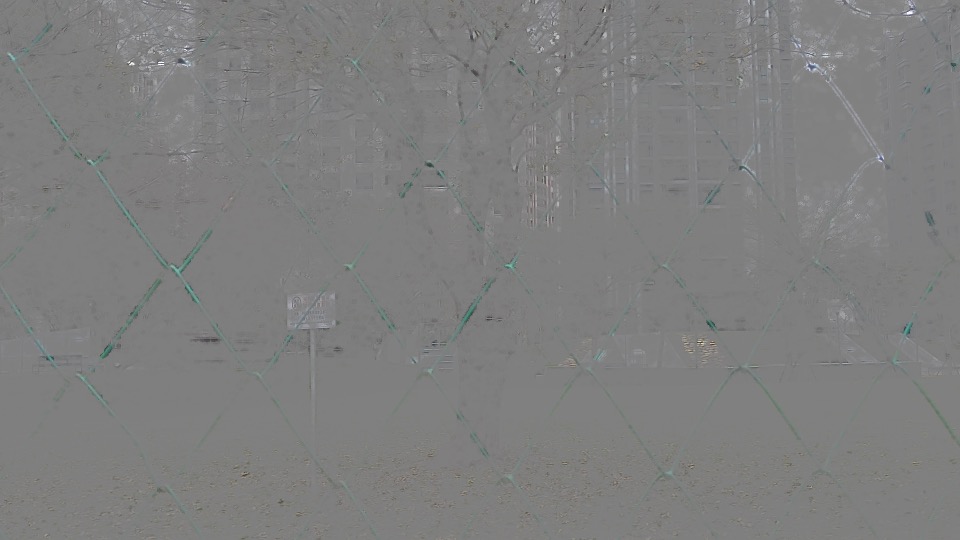}};
\node [anchor=north] at ($(Fig2C.south)+(0,-0.01)$) {\small Liu~\etal~\cite{Liu:2020:LearningToSeeJournal}};
    
\node[anchor=south] (Fig2D) at (3,-1) {\includegraphics[width=0.23\textwidth]{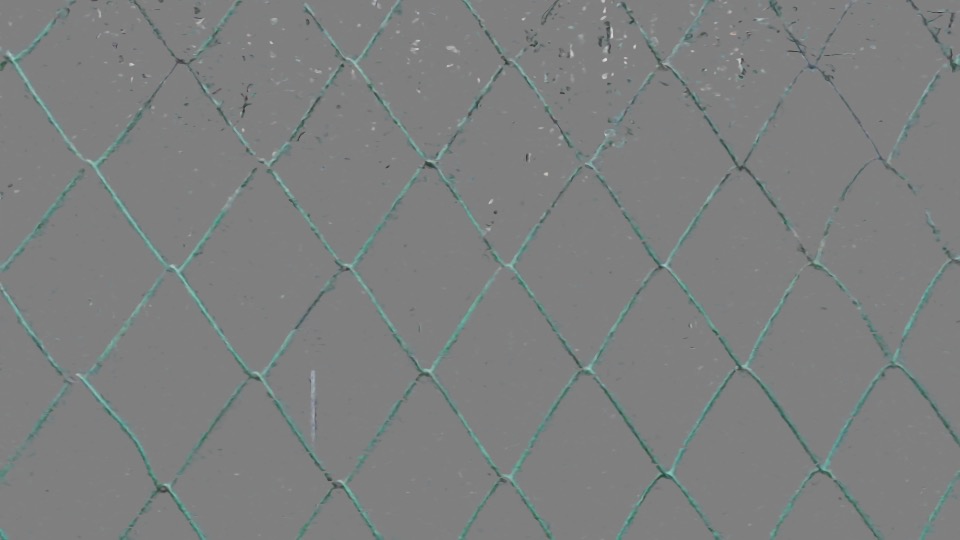}};
\node [anchor=north] at ($(Fig2D.south)+(0,-0.01)$) {\small Ours};
\end{tikzpicture}
\caption{Qualitative comparison of fence removal on real images in~\cite{Liu:2020:LearningToSeeJournal}.}
\label{fig:supp_fence}
\end{figure*}

%% file: figures/supp_rain.tex
\tikzstyle{closeup_rain} = [
  opacity=1.0,          
  height=0.9cm,         
  width=0.226\textwidth, 
  connect spies, red  
]

\begin{figure*}
\centering
\begin{tikzpicture}[x=0.24\textwidth, y=0.2\textheight, spy using outlines={every spy on node/.append style={smallwindow}}]
\node[anchor=south] (Fig2A) at (0,0) {\includegraphics[width=0.23\textwidth]{figures/rain/supp/2/input.jpg}};
\spy [closeup_rain,magnification=3] on ($(Fig2A)+(-0.25,-0.12)$) 
    in node[largewindow,anchor=north west] at ($(Fig2A.south west) + (0.04,0)$);

\node[anchor=south] (Fig2B) at (1,0) {\includegraphics[width=0.23\textwidth]{figures/rain/supp/2/fastderain.jpg}};
\spy [closeup_rain,magnification=3] on ($(Fig2B)+(-0.25,-0.12)$) 
    in node[largewindow,anchor=north west] at ($(Fig2B.south west) + (0.04,0)$);

\node[anchor=south] (Fig2C) at (2,0) {\includegraphics[width=0.23\textwidth]{figures/rain/supp/2/spaccnn.jpg}};
\spy [closeup_rain,magnification=3] on ($(Fig2C)+(-0.25,-0.12)$) 
    in node[largewindow,anchor=north west] at ($(Fig2C.south west) + (0.04,0)$);
    
\node[anchor=south] (Fig2D) at (3,0) {\includegraphics[width=0.23\textwidth]{figures/rain/supp/2/ours.jpg}};
\spy [closeup_rain,magnification=3] on ($(Fig2D)+(-0.25,-0.12)$) 
    in node[largewindow,anchor=north west] at ($(Fig2D.south west) + (0.04,0)$);
\end{tikzpicture}

\begin{tikzpicture}[x=0.24\textwidth, y=0.2\textheight, spy using outlines={every spy on node/.append style={smallwindow}}]
\node[anchor=south] (Fig1A) at (0,0) {\includegraphics[width=0.23\textwidth]{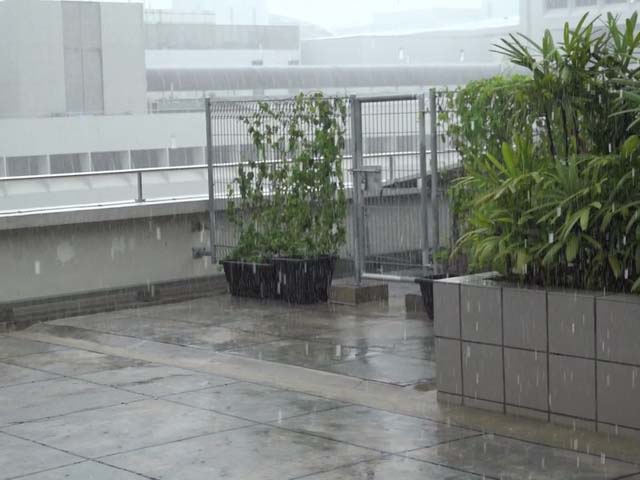}};
\spy [closeup_rain,magnification=3] on ($(Fig1A)+(-0.05,-0.05)$) 
    in node[largewindow,anchor=north west] at ($(Fig1A.south west) + (0.04,0)$);

\node[anchor=south] (Fig1B) at (1,0) {\includegraphics[width=0.23\textwidth]{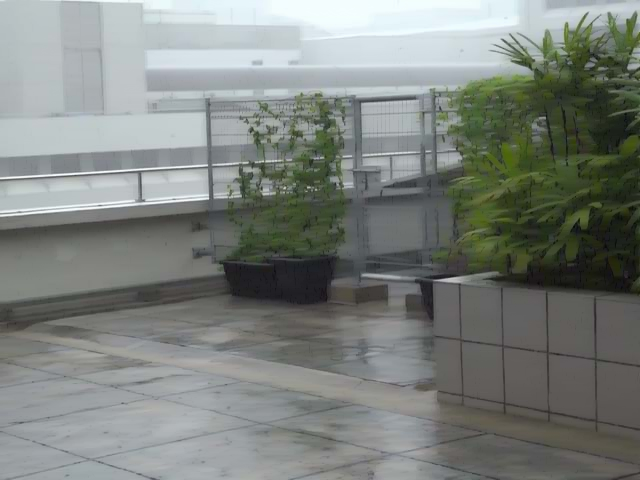}};
\spy [closeup_rain,magnification=3] on ($(Fig1B)+(-0.05,-0.05)$) 
    in node[largewindow,anchor=north west] at ($(Fig1B.south west) + (0.04,0)$);

\node[anchor=south] (Fig1C) at (2,0) {\includegraphics[width=0.23\textwidth]{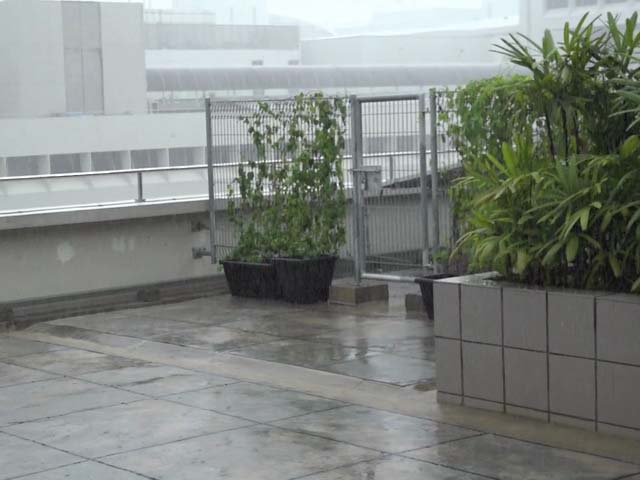}};
\spy [closeup_rain,magnification=3] on ($(Fig1C)+(-0.05,-0.05)$) 
    in node[largewindow,anchor=north west] at ($(Fig1C.south west) + (0.04,0)$);
    
\node[anchor=south] (Fig1D) at (3,0) {\includegraphics[width=0.23\textwidth]{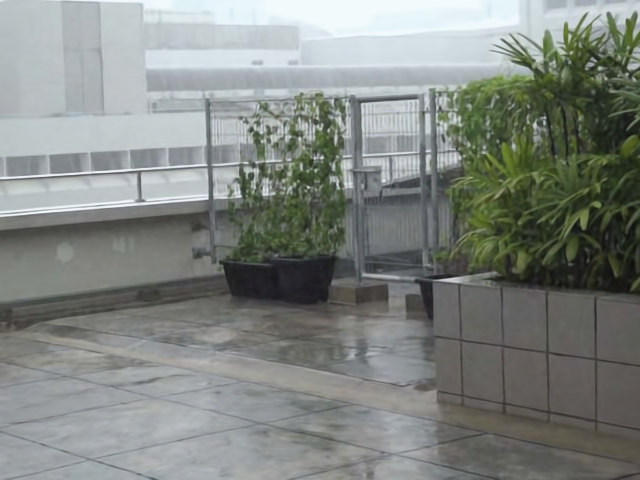}};
\spy [closeup_rain,magnification=3] on ($(Fig1D)+(-0.05,-0.05)$) 
    in node[largewindow,anchor=north west] at ($(Fig1D.south west) + (0.04,0)$);
\end{tikzpicture}



    

\begin{tikzpicture}[x=0.24\textwidth, y=0.2\textheight, spy using outlines={every spy on node/.append style={smallwindow}}]
\node[anchor=south] (Fig4A) at (0,0) {\includegraphics[width=0.23\textwidth]{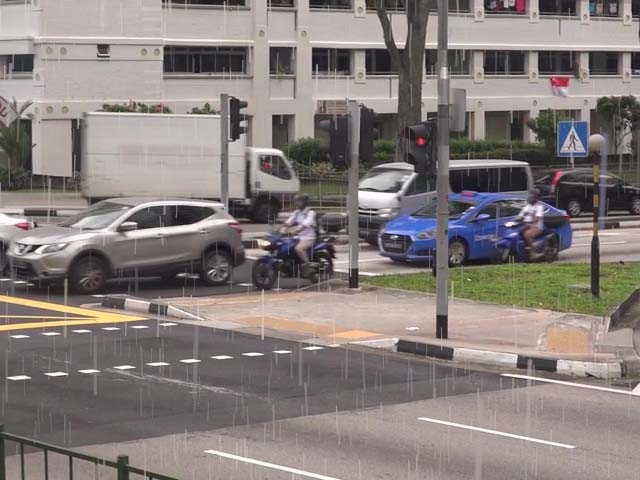}};
\spy [closeup_rain,magnification=3] on ($(Fig4A)+(0,0.05)$) 
    in node[largewindow,anchor=north west] at ($(Fig4A.south west) + (0.04,0)$);
\node [anchor=north] at ($(Fig4A.south)+(0,-0.25)$) {\small Input};

\node[anchor=south] (Fig4B) at (1,0) {\includegraphics[width=0.23\textwidth]{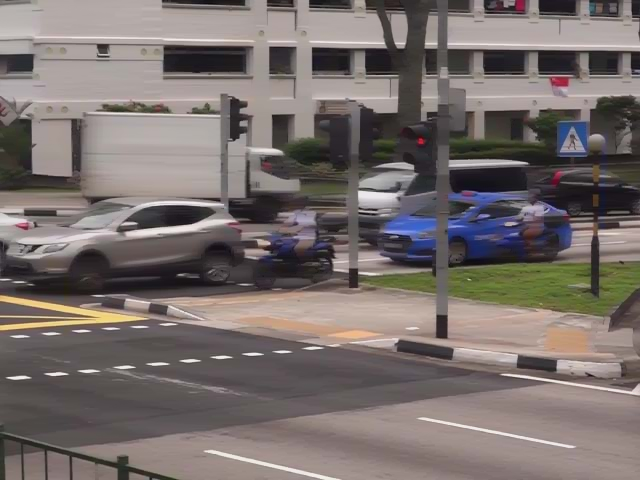}};
\spy [closeup_rain,magnification=3] on ($(Fig4B)+(0,0.05)$) 
    in node[largewindow,anchor=north west] at ($(Fig4B.south west) + (0.04,0)$);
\node [anchor=north] at ($(Fig4B.south)+(0,-0.25)$) {\small FastDeRain~\cite{Jiang:2018:FastDeRain}};

\node[anchor=south] (Fig4C) at (2,0) {\includegraphics[width=0.23\textwidth]{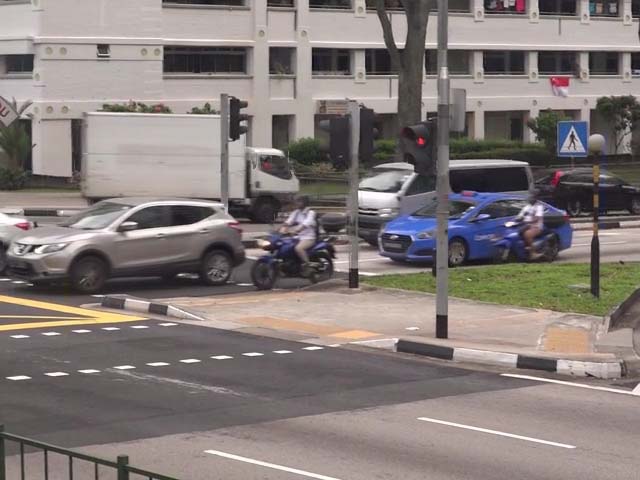}};
\spy [closeup_rain,magnification=3] on ($(Fig4C)+(0,0.05)$) 
    in node[largewindow,anchor=north west] at ($(Fig4C.south west) + (0.04,0)$);
\node [anchor=north] at ($(Fig4C.south)+(0,-0.25)$) {\small SpacCNN~\cite{Chen:2018:NTURain}};
    
\node[anchor=south] (Fig4D) at (3,0) {\includegraphics[width=0.23\textwidth]{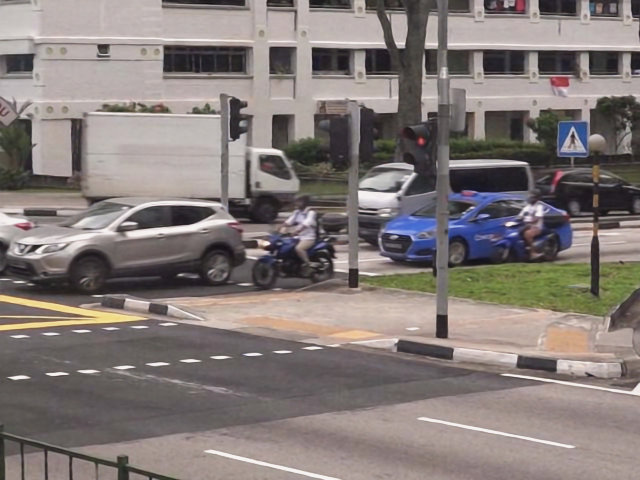}};
\spy [closeup_rain,magnification=3] on ($(Fig4D)+(0,0.05)$) 
    in node[largewindow,anchor=north west] at ($(Fig4D.south west) + (0.04,0)$);
\node [anchor=north] at ($(Fig4D.south)+(0,-0.25)$) {\small Ours};
\end{tikzpicture}
\caption{Qualitative comparison of rain removal on real images in NTURain~\cite{Chen:2018:NTURain}.}
\label{fig:supp_rain}
\end{figure*}